\documentclass[preprint,review,12pt]{elsarticle}

\usepackage{hyperref}
\usepackage{longtable}
\usepackage{amsmath,amssymb,amsfonts}
\usepackage{algorithmic}
\usepackage{graphicx}
\usepackage{textcomp}
\usepackage{dsfont}
\usepackage{booktabs} 
\usepackage{array} 
\usepackage{tabularx} 
\usepackage{makecell} 
\usepackage{algorithm}
\usepackage{algorithmic}
\usepackage{multirow}
\usepackage{booktabs} 
\usepackage{array} 
\usepackage{tabularx} 
\usepackage{makecell} 
\usepackage{multirow}
\usepackage{pifont}
\usepackage{xcolor}
\usepackage{caption}
\usepackage{float}
\definecolor{customgray}{rgb}{0.5,0.5,0.5}
\newcommand{\graytext}[1]{\textcolor{customgray}{#1}}

\makeatletter
\newenvironment{breakablealgorithm}
{
		\begin{center}
			\refstepcounter{algorithm}
			\hrule height.8pt depth0pt \kern2pt
			\renewcommand{\caption}[2][\relax]{
				{\raggedright\textbf{\ALG@name~\thealgorithm} ##2\par}
				\ifx\relax##1\relax 
				\addcontentsline{loa}{algorithm}{\protect\numberline{\thealgorithm}##2}
				\else 
				\addcontentsline{loa}{algorithm}{\protect\numberline{\thealgorithm}##1}
				\fi
				\kern2pt\hrule\kern2pt
			}
		}{
		\kern2pt\hrule\relax
	\end{center}
}
\makeatother

\journal{Medical Image Analysis}

\begin{document}

\begin{frontmatter}



\title{Post-TIPS Prediction via Multimodal Interaction: A Multi-Center Dataset and Framework for Survival, Complication, and Portal Pressure Assessment} %% Article title


\author[a]{Junhao Dong\fnref{fn1}}\ead{djh1999@bupt.edu.cn}\author[b]{Dejia Liu\fnref{fn1}}\ead{liudejia1998@163.com}\author[a]{Ruiqi Ding}\ead{dingruiqi@bupt.edu.cn}\author[a]{Zongxing Chen}\ead{czx123yf567@bupt.edu.cn}\author[a]{Yingjie Huang}\ead{hyjie@bupt.edu.cn}\author[a,c]{Fei Su}\ead{sufei@bupt.edu.cn}\author[a]{Zhu Meng\corref{cor1}}\ead{bamboo@bupt.edu.cn}\author[b]{Jianbo Zhao\corref{cor1}}\ead{zhaojianbohgl@163.com}\author[a,c]{Zhicheng Zhao\corref{cor1}}\ead{zhaozc@bupt.edu.cn} %% Author name

% \affiliation{organization={},%Department and Organization

\affiliation[a]{
% organization={School of Artificial Intelligence},
            addressline={Beijing University of Posts and Telecommunications}, 
            city={Beijing},
            postcode={100876}, 
            country={China}}

\affiliation[b]{
            organization={Division of Vascular and Interventional Radiology, Department of General Surgery, Nanfang Hospital, Southern Medical University},%Department and Organization
            addressline={No. 1838 Guangzhou Avenue North}, 
            city={Guangzhou},
            postcode={510515}, 
            country={China}}

% \affiliation[c]{organization={Graduate College for Engineers},
%             addressline={Beijing University of Posts and Telecommunications}, 
%             city={Beijing},
%             postcode={100876}, 
%             country={China}}
            
% \affiliation[d]{organization={International School},
%             addressline={Beijing University of Posts and Telecommunications}, 
%             city={Beijing},
%             postcode={100876}, 
%             country={China}}

\affiliation[c]{organization={Beijing Key Laboratory of Network System and Network Culture},
            city={Beijing},
            country={China}}
            
\cortext[cor1]{Corresponding authors.}
\fntext[fn1]{These authors contributed equally to this work.}

%% Abstract
\begin{abstract}
Transjugular intrahepatic portosystemic shunt (TIPS) is an established procedure for portal hypertension, which provides variable survival outcomes and leads to frequent overt hepatic encephalopathy (OHE), indicating the necessity of accurate preoperative prognostic modeling. Current studies typically build machine learning models from preoperative computed tomography (CT) images or clinical characteristics, but face three key challenges: (1) labor-intensive region-of-interest (ROI) annotation, (2) poor reliability and generalizability of unimodal methods, and (3) incomplete assessment from single-endpoint prediction. Moreover, the lack of publicly accessible datasets constrains its research. Therefore, we present MultiTIPS, the first public multi-center dataset for TIPS prognosis, and propose a novel multimodal prognostic framework, comprising three core modules: (1) Dual-option segmentation, which integrates semi-supervised and foundation model-based pipelines to achieve robust ROI segmentation under limited annotations and facilitate feature extraction; (2) Multimodal interaction, where multi-grained radiomics attention (MGRA), progressive orthogonal disentanglement (POD), and clinically guided prognostic enhancement (CGPE) are integrated to enable cross-modal feature interaction and complementary representation, thus improving model accuracy and robustness; (3) Multi-task prediction, where a staged training strategy is used to perform stable optimization of survival, portal pressure gradient (PPG), and OHE prediction for comprehensive prognostic assessment. Extensive experiments on MultiTIPS demonstrate the superiority of the proposed method over state-of-the-art approaches, along with cross-domain generalization and interpretability, indicating its promise for clinical application. The dataset and code are available at \url{https://github.com/djh-dzxw/TIPS_master}.
\end{abstract}


% \begin{highlights}
% \end{highlights}

%% Keywords

\end{frontmatter}



\section{Introduction}
\label{sec:introduction}
%% Use \subsubsection, \paragraph, \subparagraph commands to 
%% start 3rd, 4th and 5th level sections.
Portal hypertension (PH), secondary to chronic liver diseases such as cirrhosis, often results in severe complications such as refractory ascites and variceal bleeding \cite{ph1, ph2}, which significantly contribute to global liver disease-related mortality. To manage this, the transjugular intrahepatic portosystemic shunt (TIPS) procedure serves as a proficient treatment, effectively reducing portal pressure by creating a bypass between the portal vein and the inferior vena cava \cite{tips1}. Despite its success in lowering the recurrence risk of PH-related complications, TIPS does not provide survival benefits for all patients and is frequently accompanied by overt hepatic encephalopathy (OHE) \cite{tips2_guide,tips3_guide}. Thus, a precise preoperative prognostic model is critical for identifying suitable candidates and planning personalized treatment.

\begin{figure*}[!t]
	\centering
	\includegraphics[width=0.635\paperwidth]{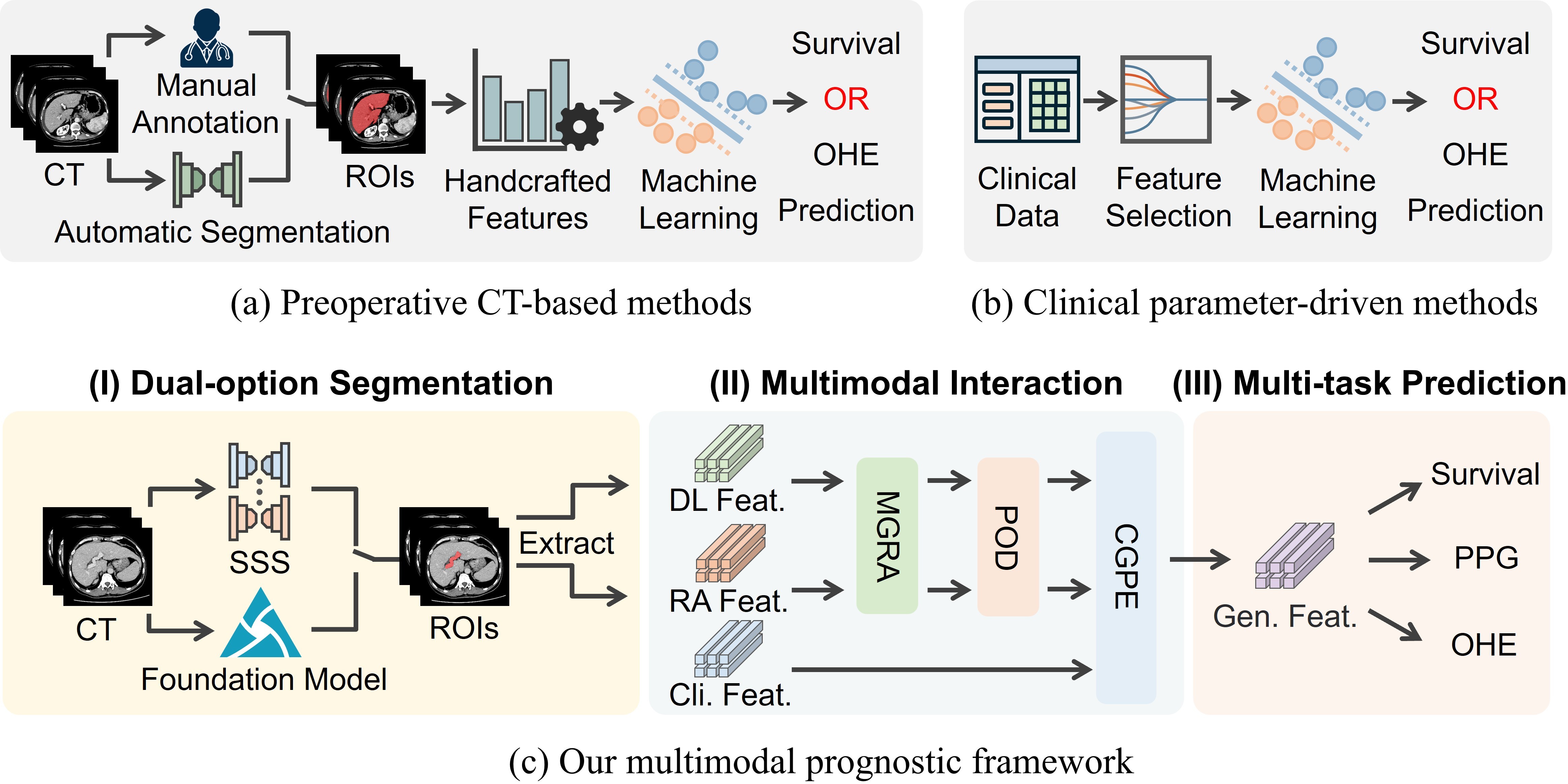}\caption{\label{fig1:process}The flowcharts of (a) preoperative CT-based methods, (b) structured clinical parameter-driven methods, and (c) our multimodal framework for TIPS prognosis. SSS: Semi-Supervised Segmentation. DL Feat.: Deep Learning Features. RA Feat.: Radiomics Features. Cli. Feat.: Clinical Features. MGRA: Multi-Grained Radiomics Attention. POD: Progressive Orthogonal Disentanglement. CGPE: Clinically Guided Prognostic Enhancement. Gen. Feat.: Generalized Features.}
\end{figure*}

Recent studies \cite{N4,N2,N6,N7} have employed machine learning algorithms for TIPS prognostic modeling. Specifically, they focus on predicting long-term survival in PH patients or assessing the occurrence of postoperative complications, especially OHE. Depending on the data utilized, these studies can be classified into two categories: (1) preoperative computed tomography (CT)–based methods, and (2) structured clinical parameter-driven methods. The core idea of the first category is to extract handcrafted features (e.g., radiomics features) from manually annotated or automatically segmented regions of interest (ROIs) and leverage traditional machine learning models for prediction (as shown in Fig.~\ref{fig1:process} (a)). For instance, in \cite{N4}, radiomics features are extracted from the delineated liver volume and fed into a linear discriminant analysis (LDA) model for OHE classification and grading. \cite{N7} implements a fully-supervised U-Net \cite{U-Net} architecture to segment skeletal muscle, visceral fat, and subcutaneous fat, facilitating the quantification of body composition metrics (area, index, and density) for each tissue. Based on these features, a logistic regression approach is developed to predict 90-day mortality after TIPS. The second category of methods relies on preoperative demographics, clinical scores, laboratory results, and procedural data for variable selection and predictive modeling (as shown in Fig.~\ref{fig1:process} (b)). For instance, \cite{N6} applies the least absolute shrinkage and selection operator (LASSO) regression to identify seven key variables, which are then incorporated into a random forest model for 1-year survival prediction.

Although machine learning-based methods have shown promising results, three key issues remain: (1) methodological limitations in CT-derived feature extraction; (2) mismatch between the unimodal framework and multimodal clinical reality; and (3) restricted prognostic assessment from single-task prediction.

The first issue involves two components, i.e., limitations of ROI annotation approaches and handcrafted feature design. On the one hand, both manual delineation and fully-supervised segmentation of ROIs require large-scale voxel-level annotations, which are time-consuming and labor-intensive, thereby hindering the clinical application of CT-based methods. On the other hand, while handcrafted features, particularly radiomics, have demonstrated effectiveness in various post-TIPS prediction tasks, their predefined computation limits the representation of prognostic information and the model’s generalization ability. To address these problems, we first propose a dual-option segmentation pipeline (depicted in Fig.~\ref{fig1:process} (c)-I) utilizing semi-supervised learning and foundation models, significantly reducing annotation cost. Specifically, the portal vein is selected as the segmentation target, as previous studies have indicated that its morphological variations are closely associated with PH and TIPS prognosis \cite{xueguan}. For the semi-supervised segmentation pipeline, 3D CT volumes are converted into 2D axial slices to alleviate the lack of labeled samples. To achieve robust portal vein segmentation, we develop a dual-stream strong data augmentation mean teacher (DSDA-MT) framework by integrating the advanced Mean Teacher \cite{mean_teacher} and UniMatch \cite{Unimatch} methods. Despite limited annotations (32 CT scans, $\sim\!10\%$), the proposed pipeline yields satisfactory segmentation results, thereby enabling effective prognostic modeling. For the foundation model-based pipeline, the same labeled data are used to fine-tune the Segment Anything Model 2 (SAM2) \cite{sam2}. More accurate prediction can be attained during inference, requiring only an extra bounding box for each ROI per slice. Therefore, clinicians can select an appropriate pipeline to balance annotation and segmentation accuracy. Moreover, we introduce deep learning (DL) features to complement the limited scope of handcrafted features. Inspired by \cite{ccl}, we employ MedicalNet \cite{MedicalNet}, a 3D medical image pretrained model, to generate multi-level voxel-wise features from segmented portal veins, ensuring precise and comprehensive representation.

For the second issue, although preoperative CT imaging and structured clinical data provide complementary information from different perspectives for prognostic evaluation, most existing approaches solely rely on one modality while ignoring valuable insights from the other, resulting in poor robustness and generalizability. Recent work \cite{T23} has proposed a combined model that directly integrates radiomics and clinical factors for OHE prediction, outperforming unimodal models. Nevertheless, it overlooks the potential correlations and interactions between multi-modalities, and the inter-modal heterogeneity hinders the effective fusion of complementary information. Motivated by this, we present a multimodal interaction framework, as shown in Fig.~\ref{fig1:process} (c)-II, aiming to enable progressive cross-modal interactions among radiomics (chosen for its common use in handcrafted features), deep learning features from the portal vein, and clinical data, thereby facilitating the generation of complementary representations. It consists of three key components: a multi-grained radiomics attention (MGRA) mechanism, a progressive orthogonal disentanglement (POD) scheme, and a clinically guided prognostic enhancement (CGPE) module. Specifically, to encourage interactions between strongly correlated CT-derived modalities, radiomics features first interact with deep features through a multi-level co-attention manner, leading to comprehensive representations. Subsequently, the two-stream features are processed by a POD scheme to gradually suppress shared redundancy and promote complementary patterns. Given the established prognostic value of clinical data \cite{N6, T30}, we employ clinical features to guide the two CT-derived streams, thus generating robust and discriminative features across all three modalities. These representations have exhibited their efficacy in diverse TIPS prognostic tasks.

As for the third concern, existing studies typically predict a single clinical endpoint (e.g., long-term survival or OHE), followed by further analyses such as patient risk stratification \cite{N6, N5} and key factor identification \cite{N2, T20}. However, single-endpoint prediction is insufficient for comprehensive postoperative assessment. Meanwhile, the use of unimodal data and methods inherently constrains the robustness and comprehensiveness of prediction outcomes and analyses. In contrast, the multimodal data and interaction framework in this work can produce representations with strong generalization capability that support multi-task prediction and comprehensive analyses. Furthermore, some recent studies \cite{ppg1, ppg2, ppg0} indicate that the post-TIPS portal pressure gradient (PPG) is a crucial prognostic indicator, defined as the pressure difference between the portal vein and the inferior vena cava. Thus, in this work, PPG prediction is also incorporated into the whole framework, resulting in a unified multi-task model covering three core prognostic dimensions: survival, complication (i.e., OHE), and PPG assessment; see Fig.~\ref{fig1:process} (c)-III. In particular, to prevent multi-task training from confusing the model, we introduce a staged training strategy: (1) Pretraining: the survival task, involving the most extensive prognostic factors, is used to train the backbone and acquire generalized features; (2) Task-specific fine-tuning: with the backbone frozen, the PPG and OHE prediction branches are sequentially fine-tuned to activate task-specific features from the generalized representations. Experimental results on a multimodal TIPS prognostic dataset validate the effectiveness of our method, which consistently outperforms the state-of-the-art (SOTA) survival models and the mono-task baselines for PPG and OHE prediction. Moreover, the results support its cross-domain generalization and demonstrate multimodal interpretability, indicating its clinical potential.

In addition, most existing studies rely on private datasets, lacking a unified benchmark for model evaluation and comparison. Therefore, we release a publicly available, multi-center, multimodal TIPS prognostic dataset to standardize evaluation and advance research in this field. Our contributions can be summarized as follows:

\begin{itemize}
\item For the first time in TIPS prognostic research, we establish MultiTIPS, a publicly accessible multi-center dataset containing preoperative CT scans, segmentation masks of key anatomical structures, clinical characteristics, and treatment outcomes from 443 patients across three distinct centers.

\item We present a novel multimodal TIPS prognostic framework comprising three components: dual-option portal vein segmentation with feature extraction, multimodal interaction, and multi-task prediction. These designs effectively alleviate the annotation burden and enhance CT-derived feature representations, make full use of multimodal information, and achieve comprehensive prognostic assessment.

\item Within multimodal interaction, the multi-grained radiomics attention (MGRA), progressive orthogonal disentanglement (POD), and clinically guided prognostic enhancement (CGPE) modules are jointly employed to facilitate cross-modal feature interaction and complementary information integration, improving predictive performance. Meanwhile, the staged training strategy ensures reliable multi-task learning and even enhances the PPG and OHE prediction.

\item Extensive experiments on the MultiTIPS dataset demonstrate the superiority of our approach over SOTA methods and mono-task models, along with cross-domain robustness and multimodal interpretability, highlighting its potential for clinical translation.

\end{itemize}

\section{MultiTIPS Dataset}

\subsection{Data Acquisition}
The MultiTIPS dataset was collected from patients who underwent TIPS for portal hypertension (PH) between March 2017 and September 2024 at three independent centers: Nanfang Hospital (NF), the First Affiliated Hospital of Guangzhou Medical University (GZ), and Guangxi People's Hospital (GX). Preoperative CT images, clinical characteristics, and postoperative outcomes were incorporated for each patient and reviewed by radiologists to exclude cases with poor imaging or incomplete clinical data. Ultimately, a total of 443 patients were included in the dataset, each with four-phase CT scans (non-contrast, arterial, portal venous, and delayed), five categories of key clinical features, and five primary treatment outcomes. Specifically, CT images from the three cohorts are stored in DICOM format, covering the entire liver, and exhibit substantial diversity in acquisition equipment, voxel sizes, and scan resolutions, as detailed in Table~\ref{tab:dataset summary}. Following recommendations from relevant clinical guidelines \cite{baveno,aasld,easl}, the preoperative clinical features are systematically classified into five groups, i.e., baseline information, pressure-related indicators, blood and biochemical indicators, clinical scores and classifications, and previous surgeries and procedures, as summarized in Table~\ref{tab: Classification of clinical features}. Moreover, the five outcome measures comprise post-TIPS PPG, overall survival (OS), and event-free survival for overt hepatic encephalopathy (OHE), variceal bleeding (VB), and stent restenosis (SR). OS is defined as the time from TIPS placement to death from any cause, whereas the other three survival metrics are measured from the same baseline to the first respective event or death. Except for the directly measured PPG, the remaining four time-dependent outcomes are assessed through scheduled follow-ups at 1 week, 1 month, 3 months, 6 months, and 12 months after TIPS, and annually thereafter. The minimum follow-up duration is set to 12 months. A summary of the outcome measures along with essential patient characteristics is presented in Table~\ref{tab:dataset summary}. Note that all the data used in this study were de-identified and acquired with the approval of the local ethics committees (NFEC-2024-106), following the Declaration of Helsinki.

% \clearpage
\scriptsize
\begin{longtable}{>{\arraybackslash}m{100pt}
                  >{\centering\arraybackslash}m{80pt}
                  >{\centering\arraybackslash}m{80pt}
                  >{\centering\arraybackslash}m{80pt}}
\caption{Summary of the MultiTIPS Dataset}\\

\toprule[1pt]
\multicolumn{1}{c}{Cohort} & NF ($n=306$) & GZ ($n=27$) & GX ($n=110$) \\
\midrule[0.5pt]
\endfirsthead

\toprule[1pt]
\multicolumn{1}{c}{Cohort} & NF ($n=306$) & GZ ($n=27$) & GX ($n=110$) \\
\midrule[0.5pt]
\endhead

\midrule[1pt]
\endfoot

\endlastfoot

\bfseries Patient characteristics & & & \\
\addlinespace[0.3em]

\hangindent=1em \hangafter=0 Time of TIPS procedures
  & Mar~2017 to Dec~2020
  & Nov~2022 to\linebreak Jul~2024
  & Mar~2017 to Sep~2024 \\
\addlinespace[0.2em]

\hangindent=1em \hangafter=0 Minimal follow-up (m)
  & 34 & 12 & 20 \\
\addlinespace[0.2em]

\hangindent=1em \hangafter=0 Age (y)
  & [14, 83] (52) & [38, 81] (59) & [25, 78] (53) \\
\addlinespace[0.2em]

\hangindent=1em \hangafter=0
\makecell[l]{Male / Female}
  & \makecell[c]{224 (73.20\%) / \\ 82 (26.80\%)}
  & \makecell[c]{20 (74.07\%) / \\ 7 (25.93\%)}
  & \makecell[c]{84 (76.36\%) / \\ 26 (23.64\%)} \\
\addlinespace[0.2em]

\hangindent=1em \hangafter=0 Pre-PPG
  & [3, 47] (21) & [18, 40] (28) & [1, 37] (22) \\
\addlinespace[0.3em]

\bfseries Treatment outcomes & & & \\
\addlinespace[0.3em]

\hangindent=1em \hangafter=0 Post-PPG
  & [1, 22] (10) & [7, 26] (15) & [0, 23] (9) \\
\addlinespace[0.2em]

\hangindent=1em \hangafter=0 OS (m)
  & [1, 78] (46) & [1, 31] (18) & [2, 106] (36) \\

\hangindent=2em \hangafter=0 \textit{Uncensored cases}
  & 65 (21.24\%) & 7 (25.93\%) & 15 (13.64\%) \\
\addlinespace[0.2em]

\hangindent=1em \hangafter=0 OHE-free survival (m)
  & [1, 78] (42) & [1, 31] (18) & [0, 106] (28) \\

\hangindent=2em \hangafter=0 \textit{Uncensored cases}
  & 53 (17.32\%) & 5 (18.52\%) & 20 (18.18\%) \\
\addlinespace[0.2em]

\hangindent=1em \hangafter=0 VB-free survival (m)
  & [1, 78] (42) & [1, 31] (18) & [0, 106] (29) \\

\hangindent=2em \hangafter=0 \textit{Uncensored cases}
  & 20 (6.54\%) & 2 (7.41\%) & 13 (11.82\%) \\
\addlinespace[0.2em]

\hangindent=1em \hangafter=0 SR-free survival (m)
  & [1, 78] (45) & [1, 31] (17) & [1, 88] (35) \\

\hangindent=2em \hangafter=0 \textit{Uncensored cases}
  & 16 (5.23\%) & 3 (11.11\%) & 9 (8.18\%) \\
\addlinespace[0.3em]

\bfseries Imaging parameters & & & \\
\addlinespace[0.3em]

\hangindent=1em \hangafter=0 \makecell[l]{Voxel spacing (mm$^3$)}
  & \makecell[c]{0.54$\times$0.54$\times$0.80 \\[-0.15em] to \\[-0.15em] 0.97$\times$0.97$\times$5.00}
  & \makecell[c]{0.63$\times$0.63$\times$1.00 \\[-0.15em] to \\[-0.15em] 0.98$\times$0.98$\times$1.00}
  & \makecell[c]{0.55$\times$0.55$\times$0.35 \\[-0.15em] to \\[-0.15em] 0.94$\times$0.94$\times$1.50} \\
\addlinespace[0.3em]

\hangindent=1em \hangafter=0 \makecell[l]{Resolution}
  & \makecell[c]{512$\times$512$\times$23 \\[-0.15em] to \\[-0.15em] 512$\times$512$\times$612}
  & \makecell[c]{512$\times$512$\times$86 \\[-0.15em] to \\[-0.15em] 512$\times$512$\times$246}
  & \makecell[c]{512$\times$512$\times$82 \\[-0.15em] to \\[-0.15em] 512$\times$512$\times$841} \\
\addlinespace[0.3em]

\hangindent=1em \hangafter=0 \makecell[l]{Manufacturer}
  & Philips (iCT 256), GE MEDICAL SYSTEMS (LightSpeed16, Revolution CT), SIEMENS (SOMATOM Definition), UIH (uCT 960+)
  & GE MEDICAL SYSTEMS (Revolution Apex/CT), SIEMENS (SOMATOM Definition AS+/Perspective)
  & Philips (iCT 256), SIEMENS (Sensation 64/Open, SOMATOM Force/go.Top), TOSHIBA (Aquilion ONE), UIH (uCT 968) \\

\bottomrule[1pt]

\multicolumn{4}{>{\arraybackslash}p{370pt}}{%
Range-based values are reported as [min, max] (median). Time units are in months (m) and years (y). Pre-PPG $=$ preoperative portal pressure gradient, Post-PPG $=$ postoperative portal pressure gradient, OS $=$ overall survival, VB $=$ variceal bleeding, SR $=$ stent restenosis.}
\label{tab:dataset summary}
\end{longtable}
\normalsize

\begin{table}[!ht]
\scriptsize
\caption{Classification of preoperative clinical features in the MultiTIPS dataset}
\label{tab: Classification of clinical features}
\centering
\setlength{\tabcolsep}{5pt}
\begin{tabular}{>{\arraybackslash}p{135pt} 
                >{\arraybackslash}p{220pt}}
\toprule[1pt]
Category & Feature list \\

\midrule[0.5pt]

Baseline information & Sex, Age, Etiology, PVT, SMVT, SVT, BCS, CTPV, HE, Ascites$^1$ \\
\addlinespace[0.3em]

Pressure-related indicators & Pre-IVCP, Pre-PVP, Pre-PPG$^2$ \\
\addlinespace[0.3em]

Blood and biochemical indicators & TBIL, ALB, Cr, ALT, Prolonged PT, INR, Na$^{\scalebox{0.8}{+}}$, PLT, WBC$^3$ \\
\addlinespace[0.3em]

Clinical scores and classifications & Child-Pugh score, Child-Pugh classification, MELD score, MELD classification, MELD-Na score, ALBI score, ALBI classification, FIPS score, CLIF-C AD score, CLIF-C AD classification$^4$ \\
\addlinespace[0.3em]

Previous surgeries and procedures & EBL, Partial splenectomy, EIS, GCVE, PSE$^5$ \\
\toprule[1pt]

\multicolumn{2}{p{370pt}}{$^1$PVT $=$ portal vein thrombosis, SMVT $=$ superior mesenteric vein thrombosis, SVT $=$ splenic vein thrombosis, BCS $=$ Budd-Chiari syndrome, CTPV $=$ cavernous transformation of the portal vein, HE $=$ hepatic encephalopathy.}\\
\addlinespace[0.2em]

\multicolumn{2}{p{370pt}}{$^2$Pre-IVCP $=$ preoperative inferior vena cava pressure, Pre-PVP $=$ preoperative portal venous pressure, Pre-PPG $=$ preoperative portal pressure gradient.}\\
\addlinespace[0.2em]

\multicolumn{2}{p{370pt}}{$^3$TBIL $=$ total bilirubin, ALB $=$ albumin, Cr $=$ creatinine, ALT $=$ alanine aminotransferase, PT $=$ prothrombin time, INR $=$ international normalized ratio, Na$^{\scalebox{0.8}{+}}$ $=$ serum sodium ion, PLT $=$ platelet count, WBC $=$ white blood cell count.}\\
\addlinespace[0.2em]

\multicolumn{2}{p{370pt}}{$^4$MELD $=$ model for end-stage liver disease, Na $=$ sodium, ALBI $=$ albumin-bilirubin, FIPS $=$ Freiburg index of post-TIPS survival, CLIF-C AD $=$ chronic liver failure consortium acute decompensation.}\\
\addlinespace[0.2em]

\multicolumn{2}{p{370pt}}{$^5$EBL $=$ endoscopic band ligation, EIS $=$ endoscopic injection sclerotherapy, GCVE $=$ gastric coronary vein embolization, PSE $=$ partial splenic embolization.}
\end{tabular}
% \label{tab3}
\end{table}
\normalsize

\subsection{Annotation Protocol for CT Scans}
\begin{figure*}[!t]
    \centering
    \includegraphics[width=0.63\paperwidth]{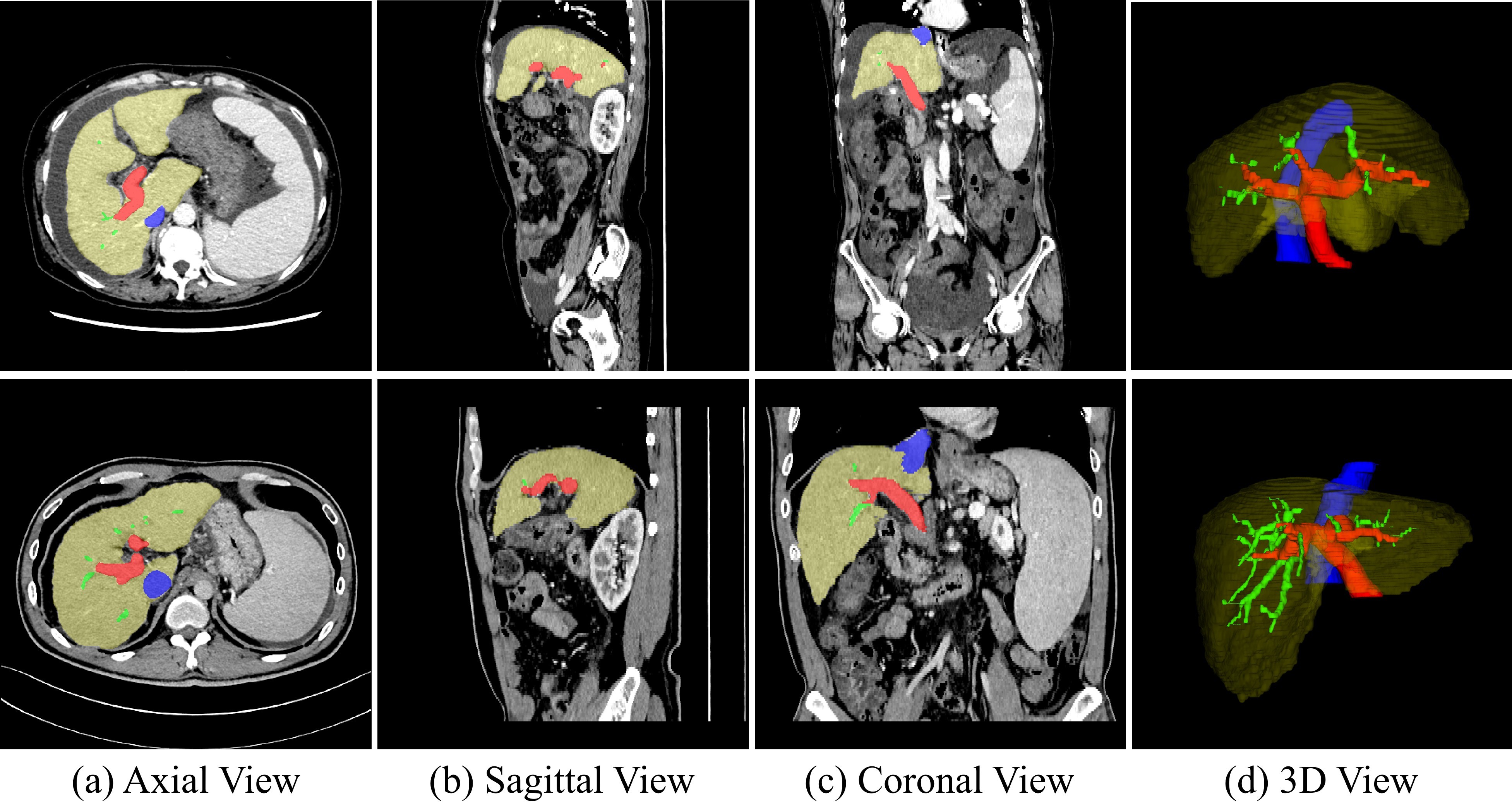}
    \caption{Illustrative examples from two patients in the MultiTIPS dataset. Columns (a)--(d) show annotations in axial, sagittal, coronal, and 3D views, respectively. Red, green, blue, and yellow represent the central portal vein, peripheral portal vein, inferior vena cava, and liver.}
    \label{fig:MultiTIPS_dataset_v2}
\end{figure*}

The annotation protocol focuses on three anatomical structures that are clinically relevant to TIPS: the liver, portal vein, and inferior vena cava. Based on the suggested operative site \cite{aasld, easl} and imaging quality, the portal vein was divided into central (main trunk, left/right first-order, and clearly visualized second-order branches) and peripheral (branches beyond second-order) segments for delineation. To enhance visualization, the portal phase CT scans served as the reference. The annotation procedures are described as follows: (1) 32 CT images from randomly selected NF patients were independently annotated by two radiologists (A and B; five years of experience) using ITK-SNAP \cite{itk-snap}, with labels for the liver, central portal vein, and inferior vena cava. (2) A semi-supervised approach (see Section~\ref{sec:semi-supervised segmentation} for details) was employed to segment the three structures separately for the remaining NF cases and all GZ and GX patients. (3) Radiologists A and B meticulously refined the segmentation masks, and their annotations were then reviewed by two experts, C and D, each with over ten years of experience. (4) For the peripheral portal vein segments with suboptimal clarity, all images were evenly assigned to experts C and D for careful delineation and subsequently evaluated by another senior expert, E. In cases of disputed annotations, the two involved experts worked to reach a consensus. The annotated results of the MultiTIPS dataset are illustrated in Fig.~\ref{fig:MultiTIPS_dataset_v2}.

\subsection{Data Usage}
\label{sec: data usage}
TIPS-related clinical practice guidelines \cite{tips_guideline_2016, aasld, easl} indicate that imaging protocols, interventional procedures, and perioperative management remain generally stable over the study period, suggesting a limited impact of temporal shifts across patient enrollment. Additionally, given its potential clinical utility, the model is expected to demonstrate strong performance on contemporaneous data while also generalizing to subsequently enrolled cases. Accordingly, the NF cohort, with sufficient training data and follow-up duration, serves as the internal dataset ($n=306$) for model development, while the GZ and GX cohorts, including both contemporaneous and more recent patients, are combined as the external test set ($n=137$) to evaluate spatial and temporal generalizability across centers and time periods. Based on the research objectives and data quality, portal phase CT images, central portal vein annotations, and five groups of clinical characteristics are leveraged to predict three major outcomes for TIPS prognosis: OS, PPG, and OHE-free survival. Particularly, in this study, only 32 patients’ voxel-level annotations (randomly sampled from the NF cohort) are employed to confirm that the proposed multimodal TIPS prognostic framework can achieve strong robustness and applicability under limited annotation conditions.

Notably, only a subset of the data components in MultiTIPS is utilized for this work, while its multiphase imaging, extensive annotations, and diverse clinical endpoints hold considerable promise for exploring more comprehensive TIPS prognostic methods.

\section{Proposed Approach}

Motivated by the clinical practice of TIPS prognosis and the available multimodal data, we propose a novel multimodal TIPS prognostic framework consisting of three components (as shown in Fig.~\ref{fig1:process} (c)): dual-option portal vein segmentation with feature extraction (Section~\ref{sec: Dual-Option Portal Vein Segmentation with Feature Extraction}), multimodal interaction (Section~\ref{sec:Multimodal Interaction}), and multi-task prediction (Section~\ref{sec:Multi-Task Prediction}), which jointly support comprehensive post-TIPS prognostic assessment. Firstly, to balance annotation and segmentation accuracy, the dual-option segmentation module is employed to generate reliable portal vein labels, thereby enabling deep learning and radiomics feature extraction for subsequent multimodal interaction. Together with clinical features, these two feature sets are then fed into the multimodal interaction module to capture complementary information and produce a unified generalized representation. Finally, the representation is shared across the multi-task prediction heads, which undergoes optimization and task-specific feature activation using a staged training strategy for comprehensive prognostic assessment.

\subsection{Dual-Option Portal Vein Segmentation with Feature Extraction}
\label{sec: Dual-Option Portal Vein Segmentation with Feature Extraction}
Based on the voxel-level annotations, patients are divided into a labeled training set $T_L^{tr}$, a labeled validation set $T_L^{val}$, and an unlabeled set $T_U$. To augment the labeled training data, all 3D CT volumes from $T_L^{tr}$ and $T_U$ are sliced axially into 2D images, resulting in $T_L^{{tr}} = \{ (x_i^l, y_i^l) \}_{i=1}^{L_{{tr}}}$ and $T_U = \{ x_i^u \}_{i=1}^U$.

\subsubsection{Semi-Supervised Segmentation Pipeline}
\label{sec:semi-supervised segmentation}
The core of semi-supervised semantic segmentation is to learn initial semantics from a small labeled set and then effectively leverage a large unlabeled set through latent supervision. Accordingly, the unlabeled set $T_U$ is further partitioned into an unlabeled training set $T_U^{tr}$ and a test set $T_U^{te}$ at the patient level.

\begin{figure*}[!t]
    \centering
    \includegraphics[width=0.6\paperwidth]{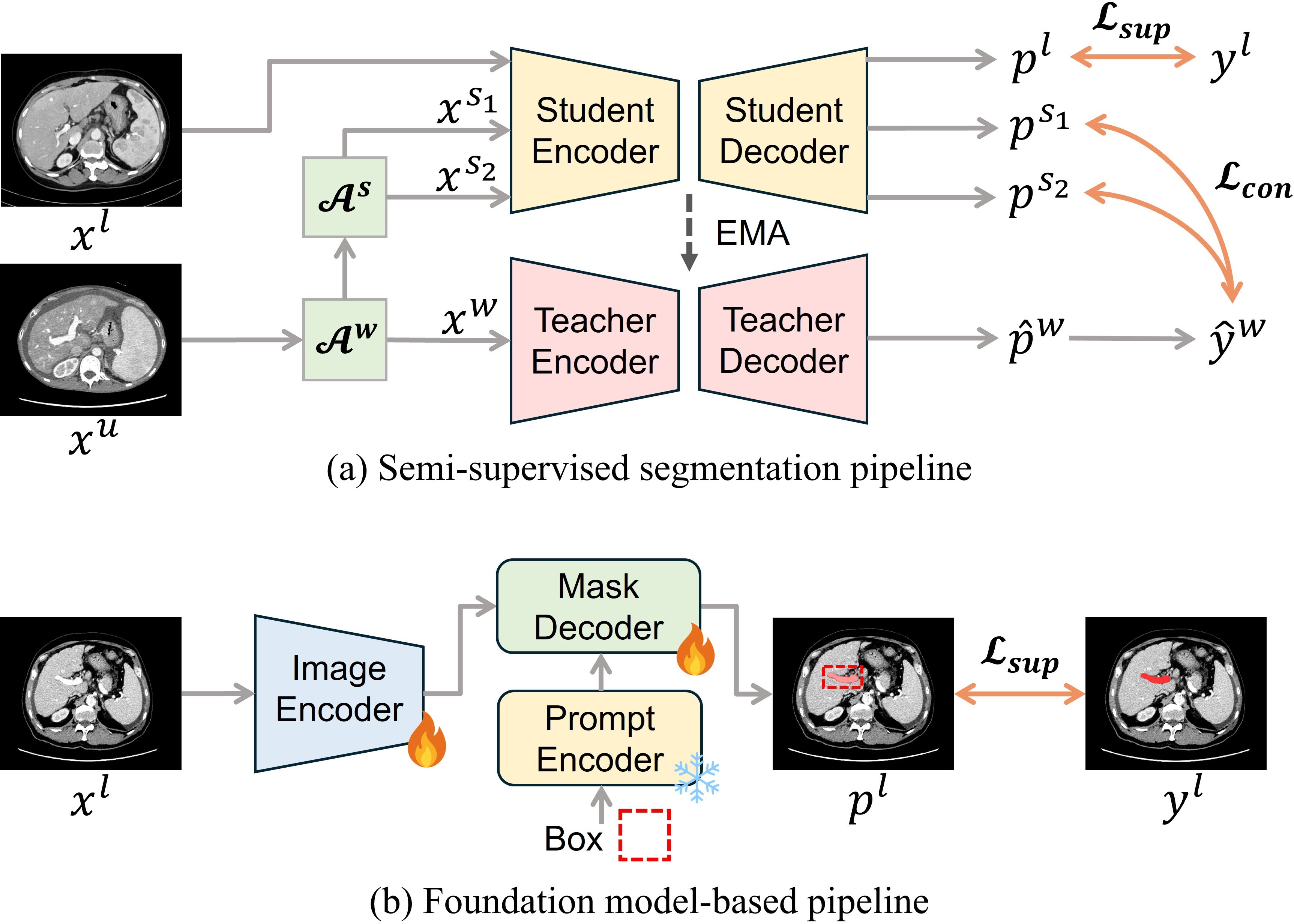}
    \caption{Dual-option pipeline for portal vein segmentation, including (a) a semi-supervised approach (DSDA-MT) and (b) a foundation model-based approach (MedSAM2).}
    \label{fig:dual-segment-pipeline}
\end{figure*}

Using the training samples from $T_L^{tr}$ and $T_U^{tr}$, as illustrated in Fig.~\ref{fig:dual-segment-pipeline} (a), a dual-stream strong data augmentation mean teacher (DSDA-MT) pipeline is proposed by combining the complementary strengths of the Mean Teacher \cite{mean_teacher} and UniMatch \cite{Unimatch} methods: Mean Teacher is applied to achieve stable and robust pseudo-label generation, while the dual-stream strong perturbation in UniMatch enforces targeted consistency across multiple views, thereby improving the generalization.

Concretely, a student segmentation network $\mathcal{F}$ and a teacher network $\hat{\mathcal{F}}$ are adopted, both composed of an image encoder and a mask decoder. $\mathcal{F}$ is optimized via backpropagation, while $\hat{\mathcal{F}}$ is updated as its exponential moving average (EMA).

For supervised learning, given a batch of labeled data $\{(x_i^l, y_i^l)\}_{i=1}^{B_l}$ from $T_L^{tr}$, the predictions of the student model are supervised by the following loss function:
\begin{equation}\label{eq:Lsup}
\mathcal{L}_{sup} = \mathcal{E}(P^l, Y^l, \mathbf{1}),
\end{equation}
where $P^l=\operatorname{Concat}\left(\left\{\sigma(\mathcal{F}(x_i^l))\right\}_{i=1}^{B_l}\right)$ and $Y^l=\operatorname{Concat}\left(\left\{y_i^l\right\}_{i=1}^{B_l}\right)$ represent the predicted foreground (portal vein) probabilities and corresponding binary ground-truth (GT) labels for this batch, with $\sigma$ denoting the sigmoid activation function. Let $\mathbf{1}$ represent a binary mask of all ones, indicating that all pixels are involved in the computation. $\mathcal{E}(\cdot)$ refers to the proposed weighted binary dice loss, designed to mitigate the imbalance between the portal vein and the background, which is defined as:
\begin{equation}\label{eq:E(·)}
\begin{split}
\mathcal{E}(P, Y, M) &= \alpha \mathcal{L}_{dice}(P \odot M, Y \odot M) \\
& + (1 - \alpha)\mathcal{L}_{dice}((\mathbf{1} - P) \odot M, \tilde{Y} \odot M),
\end{split}
\end{equation}
where the first and second terms are the dice losses for the foreground and background, respectively. $\alpha$ is a weight parameter, and $\odot ~M$ denotes element-wise multiplication with the valid mask. $(\mathbf{1} - P)$ and $\tilde{Y}$ represent the inverted probabilities and binary labels used to calculate the background loss.

For unlabeled images $\{x_i^u\}_{i=1}^{B_u}$, weak and strong augmentation strategies, $\mathcal{A}^w(\cdot)$ and $\mathcal{A}^s(\cdot)$, are applied to generate multiple views for each image:
\begin{equation}\label{eq:x_w_s1_s2}
x_i^w = \mathcal{A}^w(x_i^u),\quad x_i^{s_1} = \mathcal{A}^s(x_i^w),\quad x_i^{s_2} = \mathcal{A}^s(x_i^w),
\end{equation}
where $x_i^{s_1}$ and $x_i^{s_2}$ are two augmented versions of the image produced by dual-stream strong perturbations. Note that they are not identical, as the predefined strong augmentation pool $\mathcal{A}^s(\cdot)$ is not deterministic. Consequently, a weak-to-strong consistency regularization is implemented between the strongly and weakly augmented versions, formulated as:
\begin{equation}\label{eq:L_con}
\mathcal{L}_{con} = \frac{1}{2} \left( \mathcal{E}(P^{s_1}, \hat{Y}^w, M^w) + \mathcal{E}(P^{s_2}, \hat{Y}^w, M^w) \right).
\end{equation}

Here, $P^{s_1}$ and $P^{s_2}$ denote the student model’s predicted probabilities for the two strongly augmented versions of the batch. $\hat{Y}^w = \operatorname{argmax}(\mathbf{1} - \hat{P}^w, \hat{P}^w)$ represents the pseudo-labels generated by the teacher model on the weakly augmented inputs, where $\hat{P}^w = \operatorname{Concat}\left( \left\{ \sigma(\hat{\mathcal{F}}(x_i^w)) \right\}_{i=1}^{B_u} \right)$. The mask $M^w = \mathds{1}\left( \max(\mathbf{1} - \hat{P}^w, \hat{P}^w) \geq E(\tau) \right)$ is used to filter out uncertain pseudo-labels, where $\mathds{1}(\cdot)$ is an indicator function and $E(\tau)$ denotes a dynamic confidence threshold.

Therefore, the total loss function can be formulated as: 
\begin{equation}\label{eq:L_total}
\mathcal{L} = \mathcal{L}_{sup} + \lambda \mathcal{L}_{{con}},
\end{equation}
where $\lambda$ is the weight assigned to the consistency loss, and is set to 1 in our experiments.

\subsubsection{Foundation Model-Based Pipeline}
Considering that Segment Anything Model 2 (SAM2) \cite{sam2} possesses excellent performance, driven by a multi-scale Hiera \cite{hiera} image encoder, we use SAM2 and fine-tune it on the labeled training set $T_L^{tr}$ for portal vein segmentation in 2D CT images. The training pipeline is depicted in Fig.~\ref{fig:dual-segment-pipeline} (b).

Specifically, inspired by MedSAM2 \cite{medsam2_ours}, we introduce a fine-tuning strategy that incorporates three pretrained modules: an image encoder, a prompt encoder, and a mask decoder. During training, the image encoder and mask decoder are updated to adapt to CT-specific characteristics, whereas the prompt encoder remains frozen due to its domain-agnostic nature. For each CT slice, the image encoder extracts feature embeddings, while the prompt encoder derives prompt features from bounding boxes generated as the minimum bounding rectangles of each ROI in the ground-truth mask. Both image and prompt features are jointly fed into the mask decoder to produce the predictions. With a batch of samples $\{(x_i^l, y_i^l)\}_{i=1}^{B_l}$, the loss function remains consistent with Eq.~(\ref{eq:Lsup}), except that $\mathcal{E}(\cdot)$ is computed as an unweighted sum of dice and cross-entropy losses:
\begin{equation}\label{eq:medsam2_loss}
\mathcal{E}(P, Y, M) = \mathcal{L}_{dice}(P \odot M, Y \odot M) + \mathcal{L}_{ce}(P \odot M, Y \odot M),
\end{equation}
where $\mathcal{L}_{ce}$ represents the cross-entropy loss.

\subsubsection{Feature Extraction}
We first employ the trained segmentation model to generate portal vein labels, which are then used for feature extraction. For the semi-supervised pipeline, five-fold cross-validation is adopted to divide $T_U$ into $T_U^{tr}$ and $T_U^{te}$, ensuring that each patient in $T_U$ serves once as a test case for segmentation. The teacher network is applied to yield the stable predictions. For the foundation model-based pipeline, the fine-tuned model performs inference on the entire $T_U$, but requires a bounding box prompt for each ROI. After the inference stage, 2D segmentation results are reassembled into 3D volumes. Consequently, each patient from $T_L^{tr} \cup T_L^{val} \cup T_U$ includes a 3D CT scan $x_{3D}$ and a label volume $y_{3D}$ (patient indices omitted for ease of notation). For patients with a valid $y_{3D}$ containing portal-vein voxels, the processing steps are as follows: (1) $x_{3D}$ and $y_{3D}$ are resampled to an isotropic voxel spacing of $1\times 1\times 1 mm^3$. (2) To connect potentially discontinuous components, the portal vein annotations in $y_{3D}$ are dilated with an $m\times m\times m$ cubic kernel, yielding $y_{3D}^d$. (3) A minimal 3D bounding box enclosing the largest connected component in $y_{3D}^d$ is extracted, expanded by $d\%$ along each spatial dimension, and used to crop $x_{3D}$ and $y_{3D}$, resulting in $x_{3D}^{crop}$ and $y_{3D}^{crop}$. (4) $x_{3D}^{crop}$ and $y_{3D}^{crop}$ are adjusted to a uniform size of $n\times n\times n$ using zero padding or center cropping, producing $\tilde{x}_{3D}^{{crop}}$ and $\tilde{y}_{3D}^{{crop}}$. These steps preserve the true-volume portal vein region and surrounding structures, with dimensions adapted to the network input. The parameters $m$, $d$, and $n$ are set to 2, 20, and 128, respectively.

To enhance the representation of prognostic visual information, both deep learning and radiomics features are extracted for each patient with a valid $y_{3D}$. For the former, voxel-level features of the annotated portal vein are obtained using MedicalNet \cite{MedicalNet} (with 3D ResNet-50 \cite{resnet} as the backbone). To capture multi-scale contextual semantics, outputs from the first and third blocks (256 and 1,024 dimensions) of $\tilde{x}_{3D}^{{crop}}$ are concatenated into a 1280-dimensional feature map. Then, $\tilde{y}_{3D}^{{crop}}$ serves as a mask to select voxel-wise features within the portal vein region. This process can be formulated as $F^{d, pre} = \left\{ f_i^{d, pre} \in {MedicalNet}(\tilde{x}_{3D}^{{crop}}) \;\middle|\; i \in \Omega, \; \tilde{y}_{3D}^{{crop}}(i) = 1 \right\} = \left\{ f_s^{d, pre} \right\}_{s=1}^{N_d}
$, where $\Omega = \{1, 2, \dots, n^3\}
$ denotes the set of voxel indices, and $N_d$ is the total number of selected features for the patient.

Radiomics features typically consist of $1 \times 1$ attributes. For each patient with portal-vein voxels in $y_{3D}$, these features are extracted from the resampled $x_{3D}$ and $y_{3D}$ using Pyradiomics\footnote{https://github.com/Radiomics/pyradiomics.} (version 3.1.0). In total, 1,595 radiomics features are derived from 17 types of preprocessed images (including the original) and 7 feature classes, represented as $F^{r, pre}={Pyradiomics}(x_{3D}, y_{3D})=\{ f_j^{r, pre} \}_{j=1}^{N_r}$, where $N_r=1,595$.

Additionally, for cases with no portal-vein voxels in $y_{3D}$ due to segmentation failure, a single 1280-dimensional zero vector serves as the deep learning feature $F^{d,pre}$, and all radiomics feature values are set to zero to form $F^{r,pre}$, thereby ensuring compatibility with the subsequent multimodal framework.

\begin{figure*}[!t]
	\centering
	\includegraphics[width=0.655\paperwidth]{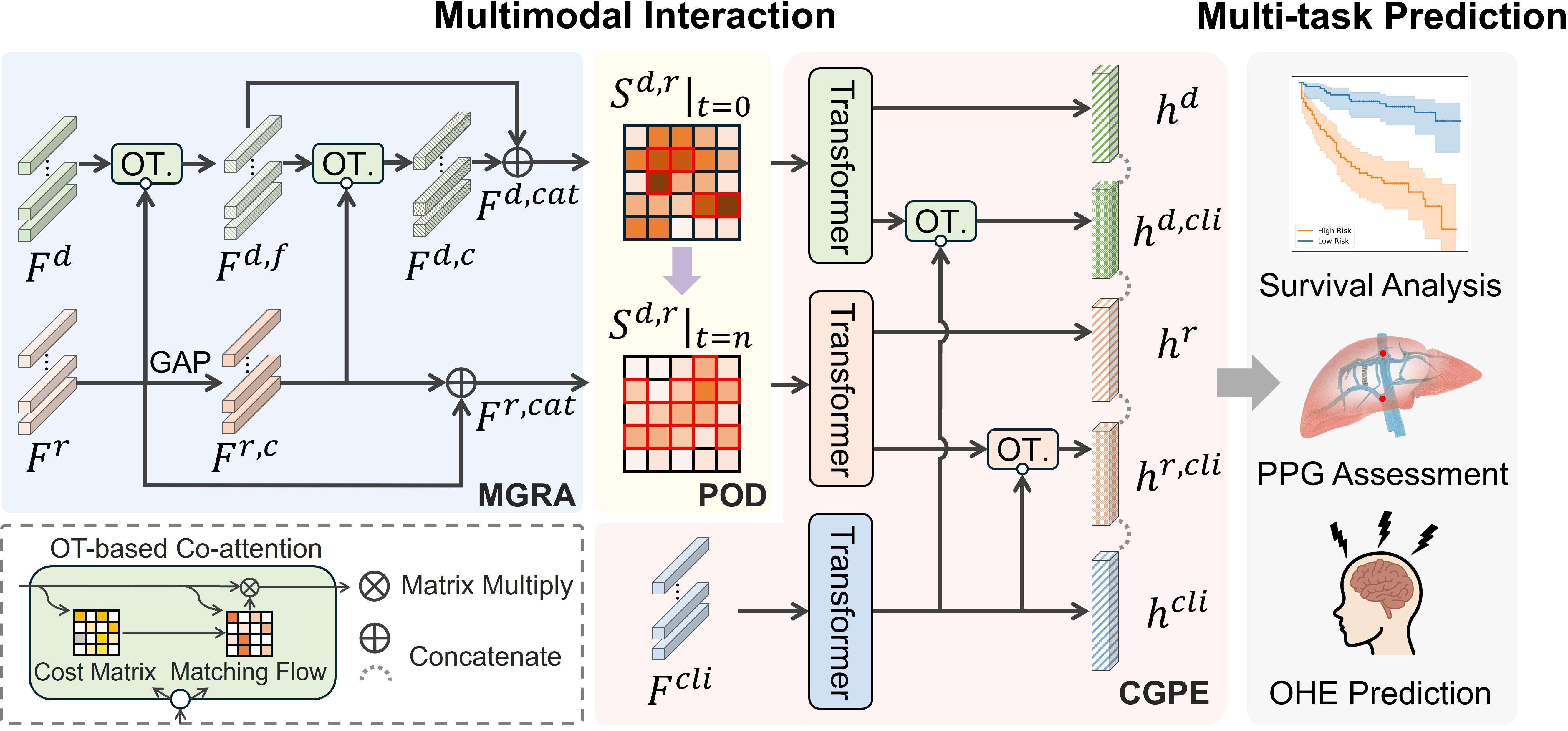}\caption{\label{fig:main_method}Overview of the proposed multimodal prediction architecture, which consists of two components: multimodal interaction and multi-task prediction. The former includes (1) a multi-grained radiomics attention (MGRA) mechanism to capture comprehensive deep learning features guided by hierarchical radiomics features; (2) a progressive orthogonal disentanglement (POD) strategy to reduce their redundancy and enhance complementary patterns; and (3) a clinically guided prognostic enhancement (CGPE) module to obtain a generalized and discriminative unified representation across the three modalities. Then, this representation serves three TIPS prognostic tasks involving survival analysis, PPG assessment and OHE prediction. OT.: OT-based Co-attention. GAP: Global Attention Pooling.
}
\end{figure*}

\subsection{Multimodal Interaction}
\label{sec:Multimodal Interaction}
We propose a novel multimodal interaction framework to fully leverage prognostic information across modalities and learn robust complementary representations, as illustrated in Fig.~\ref{fig:main_method}. This framework comprises three key modules: multi-grained radiomics attention (MGRA), progressive orthogonal disentanglement (POD), and clinically guided prognostic enhancement (CGPE). MGRA first promotes multi-grained interaction between deep learning and radiomics features to obtain comprehensive representations. To reduce shared redundancy in these representations and encourage cross-modal complementary patterns, POD is leveraged to progressively enforce orthogonality between the two feature streams. Furthermore, CGPE introduces prognostically discriminative clinical features to guide and enhance the two feature streams, ultimately yielding a trimodal discriminative representation for multi-task prediction.

\begin{figure*}[!t]
\centering
\includegraphics[width=0.6\paperwidth]{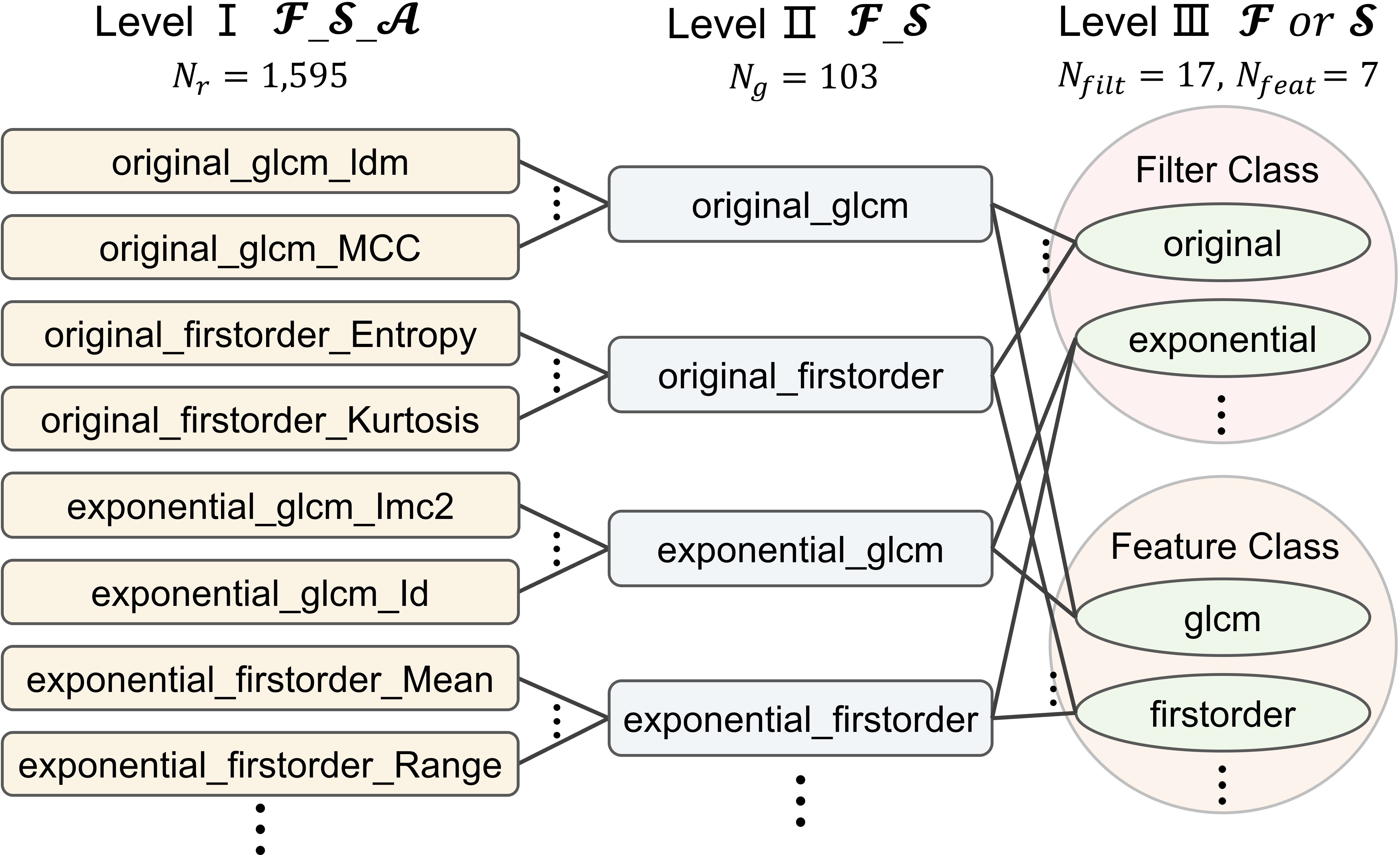}
\caption{Example of the hierarchical structure of radiomics features. $\mathcal{F}$: Filter Class. $\mathcal{S}$: Feature Class. $\mathcal{A}$: Attribute. \textbf{Level~\(\mathrm{I}\)} represents 1,595 individual $1\times1$ features, each defined by a specific ($\mathcal{F}$, $\mathcal{S}$, $\mathcal{A}$) combination. \textbf{Level~\(\mathrm{II}\)} groups features by joint filter-feature categories ($\mathcal{F}\_\mathcal{S}, N_g = 103$). \textbf{Level~\(\mathrm{III}\)} further aggregates Level~\(\mathrm{II}\) into higher-level categories based on either $\mathcal{F}$ ($N_{filt} = 17$) or $\mathcal{S}$ ($N_{feat} = 7$).}
\label{fig:radiomics_hierarchy}
\end{figure*}

\subsubsection{Multi-Grained Radiomics Attention (MGRA)}Given the CT-derived correlation between deep learning and radiomics features, the former offer complementary perspectives to handcrafted radiomics representations. To fully leverage this advantage, it is essential to learn discriminative deep representations through multimodal interactions, thereby improving prognostic modeling. However, there is currently no standard framework for modeling such interactions between radiomics and deep learning features. Some related studies on cancer survival analysis \cite{mcat, motcat} group genes, which are also represented as $1\times1$ attributes, into functional categories (typically $N=6$) to provide coarse-grained guidance for histology feature aggregation. In contrast, \cite{survpath} tokenizes genes according to established biological pathways ($N=331$) associated with specific cellular functions, enabling fine-grained semantic interactions with histology features. Nevertheless, both strategies suffer from granularity deficiency, as they model only coarse- or fine-level interactions, resulting in limited representations. Given this and the intrinsic hierarchy of radiomics features (as shown in Fig.~\ref{fig:radiomics_hierarchy}), we develop a multi-grained radiomics attention (MGRA) module to produce comprehensive deep learning features under multi-level radiomic guidance.

Specifically, to facilitate fine-grained guidance, we adopt the Level II partitioning scheme (Fig.~\ref{fig:radiomics_hierarchy}) to divide the radiomics features $F^{r, pre}$ into $N_g = 103$ groups based on joint filter-feature categories, formulated as $F^{r, pre} = \{f_j^{r, pre}\}_{j=1}^{N_r} = \{g_k^{r, pre}\}_{k=1}^{N_g}$
, where $g_k^{r, pre} \in \mathbb{R}^{d_k}$ denotes the $k$-th group with $d_k$-dimensional attributes. Unlike the sparse individual features in Level I that struggle to provide consistent guidance, and the coarse groupings in Level III that tend to obscure local details, Level II groups encourage both semantic consistency and fine-grained interactions with deep learning features. For example, in the “exponential\_glcm” group, exponential filtering enhances high-intensity regions, where texture descriptors such as fineness, complexity, and asymmetry are integrated into a unified pattern to facilitate co-expression with semantically related deep learning representations. Therefore, in this work, each feature group $g_k^{r, pre}$ is encoded into a radiomics embedding via a separate network 	$\phi_k^r(\cdot)$, as follows:
\begin{equation}\label{eq:F^r}
F^r = \left\{ \phi_k^r\left(g_k^{r, pre}\right) \right\}_{k=1}^{N_g} = \left\{ f_k^r \right\}_{k=1}^{N_g} \in \mathbb{R}^{N_g \times d},
\end{equation}
where $\phi_k^r(\cdot)$ consists of a two-layer self-normalizing neural network (SNN) \cite{snn} with an output dimension of $d = 256$.

Accordingly, each deep learning feature $f_s^{d, pre}$ is mapped to the token dimension $d$ using a shared fully connected layer $\phi^d(\cdot)$, yielding:
\begin{equation}\label{eq:F^d}
F^d = \left\{ \phi^d\left(f_s^{d, pre}\right) \right\}_{s=1}^{N_d} = \left\{ f_s^d \right\}_{s=1}^{N_d} \in \mathbb{R}^{N_d \times d}.
\end{equation}

Therefore, the fine-grained interactions are modeled between $F^d$ and $F^r$, as illustrated in Fig.~\ref{fig:main_method}. Following previous works \cite{motcat, mmp, icfnet}, we employ a widely used optimal transport (OT)-based co-attention module to learn the optimal matching flow between $F^d \in \mathbb{R}^{N_d \times d} \quad \text{and} \quad F^r \in \mathbb{R}^{N_g \times d}$, thereby guiding the aggregation of $F^d$. In particular, the pairwise cost matrix $C^{d, r} \in \mathbb{R}^{N_d \times N_g}
$ between the two feature sets is computed using the $l_2$ distance. Then, a discrete Kantorovich formulation \cite{kantorovich} is employed to derive the optimal matching flow $T^{d, r}$:
\begin{equation}\label{eq: OT}
\mathcal{W}(F^d, F^r) = \min_{T^{d,r} \in \Pi(\mu_d, \mu_r)} < T^{d,r}, C^{d,r} >_F,
\end{equation}
where $\mathcal{W}(F^d, F^r)$ denotes the optimal transport from $F^d$ to $F^r$, and $\Pi(\mu_d, \mu_r)$ $= \left\{ T^{d,r} \in \mathbb{R}_+^{N_d \times N_g} \ \middle| \ T^{d,r} \mathbf{1}_{N_g} = \mu_d,\ (T^{d,r})^\top \mathbf{1}_{N_d} = \mu_r \right\}
$ defines the marginal constraints that enforce total mass equality between the marginal distributions $\mu_d$ and $\mu_r$ of $F^d$ and $F^r$, respectively. Here, $\mathbf{1}_m$ represents an $m$-dimensional vector of ones. $<\cdot, \cdot>_F$ refers to the Frobenius inner product, and thus Eq.~(\ref{eq: OT}) aims to identify the optimal matching flow $T^{d,r}$ that minimizes the overall matching cost. Given the large size of $F^d$, we utilize the generalized Sinkhorn-Knopp \cite{Sinkhorn-Knopp1, Sinkhorn-Knopp2} algorithm  to estimate $T^{d,r}$, yielding $\hat{T}^{d,r}$, which is then used to guide the fine-grained aggregation of deep learning features, expressed as $F^{d,f} = (\hat{T}^{d,r})^\top F^d \in \mathbb{R}^{N_g \times d}
$. Notably, apart from reducing the dimensionality of $F^d$, the OT-based approach encourages the alignment of deep feature distribution with that of radiomics, thereby promoting correlated co-expression and generating novel semantic structures.

For coarse-level interactions, instead of aggregating the fine-grained radiomics features $F^r$ by filter or feature classes as in Level III, we apply the global attention pooling (GAP) \cite{attnmil} module to adaptively compute $N_c$ coarse-grained features $F^{r,c}$ across all embeddings:
\begin{equation}\label{eq: GAP_whole}
F^{r,c} = \mathrm{GAP}(F^r) \in \mathbb{R}^{N_c \times d}.
\end{equation}
Specifically, each $f_e^{r,c} \in F^{r,c}$ can be calculated as follows:
\begin{equation}\label{eq: GAP_explain}
\begin{aligned}
f_{e}^{r,c} &= \rho \left( \sum_{i=1}^{N_g} a_{e,i} \cdot f_i^r \right),~e = 1, \dots, N_c, \\
\text{where}~a_{e,i} &= \frac{
    \exp\left( \mathbf{w}_e^\top \left( \tanh(\mathbf{W}_a f_i^r) \odot \sigma(\mathbf{W}_b f_i^r) \right) \right)
}{
    \sum_{j=1}^{N_g} \exp\left( \mathbf{w}_e^\top \left( \tanh(\mathbf{W}_a f_j^r) \odot \sigma(\mathbf{W}_b f_j^r) \right) \right)
}.
\end{aligned}
\end{equation}
Here, $N_c$ is set to 6. $\mathbf{W}_a, \mathbf{W}_b\in\mathbb{R}^{d \times d}$ are learnable weight matrices, followed by the non-linear activations $\tanh$ and $\sigma$ (sigmoid), respectively. The element-wise multiplication is denoted by $\odot$, and $\mathbf{w}_e \in \mathbb{R}^d$ represents the $e$-th attention vector derived from $\mathbf{W}_c \in \mathbb{R}^{d \times N_c}$. Consequently, $	a_{e,i}$ indicates the contribution of the embedding $f_i^r$ to the $e$-th aggregated feature, which is then transformed by a fully connected layer $\rho(\cdot)$ mapping $\mathbb{R}^d$ to $\mathbb{R}^d$.

Note that $F^{r,c}$ integrates local details from $F^r$, resulting in a coarse-level global representation. To enhance co-expression at this level, it is used to guide the aggregation of $F^{d, f}$ through the OT-based module:
\begin{equation}\label{eq: F d,c}
F^{d,c} = \left( \hat{T}^{(d,f),(r,c)} \right)^\top F^{d,f},
\end{equation}
where $F^{d,c} \in \mathbb{R}^{N_c \times d}$ represents the coarse-grained deep learning features, and $\hat{T}^{(d,f),(r,c)}$ denotes the estimated optimal matching flow between $F^{d,f}$ and $F^{r,c}$.

Thus, comprehensive deep learning representations can be attained by concatenating the fine- and coarse-grained features $F^{d,f}$ and $F^{d,c}$, formulated as:
\begin{equation}\label{eq: F d,cat}
F^{d, cat} = \mathrm{Concat}(F^{d,f}, F^{d,c}),
\end{equation}
where $F^{d,cat} \in \mathbb{R}^{N_{cat} \times d}$ and $N_{cat} = N_g + N_c$.

Similarly, the multi-level radiomics features $F^{r,cat} \in \mathbb{R}^{N_{cat} \times d}
$ can be obtained as follows:
\begin{equation}\label{eq: F r,cat}
F^{r,cat} = \mathrm{Concat}(F^r, F^{r,c}).
\end{equation}

\subsubsection{Progressive Orthogonal Disentanglement (POD)}
Considering that the OT-based co-attention in MGRA often emphasizes common information via inter-modal distribution alignment, this suppresses complementary information and thereby degrades the discriminative performance of the generated features. To address this issue, we propose a progressive orthogonal disentanglement (POD) strategy (Fig.~\ref{fig:main_method}) that refines MGRA-derived representations by gradually enforcing orthogonality between deep learning and radiomics features, thus mitigating shared redundancy and enhancing complementary patterns.

Concretely, a pairwise similarity $S^{d,r} \in \mathbb{R}^{N_{cat} \times N_{cat}}
$ is first computed between $F^{d,cat}$ and $F^{r,cat}$, where each element $s_{mn}^{d,r} = \cos(f_m^{d,cat}, f_n^{r,cat})
$ denotes the cosine similarity between paired deep learning and radiomics features.

To ensure stable training, instead of directly enforcing orthogonality across all feature pairs, a novel dynamic masking strategy is presented that initially focuses on the most similar pairs in $S^{d,r}$ to eliminate dominant redundancy. As training proceeds and redundancy decreases, the mask progressively expands to cover the entire $S^{d,r}$, enabling smooth orthogonalization of all feature pairs. In particular, a dynamic ratio threshold $\gamma_t \in (0, 1]$ is defined to control the proportion of top similarity values subject to orthogonality constraints at iteration $t$: 
\begin{equation}\label{eq: γ_t_idea2}
\gamma_t = \exp\left[ -\frac{\max_0 - \mu_0}{\sigma_0} \cdot \frac{\widetilde{\max}_t}{\max_0} \cdot \left(1 - \frac{t}{T} \right)^\alpha \right],
\end{equation}
where the first term $\frac{\max_0 - \mu_0}{\sigma_0}$ serves as the initial parameter, and $\max_0$, $\mu_0$, and $\sigma_0$ denote the maximum, mean, and standard deviation of the similarity matrix $S^{d,r}$ at $t=0$, respectively. Thus, the initial parameter measures the deviation of the initial maximum similarity from the overall distribution. The second term $	\frac{\widetilde{\max}_t}{\max_0}$  represents the similarity scaling factor, indicating the relative change in maximum similarity from its initial value. To alleviate sharp fluctuations in $\max_t$, we use its exponential moving average (EMA), given by $\widetilde{\max}_t = \beta \cdot \widetilde{\max}_{t-1} + (1 - \beta) \cdot \max_t$, with $\beta = 0.99$. The third term $\left(1 - \frac{t}{T} \right)^\alpha$ is the temporal decay factor, designed to gradually extend the orthogonalization range over iterations. Here, $T$ denotes the maximum iteration index, and $\alpha$, the decay exponent, is set to 0.9 in our implementation.

Therefore, when $t=0$, the initial ratio $\gamma_0 = \exp\left(-\frac{\max_0 - \mu_0}{\sigma_0}\right)$ solely depends on the initial parameter. A larger value indicates a more prominent maximum similarity, resulting in a smaller $\gamma_0$, which guides the model to focus on the most similar pairs in $S^{d,r}
$. During early training, $\widetilde{\max}_t$ typically shows a significant decline due to the orthogonalization of highly similar pairs, making the increase in $\gamma_t$ dominated by the similarity scaling factor $\frac{\widetilde{\max}_t}{\max_0}$. As training advances, $\frac{\widetilde{\max}_t}{\max_0}$ gradually stabilizes, indicating the suppression of major redundancy. Hence, the rise of $\gamma_t$ is primarily driven by the temporal factor $\left(1 - \frac{t}{T}\right)^{\alpha}
$, facilitating smooth orthogonalization across all pairs. Notably, this dynamic thresholding strategy requires no extra hyper-parameters, only relying on similarities and training progress to adaptively reduce redundancy and encourage complementary patterns.

Thus, the orthogonality constraints can be implemented via the following loss function:
\begin{equation}\label{eq: L_ortho}
\mathcal{L}_{ortho} = \frac{1}{|M^{d,r}|} \sum_{m=1}^{N_{cat}} \sum_{n=1}^{N_{cat}} \left|s^{d,r}_{mn}\right| \cdot M^{d,r}_{mn},
\end{equation}
where $|M^{d,r}| = \sum_{m=1}^{N_{cat}} \sum_{n=1}^{N_{cat}} M^{d,r}_{mn}
$ denotes the number of elements contributing to the loss. Here, $M^{d,r} = \mathds{1}(S^{d,r} \ge \eta_t)$ is a binary valid mask, with $\eta_t$ being the similarity threshold corresponding to $\gamma_t$, i.e., $\eta_t = \texttt{np.percentile}(S^{d,r},\ 100 * (1 - \gamma_t))
$. $\left|s^{d,r}_{mn}\right|$ represents the absolute value of the cosine similarity, used to constrain the two features to be orthogonal.

\subsubsection{Clinically Guided Prognostic Enhancement (CGPE)}
Recent advances in multimodal learning have demonstrated that structured data, such as tabular clinical variables \cite{tab_cli} and genomic profiles \cite{motcat, ld-cvae}, can effectively guide visual feature aggregation and improve prognostic performance. Moreover, they tend to be more discriminative than visual features in prognostic modeling, as reflected in the superior performance of the corresponding unimodal models. This may be attributed to their explicit representation of biomedical factors relevant to prognosis, which are only implicitly encoded in images. Thus, given the established value of structured clinical data in post-TIPS prediction, a clinically guided prognostic enhancement (CGPE) module (as illustrated in Fig.~\ref{fig:main_method}) is introduced to guide deep learning and radiomics representations using clinical features, thereby generating more discriminative features to enhance prognostic effectiveness.

Specifically, following \cite{mcat, motcat, ld-cvae}, we first apply two transformer encoders, $\mathcal{T}^d(\cdot)$ and $\mathcal{T}^r(\cdot)$, to separately integrate multi-level features within the deep learning and radiomics modalities:
\begin{equation}\label{eq: H^d, H^r}
H^d = \mathcal{T}^d(F^{d,cat}), \quad H^r = \mathcal{T}^r(F^{r,cat}),
\end{equation}
where $H^d, H^r \in \mathbb{R}^{N_{cat} \times d}
$, sharing the same dimensions as $F^{d,cat}$ and $F^{r,cat}$.

Then, to generate clinical embeddings for guidance, the clinical features are divided into $N_{cli}=5$ groups based on the classification in Table~\ref{tab: Classification of clinical features}, denoted as $F^{\mathit{cli, pre}} = \{g_k^{\mathit{cli, pre}}\}_{k=1}^{N_{\mathit{cli}}}
$. Each feature group $g_k^{cli,pre}$ is transformed into a $d$-dimensional vector by a dedicated two-layer SNN $\phi_k^{\mathit{cli}}(\cdot)$ (not shown in Fig.~\ref{fig:main_method}), yielding $F^{\mathit{cli}} = \left\{ \phi_k^{\mathit{cli}}(g_k^{\mathit{cli, pre}}) \right\}_{k=1}^{N_{\mathit{cli}}} = \left\{ f_k^{\mathit{cli}} \right\}_{k=1}^{N_{\mathit{cli}}} \in \mathbb{R}^{N_{\mathit{cli}} \times d}
$. Similarly, $F^{cli}$ is subsequently fed into the transformer $\mathcal{T}^{cli}(\cdot)$ to extract intra-modal representations:
\begin{equation}\label{eq: H^cli}
H^{\mathit{cli}} = \mathcal{T}^{\mathit{cli}}(F^{\mathit{cli}}).
\end{equation}

Therefore, clinically guided co-attention can be performed between $H^{cli}$ and each of $H^d$ and $H^r$ to generate prognostically enhanced discriminative representations. We adopt the OT-based algorithm to implement this process:
\begin{equation}\label{eq: H^d, cli; H^r, cli}
H^{d,\mathit{cli}} = (\hat{T}^{d,\mathit{cli}})^{\top} H^d, \quad H^{r,\mathit{cli}} = (\hat{T}^{r,\mathit{cli}})^{\top} H^r,
\end{equation}
where $H^{d,\mathit{cli}},\ H^{r,\mathit{cli}} \in \mathbb{R}^{N_{\mathit{cli}} \times d}$ are the clinically guided aggregated features, and $\hat{T}^{d,\mathit{cli}}$, $\hat{T}^{r,\mathit{cli}}$ represent the optimal matching flows between $H^d$ and $H^{cli}$, and $H^r$ and $H^{cli}$, respectively.

Finally, to obtain generalized and discriminative representations across the three modalities, the global attention pooling (GAP) modules are separately applied to the intra-modal features $H^d$, $H^r$, $H^{cli}$ and the cross-modal features $H^{d,cli}$, $H^{r,cli}$ (GAP modules are omitted in Fig.~\ref{fig:main_method} for brevity) to produce their respective aggregated embeddings. The five embeddings are then concatenated into a unified multimodal representation shared across various TIPS prognostic tasks. The overall process can be formulated as follows:
\begin{equation}\label{eq: h^x, 5 path}
h^x = \mathrm{GAP}_x(H^x),\quad x \in \{d;\ d,\mathit{cli};\ r;\ r,\mathit{cli};\ \mathit{cli}\},
\end{equation}
then,
\begin{equation}\label{eq: h^final}
h^{\mathit{final}} = \mathrm{Concat}(h^d,\ h^{d,\mathit{cli}},\ h^r,\ h^{r,\mathit{cli}},\ h^{\mathit{cli}}).
\end{equation}

\subsection{Multi-Task Prediction}
\label{sec:Multi-Task Prediction}
The unified multimodal representation can be shared across three key TIPS prognostic tasks, i.e., survival analysis, PPG assessment, and OHE prediction (Fig.~\ref{fig:main_method}), for comprehensive postoperative evaluation. To ensure reliable multi-task learning, a staged training strategy is introduced, starting with generalized representation learning from the survival task, followed by the activation of task-specific features for PPG and OHE prediction.

\begin{figure*}[!t]
\centering
\includegraphics[width=0.47\paperwidth]{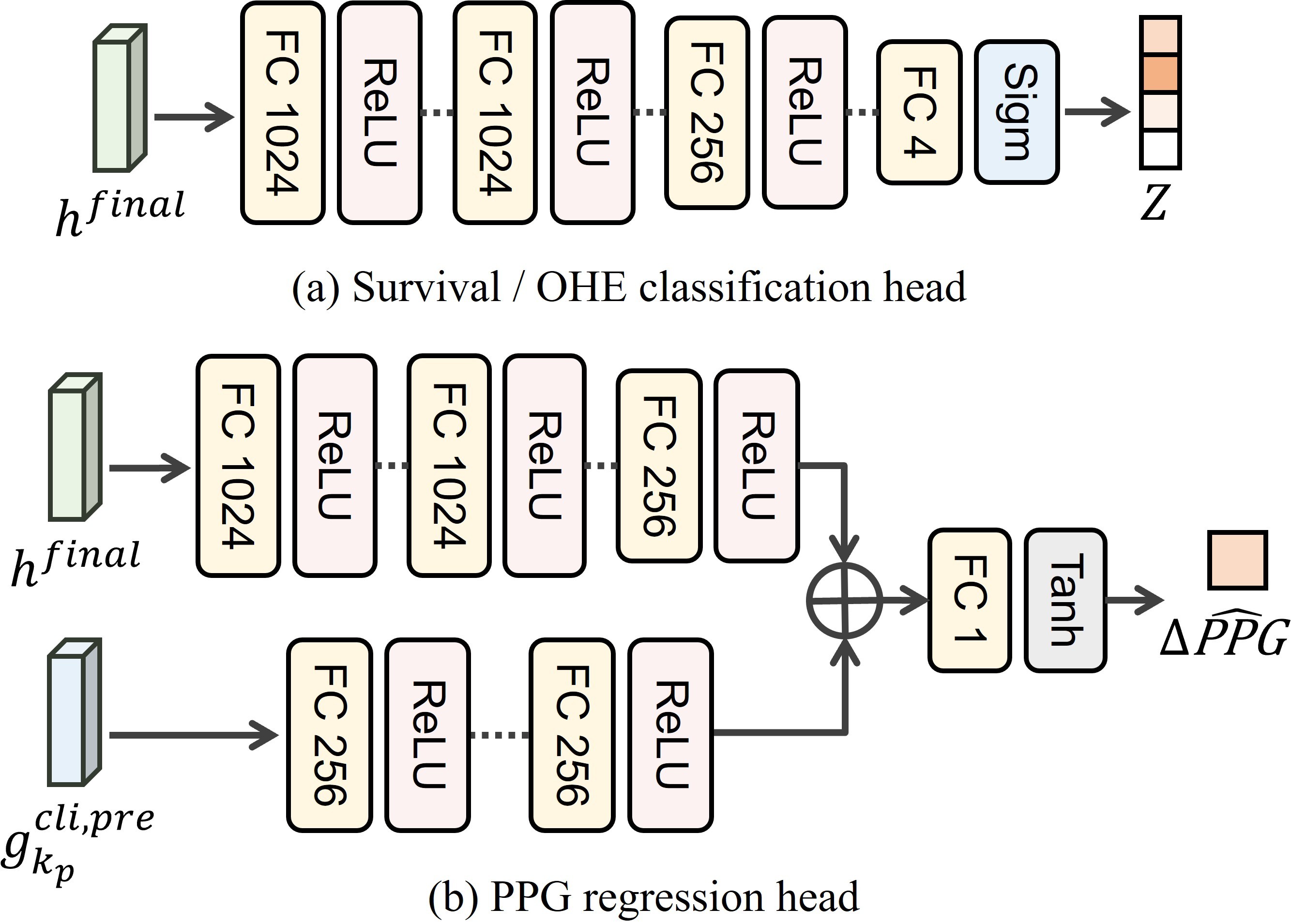}
\caption{Architecture of the multi-task prediction heads for (a) survival or OHE classification and (b) PPG regression. FC [num]: Fully connected layer with output dimension [num]. Sigm: Sigmoid activation function. $Z$: $Z^{surv}$ or $Z^{ohe}$. $\oplus$: Concatenation.
}
\label{fig:predict_head}
\end{figure*}

\subsubsection{Task Formulation}
\textbf{Survival Analysis.} This task aims to estimate the probability of an event (typically death) occurring before a specific time, where the event outcome is not always observed (i.e., right-censored data). Let $c \in \{0,1\}$ denote the censoring status, indicating whether the event is right-censored ($c=1$) or not ($c=0$), and $t^{os} \in \mathbb{R}^{+}$ the overall survival time (in months). Following previous works \cite{mcat, motcat}, we construct $n$ non-overlapping time intervals $[t_{i-1}, t_i)$ for $i \in \{1,2,\dots,n\}$, based on the quartiles of uncensored survival times in the training set. Under this setting, the task is transformed into a classification problem, with the class label $y^{os} \in \{1,2,\dots,n\}$ determined by the time interval in which $t^{os}$ is located. Then, several fully connected layers appended with a sigmoid function are utilized to build the survival classification head $\Phi^{surv}$, as shown in Fig.~\ref{fig:predict_head} (a), to predict the discrete hazard function $Z^{surv}$:
\begin{equation}\label{eq: Z^surv}
Z^{surv} = \Phi^{surv}\bigl(h^{final}\bigr),
\end{equation}
where $Z^{surv} = \{ z^{surv}_1, z^{surv}_2, \dots, z^{surv}_n \}$ serves as the predicted hazard vector, describing the probabilities of $t^{os}$ being located in each time interval. Here, $n$ is set to 4. Consequently, each patient is represented as a triplet $\{Z^{surv}, y^{os}, c\}$, with the cumulative survival function given by $f^{surv}\bigl(Z^{surv}, y^{os}\bigr)$ $= \prod_{j=1}^{y^{os}} \bigl( 1 - z^{surv}_{j} \bigr)$. Finally, the survival prediction objective can be formulated by the negative log-likelihood (NLL) loss \cite{nllloss}:
\begin{equation}\label{eq:Loss_surv}
\begin{aligned}
\mathcal{L}_{surv} = &-c \log \left( f^{surv}\bigl(Z^{surv}, y^{os}\bigr) \right) \\
&-(1 - c) \log \left( f^{surv}\bigl(Z^{surv}, y^{os} - 1\bigr) \right) \\
&-(1 - c) \log z^{surv}_{y^{os}}.
\end{aligned}
\end{equation}

\textbf{PPG Assessment.} Its goal is to accurately estimate the postoperative PPG value. Since post-TIPS PPG indicates changes in the pressure difference between the portal vein and the inferior vena cava, preoperative pressure measurements provide essential information for its prediction. Hence, apart from the inherent feature activation pathway for learning task-specific features from the shared representation $h^{final}$, an extra feature extraction subnet with pressure-related indicators as input is introduced to enhance their contribution to PPG prediction. In addition, given the availability of preoperative PPG, the post-TIPS PPG variation serves as the regression target, supporting an indirect assessment of the postoperative PPG value. The regression head $\Phi^{ppg}$ is depicted in Fig.~\ref{fig:predict_head} (b), and its prediction process can be formulated as follows:
\begin{equation}\label{eq: delta_PPG}
\Delta\widehat{PPG} = \Phi^{ppg}\left( h^{final}, g^{cli,pre}_{k_p} \right),
\end{equation}
where $\Delta\widehat{PPG}$ indicates the predicted PPG variation, and $g^{cli,pre}_{k_p}$ represents the input feature group of pressure-related indicators (as presented in Table~\ref{tab: Classification of clinical features}), with $k_p$ denoting its group index. Therefore, the PPG regression head can be optimized using the following loss function:
\begin{equation}\label{eq: Loss_ppg}
	\mathcal{L}_{ppg} = \left\lVert PPG_{post} - \left( PPG_{pre} + \Delta\widehat{PPG} \right) \right\rVert_2^2
,
\end{equation}
where $\left\lVert \cdot \right\rVert_2$ denotes the $l_2$ norm. $PPG_{post}$ and $PPG_{pre}$ refer to the true postoperative and preoperative PPG values, respectively.

\textbf{OHE Prediction.} The occurrence of postoperative OHE within a defined period is an important prognostic metric for TIPS. Thus, we aim to predict the probability of OHE occurring before a specified time. Considering the high similarity of targets and the presence of right-censoring, we adopt the same task formulation as in survival analysis, with the event of interest defined as the occurrence of OHE instead of death. Accordingly, the OHE prediction head $\Phi^{ohe}$ employs an identical structure to $\Phi^{surv}$ (Fig.~\ref{fig:predict_head} (a)), and its loss function $\mathcal{L}_{ohe}$ retains the form of Eq.~(\ref{eq:Loss_surv}), adapted with variables relevant to OHE.

\subsubsection{Training Strategy}
To ensure the effectiveness of the TIPS prognostic framework across all three tasks, a staged training strategy is adopted to avoid potential confusion from concurrent multi-task training. Particularly, the overall multimodal framework in Fig.~\ref{fig:main_method} is divided into two parts: a shared backbone $\Phi^{back}$ for generalized feature extraction and three prediction heads $\Phi^{surv}$, $\Phi^{ppg}$, and $\Phi^{ohe}$. In the first training stage, a pretraining process is conducted, where $\Phi^{back}$ and $\Phi^{surv}$ are jointly optimized using $\mathcal{L}_{surv}$ and $\mathcal{L}_{ortho}$, enabling the shared feature $h^{final}$ to capture generalized representation from the survival task. After pretraining, task-specific fine-tuning is performed with the backbone $\Phi^{back}$ frozen, sequentially optimizing $\Phi^{ppg}$ and $\Phi^{ohe}$ with $\mathcal{L}_{ppg}$ and $\mathcal{L}_{ohe}$, respectively, to achieve PPG and OHE estimation based on $h^{final}$. The staged training procedure is outlined in Algorithm \ref{alg:staged training strategy}.

\clearpage
\begin{breakablealgorithm}
\caption{Overview of the Staged Training Strategy}
\renewcommand{\algorithmicrequire}{\textbf{Input:}}
\renewcommand{\algorithmicensure}{\textbf{Output:}}
\label{alg:staged training strategy}
\begin{algorithmic}[1]
\REQUIRE Deep learning features $F^{d,pre}$, radiomics features $F^{r,pre}$, and clinical features $F^{cli, pre}$ at epoch $e$ and iteration $t$ (both starting from zero).
\ENSURE Optimized parameters of the model.
\renewcommand{\algorithmicensure}{\textbf{Initialization:}}
\ENSURE Number of epochs $E_0$, $E_1$, and $E_2$ for the pretraining, PPG estimation, and OHE prediction stages, respectively; maximum iteration index $T$ in the pretraining stage; weight $\delta$ of the orthogonality loss.

\renewcommand{\algorithmicensure}{\textbf{Process:}}

\ENSURE
\STATE \textcolor[rgb]{0.5,0.5,0.5}{// Stage I: pretraining (survival task)}
\IF{$e < E_0$}
    \STATE $h^{final} \leftarrow \Phi^{back}(F^{d, pre}, F^{r, pre}, F^{cli, pre})$
    \STATE $Z^{surv} \leftarrow \Phi^{surv}(h^{final})$
    \STATE Calculate the survival loss $\mathcal{L}_{surv}$ based on Eq.~(\ref{eq:Loss_surv}).
    \STATE Calculate the orthogonality loss $\mathcal{L}_{ortho}$ with $t$ and $T$ based on Eq.~(\ref{eq: L_ortho}).\STATE Calculate the overall loss $\mathcal{L} \leftarrow \mathcal{L}_{surv} + \delta \mathcal{L}_{ortho}$.
    \STATE Update the parameters of $\Phi^{back}$ and $\Phi^{surv}$ via backpropagation.
\STATE \hspace*{-1.25\algorithmicindent} \textcolor[rgb]{0.5,0.5,0.5}{// Stage II: task-specific fine-tuning}
\ELSE
    \STATE \textbf{if} $e < E_0 + E_1$ \textbf{then} \textcolor[rgb]{0.5,0.5,0.5}{// PPG estimation}
        \STATE \hspace{0.73em} Freeze the parameters of $\Phi^{back}$ and $\Phi^{surv}$.
        \STATE \hspace{0.73em} $h^{final} \leftarrow \Phi^{back}(F^{d, pre}, F^{r, pre}, F^{cli, pre})$
        \STATE \hspace{0.73em} \parbox[t]{0.95\linewidth}{$\Delta\widehat{PPG} \leftarrow \Phi^{ppg}(h^{final}, g_{k_p}^{cli, pre})$ \textcolor[rgb]{0.5,0.5,0.5}{// $g_{k_p}^{cli, pre}$ derived from $F^{cli, pre}$}}
        \vspace{0.01em}
        \STATE \hspace{0.73em} \parbox[t] {0.95\linewidth} {Calculate the PPG prediction loss $\mathcal{L}_{ppg}$ based on Eq.~(\ref{eq: Loss_ppg}).}
        \vspace{0.001em}
        \STATE \hspace{0.73em} Update the parameters of $\Phi^{ppg}$ via backpropagation.
     \STATE \textbf{else} \textcolor[rgb]{0.5,0.5,0.5}{// $e < E_0 + E_1 + E_2$, OHE prediction}
        \STATE \hspace{0.73em} Freeze the parameters of $\Phi^{back}$, $\Phi^{surv}$ and $\Phi^{ppg}$.
        \STATE \hspace{0.73em} $h^{final} \leftarrow \Phi^{back}(F^{d, pre}, F^{r, pre}, F^{cli, pre})$
        \STATE \hspace{0.73em} $Z^{ohe} \leftarrow \Phi^{ohe}(h^{final})$
        \STATE \hspace{0.73em} Calculate the OHE prediction loss $\mathcal{L}_{ohe}$.
        \STATE \hspace{0.73em} Update the parameters of $\Phi^{ohe}$ via backpropagation.
    \STATE \textbf{end if}
\ENDIF
\end{algorithmic}
\end{breakablealgorithm}

\section{Experiments}
\subsection{Experimental Setup}
\subsubsection{Dataset and Settings}
As described in Section~\ref{sec: data usage}, the internal dataset of 306 patients, including CT images and clinical characteristics, is used to develop and validate the proposed approach. Within this cohort, voxel-level portal vein annotations are available for only 32 patients ($\sim\!10\%$), leaving the remaining 274 unlabeled. The experimental settings for each task are summarized as follows:
\begin{itemize}
\item The segmentation task is performed solely on patient imaging data. For the semi-supervised pipeline, the 32 labeled patient samples are split into a labeled training set ($n=22$) and a validation set ($n=10$), while the 274 unlabeled samples are partitioned using a five-fold cross-validation scheme. Specifically, we evenly divide them into five subsets; in each cross-validation round, four subsets serve as the unlabeled training set, and the remaining one as the test set for inference. This ensures that each unlabeled sample is segmented once as a test case over five rounds, yielding its predicted labels. For the foundation model-based pipeline, the same labeled training and validation sets are leveraged to train and evaluate the model, and inference is performed on the entire unlabeled set with weakly supervised bounding-box prompts.
\item For the post-TIPS prediction task, since complete multimodal features can be obtained for each patient (through inherent or generated portal vein labels), we conduct five-fold cross-validation on the full internal dataset ($n=306$) to demonstrate the effectiveness of the proposed method.
\end{itemize}

Additionally, the external test set of 137 cases is employed to assess the robustness and generalization of the overall framework.

\subsubsection{Implementation Details}
Given the varying tasks and approaches, the implementation details are organized into the following three parts: 

For semi-supervised segmentation, we adopt DeepLabV3+ \cite{deeplabv3+} as the segmentation network with ResNet-101 \cite{resnet} as its backbone. Due to the limited size of the labeled training set, the network is first pretrained in a fully supervised manner on portal vein annotations from the public 3D-IRCADB \cite{3d-ircadb} dataset to enhance its robustness. For semi-supervised training, 3D CT volumes are axially sliced into 2D images, with each image truncated to [-50, 150] Hounsfield units (HU) and rescaled to [0, 1]. Weak data augmentations are applied to both labeled and unlabeled images, including random resizing within [0.5, 2.0], horizontal flipping with a probability of 0.5, and random cropping to $512\times512$, while strong augmentations are only performed on unlabeled data, consisting of contrast enhancement. For training strategies, we employ the stochastic gradient descent (SGD) optimizer with an initial learning rate of 0.001, momentum of 0.9, and weight decay of 1e-4. The batch size is $B_l=B_u=16$, and the model is trained for 200 epochs using a polynomial learning rate scheduler: $lr = lr_{init} \cdot \left(1 - \frac{iter}{max\_iter} \right)^{0.9}$. The weight parameter $\alpha$ in Eq.~(\ref{eq:E(·)}) is set to $2/3$ to balance the foreground and background losses, and the confidence threshold $E(\tau)$ increases linearly from 0.8 to 0.95 during training to match the growing confidence of pseudo-labels. Additionally, the teacher network is used for validation and inference to produce stable predictions. The experiments are conducted on four NVIDIA Tesla V100 GPUs with mixed precision training \cite{mixed} to overcome memory constraints.

For foundation model-based segmentation, we follow the configuration of MedSAM2 \cite{medsam2_ours}, employing the pretrained SAM2-Tiny \cite{sam2} model with the smallest Hiera \cite{hiera} image encoder. The preprocessing of CT scans involves intensity clipping to [-50, 150] HU followed by rescaling to [0, 255], splitting into 2D slices, resizing to $1024\times1024$, and applying z-score normalization. In the fine-tuning stage, the model is trained for 100 epochs using the AdamW optimizer on a single NVIDIA Tesla V100 GPU, with a batch size $B_l=8$, a learning rate of 3e-5, beta parameters of (0.9, 0.999), and a weight decay of 0.01. During inference, the required bounding box prompts are automatically generated by computing the minimum bounding rectangles for each ROI in the ground-truth masks. Furthermore, a bounding box perturbation strategy is applied in both stages by extending each side by up to 5 pixels at the original image scale to simulate deviations in manual annotations.

For post-TIPS prediction, categorical variables in radiomics and clinical features are processed via one-hot encoding, while numerical attributes are normalized to the [0, 1] range. During training, the Adam optimizer is utilized with a learning rate of 2e-4 and weight decay of 1e-5. We set the batch size to 1 and the number of epochs for each stage to $E_0=E_1=E_2=20$. The orthogonality loss uses a weight of $\delta=0.1$, and the generalized Sinkhorn-Knopp \cite{Sinkhorn-Knopp1, Sinkhorn-Knopp2} algorithm is implemented following \cite{motcat}. All the experiments are conducted on a single NVIDIA Tesla V100 GPU.

\subsubsection{Evaluation Metrics}
To quantitatively evaluate the segmentation performance, four widely used metrics are employed, including Dice, Jaccard, the 95\% Hausdorff distance (95HD), and the average surface distance (ASD). For TIPS prognosis, we apply the concordance index (C-index) and mean Brier score (mBS) to assess the survival and OHE prediction tasks \cite{brier_score}. Taking survival prediction as an example, the C-index quantifies the proportion of all comparable patient pairs whose predicted risk scores (the negative sum of cumulative survival functions) are consistent with their observed survival outcomes, indicating the model’s discriminative capacity in risk ordering. The mBS measures the absolute accuracy of predicted survival probabilities across various patient categories, defined as:
\begin{equation}\label{eq:mBS}
\mathrm{mBS} = \frac{1}{|\mathcal{C}|} \sum_{(y,c) \in \mathcal{C}} 
\frac{1}{|\mathcal{I}_{(y,c)}|} \sum_{i \in \mathcal{I}_{(y,c)}}\mathrm{BS}_i,
\end{equation}
where $\mathrm{BS}_i$ denotes the average squared distance between the predicted survival probabilities and the observed survival status for patient $i$. $(y,c)$ represents a combination of the patient label and censoring status; $\mathcal{I}_{(y,c)}$ comprises patients in category $(y,c)$, while $\mathcal{C}$ is the set of all such categories. Furthermore, Kaplan-Meier (KM) analysis \cite{KM} and the log-rank test \cite{logrank} are conducted to evaluate the risk stratification ability among different methods. For the estimation of PPG, mean absolute error (MAE) and root mean squared error (RMSE) serve as the quantitative measures.

\begin{table}[!t]
\scriptsize
\caption{Quantitative comparisons of the SOTA methods under two pipelines on the MultiTIPS validation set. ``Semi.'' and ``Found.'' denote the semi-supervised and foundation model-based pipelines, respectively. ``Lab.'' and ``Unlab.'' denote the use of labeled or unlabeled data for training, while ``Box'' refers to inference with bounding box prompts. $\uparrow$ indicates that higher values are better and $\downarrow$ indicates that lower values are better. $\dagger$ denotes the use of dual-stream strong perturbations only.
$\mathcal{F}$ indicates a fine-tuned model. $\ddagger$ represents the adoption of 3D inference. The best results are in \textbf{bold}, and the second-best are \underline{underlined}.}
\label{table: segmentation}
\centering
\setlength{\tabcolsep}{1pt}
\begin{tabular}{>{\centering\arraybackslash}m{40pt} 
                >{\centering\arraybackslash}m{83pt} 
                >{\centering\arraybackslash}m{25pt}
                >{\centering\arraybackslash}m{25pt}
                >{\centering\arraybackslash}m{25pt}
                >{\centering\arraybackslash}m{52pt}
                >{\centering\arraybackslash}m{52pt}
                >{\centering\arraybackslash}m{52pt}
                >{\centering\arraybackslash}m{52pt}}
\toprule[1pt]
\multirow{2}{*}[-0.3em]{Pipeline} & \multirow{2}{*}[-0.3em]{Method} & \multicolumn{3}{c}{Data/Prompt used} & \multicolumn{4}{c}{Metric} \\
\cmidrule(lr){3-5} \cmidrule(lr){6-9}
& & Lab. & Unlab. & Box & Dice(\%)$\uparrow$ & Jaccard(\%)$\uparrow$ & 95HD(Voxel)$\downarrow$ & ASD(Voxel)$\downarrow$ \\
\midrule
 / & SupOnly & \ding{51} & \ding{55} & \ding{55} & 71.41 & 55.99 & 11.30 & 2.78 \\
\midrule

\multirow{6}{*}[0em]{Semi.} & FixMatch \cite{fixmatch} & \ding{51} & \ding{51} & \ding{55} & 72.77\textsubscript{{$\pm$0.75}} & 57.93\textsubscript{{$\pm$0.76}} & 15.28\textsubscript{{$\pm$5.36}} & 1.76\textsubscript{{$\pm$1.08}} \\

& Mean Teacher \cite{mean_teacher} & \ding{51} & \ding{51} & \ding{55} & 76.43\textsubscript{{$\pm$0.62}} & 62.36\textsubscript{{$\pm$0.76}} & 13.56\textsubscript{{$\pm$3.45}} & 1.92\textsubscript{{$\pm$0.57}} \\

& UniMatch\textsuperscript{$\dagger$} \cite{Unimatch} & \ding{51} & \ding{51} & \ding{55} & 75.11\textsubscript{{$\pm$1.05}} & 60.83\textsubscript{{$\pm$1.20}} & 13.91\textsubscript{{$\pm$2.89}} & 2.18\textsubscript{{$\pm$0.63}} \\

& CorrMatch \cite{corrmatch} & \ding{51} & \ding{51} & \ding{55} & 72.37\textsubscript{{$\pm$0.43}} & 57.34\textsubscript{{$\pm$0.55}} & 23.31\textsubscript{{$\pm$4.55}} & 4.56\textsubscript{{$\pm$1.22}} \\

& DDFP \cite{ddfp} & \ding{51} & \ding{51} & \ding{55} & \underline{78.54\textsubscript{{$\pm$0.80}}} & \underline{65.05\textsubscript{{$\pm$1.05}}} & \underline{9.31\textsubscript{{$\pm$0.54}}} & \underline{1.25\textsubscript{{$\pm$0.33}}} \\

& DSDA-MT (Ours) & \ding{51} & \ding{51} & \ding{55} & \textbf{79.53}\textsubscript{{$\pm$0.56}} & \textbf{66.42}\textsubscript{{$\pm$0.68}} & \textbf{8.96}\textsubscript{{$\pm$1.35}} & \textbf{1.16}\textsubscript{{$\pm$0.33}} \\

\midrule

\multirow{7}{*}[0em]{\makecell[c]{Found.}} & SAM \cite{sam} & \ding{55} & \ding{55} & \ding{51} & 81.69 & 69.40 & 2.90 & 0.82 \\

& MedSAM \cite{medsam} & \ding{55} & \ding{55} & \ding{51} & 74.25 & 59.77 & 7.23 & 1.73 \\

& SAM2 \cite{sam2} & \ding{55} & \ding{55} & \ding{51} & 83.80 & 72.41 & 3.92 & 1.07 \\

& SAM\textsubscript{$\mathcal{F}$} \cite{sam} & \ding{51} & \ding{55} & \ding{51} & \underline{88.82} & \underline{79.98} & \underline{1.48} & \underline{0.33} \\

& MedSAM\textsubscript{$\mathcal{F}$} \cite{medsam} & \ding{51} & \ding{55} & \ding{51} & 84.16 & 72.88 & 2.70 & 0.70 \\

& MedSAM2 (Ours) \cite{medsam2_ours} & \ding{51} & \ding{55} & \ding{51} & \textbf{91.32} & \textbf{84.07} & \textbf{1.00} & \textbf{0.20} \\

& \graytext{MedSAM2\textsuperscript{$\ddagger$} \cite{medsam2_ours}} & \graytext{\ding{51}} & \graytext{\ding{55}} & \graytext{One} & \graytext{35.52} & \graytext{22.71} & \graytext{26.49} & \graytext{8.35} \\

\bottomrule[1pt]

\end{tabular}
\end{table}

\normalsize

\subsection{Portal Vein Segmentation Results}
The proposed methods under two segmentation pipelines are assessed on the MultiTIPS validation set and compared with representative SOTA approaches. For semi-supervised segmentation, we include FixMatch \cite{fixmatch}, Mean Teacher \cite{mean_teacher}, UniMatch \cite{Unimatch}, CorrMatch \cite{corrmatch}, and DDFP \cite{ddfp} in comparison. The results for all methods are reported as mean $\pm$ standard deviation, computed from five models trained via five-fold cross-validation on the unlabeled data. In addition, a supervised baseline trained solely on labeled data (denoted as SupOnly) is introduced to demonstrate the efficacy of leveraging unlabeled data in unsupervised algorithms. To ensure fair comparisons, all the methods are implemented using the DeepLabV3+ network with a ResNet-101 backbone and share the same supervised pretraining on the 3D-IRCADB dataset. For the foundation model-based pipeline, we evaluate SAM \cite{sam}, MedSAM \cite{medsam}, and SAM2 \cite{sam2} on portal vein segmentation, considering both their pretrained and fine-tuned versions (the fine-tuned SAM2 is denoted as MedSAM2 \cite{medsam2_ours}). Their performance on MultiTIPS is obtained based on the released codebases from \cite{medsam} and \cite{medsam2_ours}, with the default network configurations. Table~\ref{table: segmentation} shows the quantitative results of SOTA approaches under the two pipelines.

\begin{figure*}[!t]
	\centering
	\includegraphics[width=0.635\paperwidth]{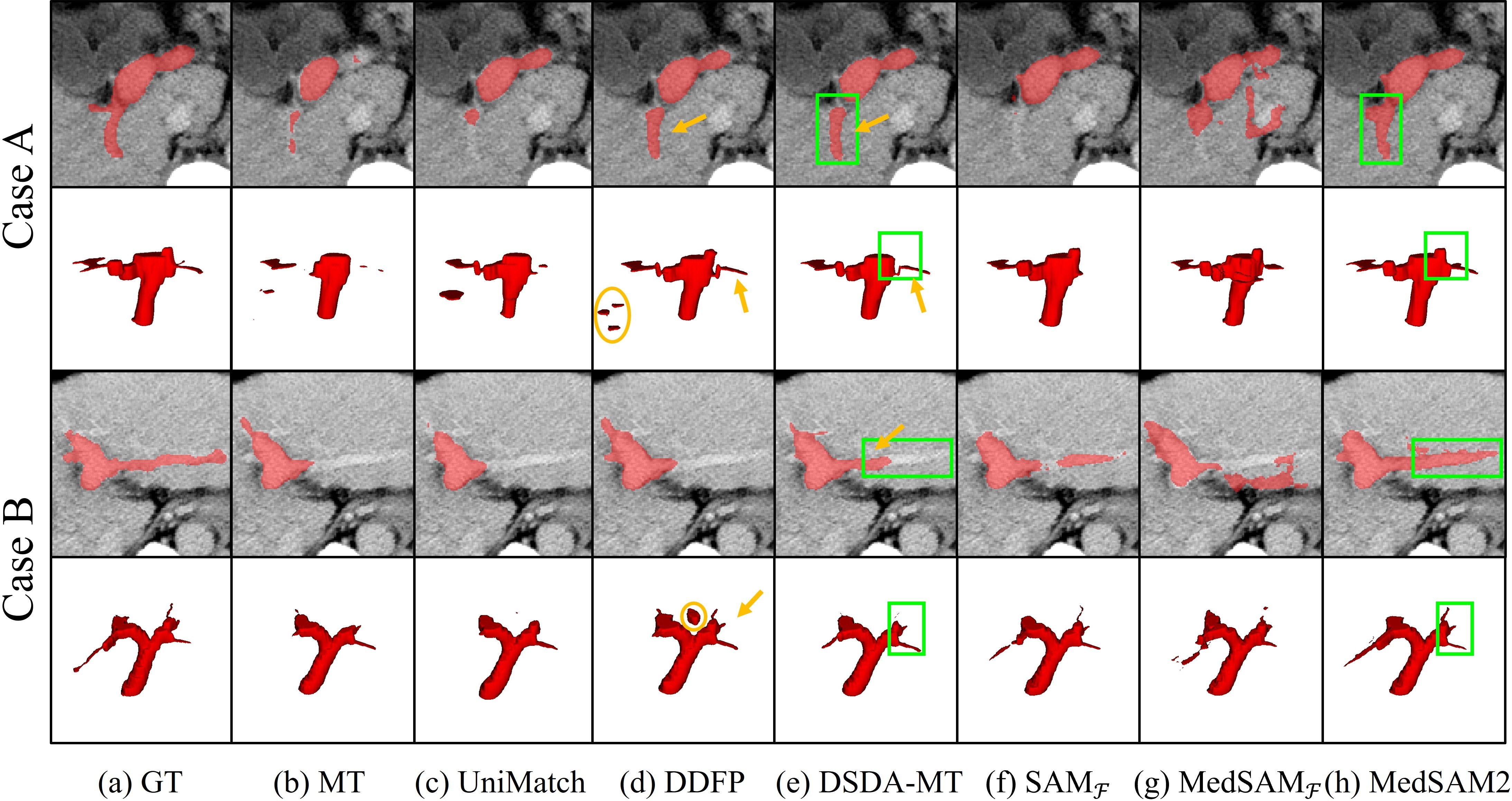}\caption{\label{fig:segment}Visual comparisons of different methods on the MultiTIPS validation set in 2D axial and 3D views. Yellow arrows in columns (d) and (e) indicate effective fine-branch segmentation, yellow ellipses in column (d) highlight false positive (FP) regions, and green rectangles in columns (e) and (h) mark comparisons of segmentation details. GT: Ground Truth. MT: Mean Teacher.
}
\end{figure*}

In the semi-supervised setting, the proposed DSDA-MT substantially surpasses the supervised baseline by 8.12\% in Dice, 10.43\% in Jaccard, 2.34 in 95HD, and 1.62 in ASD, demonstrating its ability to leverage unlabeled data for enhanced segmentation. When compared with its constituent baselines, Mean Teacher and UniMatch, our method exhibits consistent improvements across all metrics, even outperforming the previous best model, DDFP \cite{ddfp}. This confirms the effectiveness and superiority of integrating their complementary strengths in portal vein segmentation. For the foundation model-based setting, models fine-tuned on MultiTIPS (SAM\textsubscript{$\mathcal{F}$}, $\text{MedSAM}_\mathcal{F}$, and MedSAM2) all achieve better performance than their pretrained counterparts (SAM, MedSAM, and SAM2). Furthermore, MedSAM2 consistently outperforms all competing methods across the four metrics, achieving remarkable 91.32\% in Dice, 84.07\% in Jaccard, 1.00 in 95HD, and 0.20 in ASD, which can be attributed to the Hiera \cite{hiera} encoder’s multi-scale feature extraction for detailed high-resolution segmentation and large-scale image and video pretraining. Note that it also yields superior results to the semi-supervised approach DSDA-MT, leading by 11.79\% in Dice and 7.96 in 95HD, but requiring an additional bounding box prompt for each ROI during inference. Based on the configurations in \cite{medsam2_ours}, we also present the 3D inference performance of MedSAM2 (gray in Table~\ref{table: segmentation}), where the bounding box prompt is initialized only on the middle slice, followed by slice-wise propagation. However, poor segmentation results are observed, mainly due to significant inter-slice changes in ROI position, size, and count, which hinder effective bounding box propagation.

Fig.~\ref{fig:segment} illustrates the segmentation results of representative approaches in 2D axial and 3D views. For the semi-supervised approaches (columns b--e), both DDFP and DSDA-MT effectively capture challenging fine branches (indicated by yellow arrows in columns d and e), but DDFP tends to produce isolated false positive (FP) regions (highlighted by yellow ellipses in column d). In contrast, our DSDA-MT retains fine-branch segmentation accuracy while suppressing such FP regions, further validating that the Mean Teacher algorithm and the dual-stream perturbations in UniMatch enhance its stability and robustness. For the foundation model-based methods (columns f--h), the segmentation results of MedSAM2 are clearly closer to the ground truth than those of the other two methods. Moreover, compared with the semi-supervised method DSDA-MT, it achieves better completeness in segmentation details (cf. green rectangles in columns e and h), which may contribute to improved prognostic effectiveness. The robust performance of DSDA-MT and MedSAM2 on the validation set indicates their generalization to unlabeled data, thus supporting reliable multimodal prognosis.

% \clearpage
\begin{scriptsize}
\setlength{\tabcolsep}{1pt}
\begin{longtable}{>{\centering\arraybackslash}m{80pt} 
                >{\centering\arraybackslash}m{24pt}
                >{\centering\arraybackslash}m{24pt}
                >{\centering\arraybackslash}m{24pt}
                >{\centering\arraybackslash}m{55pt}
                >{\centering\arraybackslash}m{55pt}
                >{\centering\arraybackslash}m{55pt}
                >{\centering\arraybackslash}m{55pt}}
\caption{Comparative results of survival analysis (mean $\pm$ standard deviation) on the internal dataset. ``DL'', ``RA'' and ``Cli.'' represent deep learning, radiomics and clinical features, respectively. ``Metric\textsubscript{DSDA-MT}'' and ``Metric\textsubscript{MedSAM2}'' denote survival prediction metrics based on segmentation labels generated by DSDA-MT and MedSAM2 \cite{medsam2_ours}, respectively. $*$ indicates identical values for Metric\textsubscript{DSDA-MT} and Metric\textsubscript{MedSAM2}, since only clinical data are used. The best results are in \textbf{bold}, and the second-best are \underline{underlined}.}
\label{tab: survial compare SOTA}\\
% \centering
\toprule[1pt]
\multirow{2}{*}[-0.2em]{Method}  & \multicolumn{3}{c}{Modality} & \multicolumn{2}{c}{Metric\textsubscript{DSDA-MT}} & \multicolumn{2}{c}{Metric\textsubscript{MedSAM2}} \\ 
\cmidrule(l{0.25em}r{0.25em}){2-4} \cmidrule(l{0.25em}r{0.25em}){5-6} \cmidrule(l{0.25em}r{0.25em}){7-8}
% & & & & & & &
 & DL & RA & Cli. & C-index$\uparrow$ & mBS$\downarrow$ & C-index$\uparrow$ & mBS$\downarrow$ \\
\midrule
\endfirsthead
 % —— Continued page header ——
\toprule[1pt]
\multirow{2}{*}[-0.2em]{Method}  
& \multicolumn{3}{c}{Modality} 
& \multicolumn{2}{c}{Metric\textsubscript{DSDA-MT}} 
& \multicolumn{2}{c}{Metric\textsubscript{MedSAM2}} \\ 
\cmidrule(l{0.25em}r{0.25em}){2-4} 
\cmidrule(l{0.25em}r{0.25em}){5-6} 
\cmidrule(l{0.25em}r{0.25em}){7-8}
& DL & RA & Cli. & C-index$\uparrow$ & mBS$\downarrow$ & C-index$\uparrow$ & mBS$\downarrow$ \\
\midrule
\endhead

% —— Page footer except the last page ——
\midrule
\endfoot

% —— Last page footer ——
\endlastfoot

$\text{SNN}^{*}$ \cite{snn} & ~ & ~ & \ding{51} & 0.6413\textsubscript{{$\pm$0.0635}} & 0.2093\textsubscript{{$\pm$0.0492}} & 0.6413\textsubscript{{$\pm$0.0635}} & 0.2093\textsubscript{{$\pm$0.0492}} \\

$\text{SNNTrans}^{*}$ \cite{snn, transmil} & ~ & ~ & \ding{51} & 0.6471\textsubscript{{$\pm$0.0530}} & 0.1689\textsubscript{{$\pm$0.0181}} & 0.6471\textsubscript{{$\pm$0.0530}} & 0.1689\textsubscript{{$\pm$0.0181}} \\

$\text{S-MLP}^{*}$ \cite{s-mlp} & ~ & ~ & \ding{51} & 0.6361\textsubscript{{$\pm$0.0392}} & 0.1633\textsubscript{{$\pm$0.0222}} & 0.6361\textsubscript{{$\pm$0.0392}} & 0.1633\textsubscript{{$\pm$0.0222}} \\

\midrule

Deep Sets \cite{deepsets} & \ding{51} & ~ & ~ & 0.4963\textsubscript{{$\pm$0.0045}} & 0.3359\textsubscript{{$\pm$0.0284}} & 0.5020\textsubscript{{$\pm$0.0039}} & 0.3080\textsubscript{{$\pm$0.0992}} \\

AttnMIL \cite{attnmil} & \ding{51} & ~ & ~ & 0.5754\textsubscript{{$\pm$0.0899}} & 0.1665\textsubscript{{$\pm$0.0109}} & 0.5866\textsubscript{{$\pm$0.0749}} & 0.1642\textsubscript{{$\pm$0.0121}} \\

DeepAttnMISL \cite{deepattnmisl} & \ding{51} & ~ & ~ & 0.5976\textsubscript{{$\pm$0.0546}} & 0.1639\textsubscript{{$\pm$0.0073}} & 0.6007\textsubscript{{$\pm$0.0822}} & 0.1644\textsubscript{{$\pm$0.0181}} \\

CLAM-SB \cite{clam} & \ding{51} & ~ & ~ & 0.5942\textsubscript{{$\pm$0.0475}} & 0.1664\textsubscript{{$\pm$0.0104}} & 0.6196\textsubscript{{$\pm$0.0527}} & 0.1635\textsubscript{{$\pm$0.0111}} \\

CLAM-MB \cite{clam} & \ding{51} & ~ & ~ & 0.5998\textsubscript{{$\pm$0.0399}} & 0.1680\textsubscript{{$\pm$0.0103}} & 0.6326\textsubscript{{$\pm$0.0348}} & 0.1638\textsubscript{{$\pm$0.0081}} \\

TransMIL \cite{transmil} & \ding{51} & ~ & ~ & 0.5881\textsubscript{{$\pm$0.0928}} & 0.1650\textsubscript{{$\pm$0.0212}} & 0.6011\textsubscript{{$\pm$0.0824}} & 0.1627\textsubscript{{$\pm$0.0087}} \\

DTFD-MIL \cite{dtfd} & \ding{51} & ~ & ~ & 0.5868\textsubscript{{$\pm$0.0647}} & 0.1624\textsubscript{{$\pm$0.0115}} & 0.5959\textsubscript{{$\pm$0.0658}} & 0.1593\textsubscript{{$\pm$0.0126}} \\

ILRA \cite{ilra} & \ding{51} & ~ & ~ & 0.5880\textsubscript{{$\pm$0.0883}} & 0.1610\textsubscript{{$\pm$0.0108}} & 0.5938\textsubscript{{$\pm$0.0907}} & 0.1623\textsubscript{{$\pm$0.0101}} \\

WiKG \cite{wikg} & \ding{51} & ~ & ~ & 0.6317\textsubscript{{$\pm$0.0230}} & 0.1621\textsubscript{{$\pm$0.0081}} & 0.6345\textsubscript{{$\pm$0.0492}} & 0.1638\textsubscript{{$\pm$0.0070}} \\

OTSurv \cite{otsurv} & \ding{51} & ~ & ~ & 0.6020\textsubscript{{$\pm$0.1013}} & 0.1644\textsubscript{{$\pm$0.0110}} & 0.6235\textsubscript{{$\pm$0.0643}} & 0.1606\textsubscript{{$\pm$0.0081}} \\

\midrule

MCAT \cite{mcat} & \ding{51} & ~ & \ding{51} & 0.6567\textsubscript{{$\pm$0.0782}} & 0.1758\textsubscript{{$\pm$0.0256}} & 0.6710\textsubscript{{$\pm$0.0668}} & 0.1690\textsubscript{{$\pm$0.0167}} \\

Pathomic \cite{pathomic} & \ding{51} & ~ & \ding{51} & 0.5902\textsubscript{{$\pm$0.0729}} & \textbf{0.1590}\textsubscript{{$\pm$0.0111}} & 0.5953\textsubscript{{$\pm$0.0667}} & 0.1573\textsubscript{{$\pm$0.0104}} \\

Porpoise \cite{porpoise} & \ding{51} & ~ & \ding{51} & 0.5879\textsubscript{{$\pm$0.0669}} & \underline{0.1604\textsubscript{{$\pm$0.0188}}} & 0.5826\textsubscript{{$\pm$0.0763}} & \underline{0.1566\textsubscript{{$\pm$0.0141}}} \\

MOTCat \cite{motcat} & \ding{51} & ~ & \ding{51} & 0.6656\textsubscript{{$\pm$0.0553}} & 0.1697\textsubscript{{$\pm$0.0226}} & 0.6774\textsubscript{{$\pm$0.0843}} & 0.1682\textsubscript{{$\pm$0.0145}} \\

CMTA \cite{cmta} & \ding{51} & ~ & \ding{51} & 0.6588\textsubscript{{$\pm$0.0252}} & 0.1692\textsubscript{{$\pm$0.0220}} & 0.6645\textsubscript{{$\pm$0.0638}} & 0.1686\textsubscript{{$\pm$0.0212}} \\

SurvPath \cite{survpath} & \ding{51} & ~ & \ding{51} & 0.6520\textsubscript{{$\pm$0.0465}} & 0.1819\textsubscript{{$\pm$0.0186}} & 0.6652\textsubscript{{$\pm$0.0435}} & 0.1791\textsubscript{{$\pm$0.0203}} \\

MMP \cite{mmp} & \ding{51} & ~ & \ding{51} & 0.6635\textsubscript{{$\pm$0.1190}} & 0.2282\textsubscript{{$\pm$0.0127}} & 0.6626\textsubscript{{$\pm$0.1009}} & 0.2264\textsubscript{{$\pm$0.0147}} \\

MoME \cite{mome} & \ding{51} & ~ & \ding{51} & 0.6372\textsubscript{{$\pm$0.0170}} & 0.1816\textsubscript{{$\pm$0.0237}} & 0.6530\textsubscript{{$\pm$0.0450}} & 0.1764\textsubscript{{$\pm$0.0174}} \\

CCL \cite{ccl} & \ding{51} & ~ & \ding{51} & 0.6708\textsubscript{{$\pm$0.0437}} & 0.1700\textsubscript{{$\pm$0.0205}} & 0.6748\textsubscript{{$\pm$0.0653}} & 0.1690\textsubscript{{$\pm$0.0219}} \\

LD-CVAE \cite{ld-cvae} & \ding{51} & ~ & \ding{51} & 0.6728\textsubscript{{$\pm$0.0539}} & 0.1661\textsubscript{{$\pm$0.0072}} & 0.6839\textsubscript{{$\pm$0.0386}} & 0.1647\textsubscript{{$\pm$0.0176}} \\

DIMAF \cite{dimaf} & \ding{51} & ~ & \ding{51} & 0.6615\textsubscript{{$\pm$0.0706}} & 0.1672\textsubscript{{$\pm$0.0183}} & 0.6723\textsubscript{{$\pm$0.0492}} & 0.1666\textsubscript{{$\pm$0.0142}} \\

ProSurv \cite{prosurv} & \ding{51} & ~ & \ding{51} & 0.6459\textsubscript{{$\pm$0.0417}} & 0.1637\textsubscript{{$\pm$0.0199}} & 0.6548\textsubscript{{$\pm$0.0521}} & 0.1616\textsubscript{{$\pm$0.0152}} \\

HSFSurv \cite{hsfsurv} & \ding{51} & ~ & \ding{51} & 0.6751\textsubscript{{$\pm$0.0809}} & 0.1656\textsubscript{{$\pm$0.0130}} & 0.6852\textsubscript{{$\pm$0.0738}} & 0.1643\textsubscript{{$\pm$0.0179}} \\

MSRL \cite{msrl} & \ding{51} & ~ & \ding{51} & 0.6215\textsubscript{{$\pm$0.0406}} & 0.1612\textsubscript{{$\pm$0.0122}} & 0.6175\textsubscript{{$\pm$0.1005}} & 0.1597\textsubscript{{$\pm$0.0111}} \\

SlotSPE \cite{slotspe} & \ding{51} & ~ & \ding{51} & 0.6783\textsubscript{{$\pm$0.0459}} & 0.1633\textsubscript{{$\pm$0.0165}} & 0.6867\textsubscript{{$\pm$0.0713}} & 0.1632\textsubscript{{$\pm$0.0186}} \\

\midrule

MCAT \cite{mcat} & \ding{51} & \ding{51} & \ding{51} & 0.6536\textsubscript{{$\pm$0.0251}} & 0.1764\textsubscript{{$\pm$0.0100}} & 0.6601\textsubscript{{$\pm$0.0361}} & 0.1785\textsubscript{{$\pm$0.0249}} \\

MOTCat \cite{motcat} & \ding{51} & \ding{51} & \ding{51} & 0.6708\textsubscript{{$\pm$0.0292}} & 0.1685\textsubscript{{$\pm$0.0126}} & 0.6820\textsubscript{{$\pm$0.0246}} & 0.1639\textsubscript{{$\pm$0.0098}} \\

CMTA \cite{cmta} & \ding{51} & \ding{51} & \ding{51} & 0.6608\textsubscript{{$\pm$0.0576}} & 0.1730\textsubscript{{$\pm$0.0235}} & 0.6597\textsubscript{{$\pm$0.0376}} & 0.1733\textsubscript{{$\pm$0.0154}} \\

SurvPath \cite{survpath} & \ding{51} & \ding{51} & \ding{51} & 0.6569\textsubscript{{$\pm$0.0155}} & 0.1703\textsubscript{{$\pm$0.0090}} & 0.6665\textsubscript{{$\pm$0.0356}} & 0.1749\textsubscript{{$\pm$0.0108}} \\

MMP \cite{mmp} & \ding{51} & \ding{51} & \ding{51} & 0.6199\textsubscript{{$\pm$0.0430}} & 0.2134\textsubscript{{$\pm$0.0330}} & 0.6277\textsubscript{{$\pm$0.0509}} & 0.2272\textsubscript{{$\pm$0.0210}} \\

MoME \cite{mome} & \ding{51} & \ding{51} & \ding{51} & 0.6524\textsubscript{{$\pm$0.0753}} & 0.1776\textsubscript{{$\pm$0.0120}} & 0.6558\textsubscript{{$\pm$0.0432}} & 0.1760\textsubscript{{$\pm$0.0333}} \\

CCL \cite{ccl} & \ding{51} & \ding{51} & \ding{51} & 0.6771\textsubscript{{$\pm$0.0380}} & 0.1678\textsubscript{{$\pm$0.0102}} & 0.6883\textsubscript{{$\pm$0.0470}} & 0.1642\textsubscript{{$\pm$0.0129}} \\

LD-CVAE \cite{ld-cvae} & \ding{51} & \ding{51} & \ding{51} & 0.6817\textsubscript{{$\pm$0.0515}} & 0.1653\textsubscript{{$\pm$0.0184}} & 0.6904\textsubscript{{$\pm$0.0462}} & 0.1602\textsubscript{{$\pm$0.0102}} \\

DIMAF \cite{dimaf} & \ding{51} & \ding{51} & \ding{51} & 0.6333\textsubscript{{$\pm$0.0325}} & 0.1698\textsubscript{{$\pm$0.0162}} & 0.6468\textsubscript{{$\pm$0.0162}} & 0.1678\textsubscript{{$\pm$0.0142}} \\

ProSurv \cite{prosurv} & \ding{51} & \ding{51} & \ding{51} & 0.6363\textsubscript{{$\pm$0.0465}} & 0.1656\textsubscript{{$\pm$0.0147}} & 0.6556\textsubscript{{$\pm$0.0297}} & 0.1643\textsubscript{{$\pm$0.0194}} \\

HSFSurv \cite{hsfsurv} & \ding{51} & \ding{51} & \ding{51} & 0.6757\textsubscript{{$\pm$0.0312}} & 0.1664\textsubscript{{$\pm$0.0145}} & \underline{0.6930\textsubscript{{$\pm$0.0859}}} & 0.1594\textsubscript{{$\pm$0.0187}} \\

MSRL \cite{msrl} & \ding{51} & \ding{51} & \ding{51} & 0.6277\textsubscript{{$\pm$0.0843}} & 0.1622\textsubscript{{$\pm$0.0139}} & 0.6329\textsubscript{{$\pm$0.0625}} & 0.1573\textsubscript{{$\pm$0.0102}} \\

SlotSPE \cite{slotspe} & \ding{51} & \ding{51} & \ding{51} & \underline{0.6845\textsubscript{{$\pm$0.0261}}} & 0.1620\textsubscript{{$\pm$0.0110}} & 0.6918\textsubscript{{$\pm$0.0446}} & 0.1601\textsubscript{{$\pm$0.0110}} \\

Ours & \ding{51} & \ding{51} & \ding{51} & \textbf{0.7060}\textsubscript{{$\pm$0.0542}} & \underline{0.1604\textsubscript{{$\pm$0.0144}}} & \textbf{0.7178}\textsubscript{{$\pm$0.0407}} & \textbf{0.1559}\textsubscript{{$\pm$0.0050}} \\

\bottomrule[1pt]
\end{longtable}
\end{scriptsize}

\normalsize

In addition, as a preliminary exploration toward fully automated inference, we evaluate a workflow that derives MedSAM2 bounding box prompts from DSDA-MT segmentation masks. Experiments and discussion are detailed in \ref{app1: auto segment}.

\subsection{Comparison with State-of-the-Art in Survival Analysis}
To demonstrate the effectiveness of the proposed framework in leveraging multimodal data, it is compared with SOTA methods for both unimodal and multimodal survival analysis. Although most of these methods rely on pathological whole-slide images (WSIs) and genomic data, they can still adapt to post-TIPS survival prediction with minor modifications due to structural similarities in input data. Specifically, the patch-level features from WSIs and the voxel-level deep learning features in this study both constitute large sets extracted by pretrained models, whereas genomic data, similar to radiomics and clinical features, consist of $1\times1$ attributes that are often grouped into distinct subsets \cite{mcat, motcat, survpath}. Furthermore, the multimodal SOTA approaches have shown remarkable generalization across diverse cancer datasets, validating the efficacy of their interaction and fusion techniques and thereby indicating their potential competitiveness in TIPS prognosis. Thus, we implement the following methods:
\begin{itemize}
\item  \textbf{Unimodal methods}: We consider SNN \cite{snn}, SNNTrans \cite{snn, transmil}, and S-MLP \cite{s-mlp} based on clinical data, along with Deep Sets \cite{deepsets}, AttnMIL \cite{attnmil}, DeepAttnMISL \cite{deepattnmisl}, CLAM-SB \cite{clam}, CLAM-MB \cite{clam}, TransMIL \cite{transmil}, DTFD-MIL \cite{dtfd}, ILRA \cite{ilra}, WiKG \cite{wikg}, and OTSurv \cite{otsurv} using deep learning features. 
\item \textbf{Bimodal methods}: To best match the input settings of SOTA approaches, deep learning features and non-image clinical features are employed to evaluate MCAT \cite{mcat}, Pathomic \cite{pathomic}, Porpoise \cite{porpoise}, MOTCat \cite{motcat}, CMTA \cite{cmta}, SurvPath \cite{survpath}, MMP \cite{mmp}, MoME \cite{mome}, CCL \cite{ccl}, LD-CVAE \cite{ld-cvae}, DIMAF \cite{dimaf}, ProSurv \cite{prosurv}, HSFSurv \cite{hsfsurv}, MSRL \cite{msrl}, and SlotSPE \cite{slotspe}.
\item \textbf{Trimodal methods}: Extending the bimodal setting, radiomics and clinical features are jointly incorporated to interact and fuse with deep learning features, realizing MCAT \cite{mcat}, MOTCat \cite{motcat}, CMTA \cite{cmta}, SurvPath \cite{survpath}, MMP \cite{mmp}, MoME \cite{mome}, CCL \cite{ccl}, LD-CVAE \cite{ld-cvae}, DIMAF \cite{dimaf}, ProSurv \cite{prosurv}, HSFSurv \cite{hsfsurv}, MSRL \cite{msrl}, and SlotSPE \cite{slotspe}.
\end{itemize}

Table~\ref{tab: survial compare SOTA} presents the comparative results of five-fold cross-validation on the internal dataset ($n = 306$). In comparisons between unimodal and bimodal methods, the latter achieve overall higher C-index values and comparable mBS results, demonstrating their effective integration of complementary information from different modalities. Moreover, unimodal methods based on clinical features show superior overall C-index performance over those using deep learning features, which validates the discriminative value of clinical data and supports their use in our approach to guide visual feature aggregation. When comparing bimodal and trimodal methods, it can be observed that the introduction of radiomics features yields no substantial gains for the previous SOTA models and even leads to inferior performance. This may be attributed to the heterogeneity between clinical and radiomics features, which limits effective multimodal interaction and fusion, indicating the necessity of a trimodal interaction framework tailored to TIPS prognosis. In contrast, the proposed method facilitates progressive and sufficient cross-modal interactions among the three modalities, generating comprehensive multimodal representations and outperforming all bimodal and trimodal approaches across three of the four metrics. Particularly, it surpasses the previous best results in the two C-index measures by 2.15\% and 2.48\%, respectively, while achieving comparable mBS performance under Metric\textsubscript{DSDA-MT}. These results show that our framework achieves more accurate patient risk discrimination and absolute survival probability prediction, illustrating its superiority in leveraging trimodal data. In addition, it demonstrates better performance with MedSAM2 \cite{medsam2_ours} as the segmentation model compared to DSDA-MT, indicating its ability to enhance survival prediction through higher-quality segmentation.

\begin{figure*}[t]
	\centering
	\includegraphics[width=0.635\paperwidth]{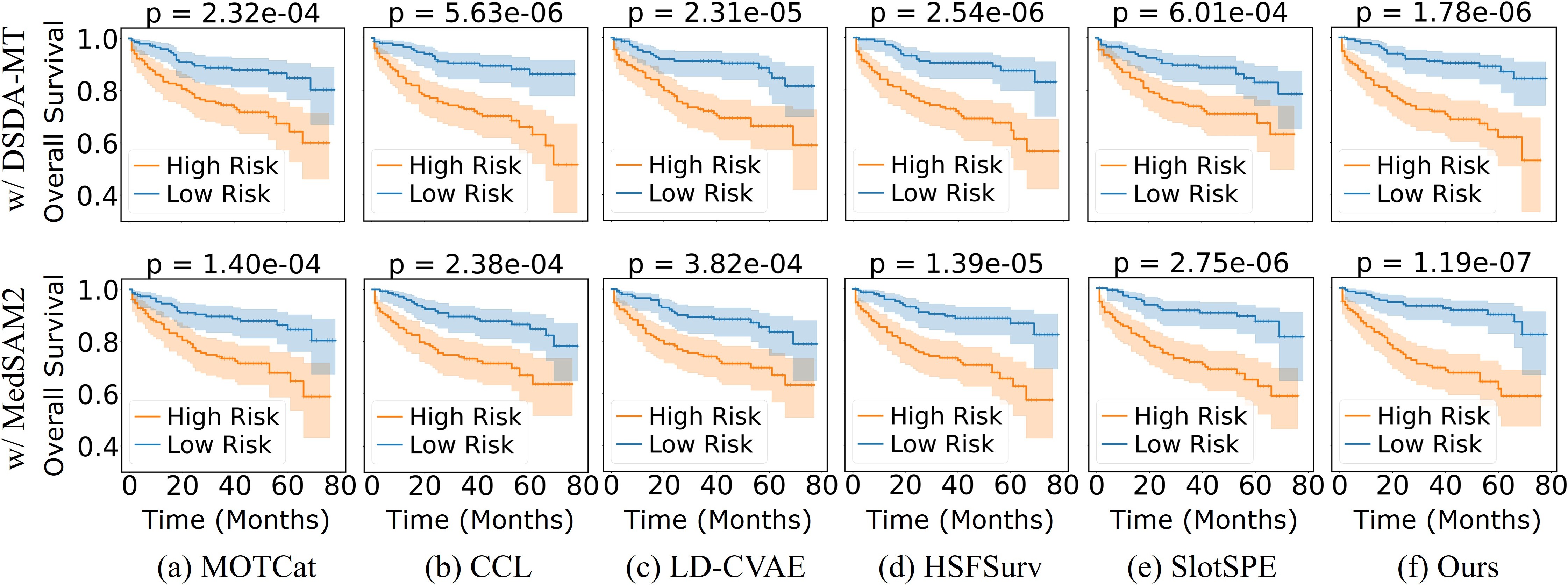}\caption{\label{fig:surv_K_M_sota}Kaplan-Meier analysis and log-rank test of the proposed method and its trimodal competitors on the internal dataset. ``w/ DSDA-MT'' and ``w/ MedSAM2'' refer to the segmentation models employed. Patients are stratified into high-risk (yellow) and low-risk (blue) groups, with shaded areas denoting confidence intervals. A lower p-value ($p<0.05$) indicates greater statistical significance.
}
\end{figure*}

To assess patient risk stratification, the Kaplan-Meier (KM) survival analysis \cite{KM} and the log-rank test \cite{logrank} are further performed on the proposed framework and its trimodal competitors, MOTCat, CCL, LD-CVAE, HSFSurv, and SlotSPE. As shown in Fig.~\ref{fig:surv_K_M_sota}, all patients from the internal dataset are divided into high- and low-risk groups based on the predicted median scores, with p-values calculated to determine whether the differences between the two groups are statistically significant. The results demonstrate that our method separates patients of high and low risk more clearly with either DSDA-MT or MedSAM2, consistently achieving lower p-values (well below 0.05) than competing approaches, thus confirming its effectiveness and superiority in patient stratification. Furthermore, the method exhibits a reduced p-value with the more precise segmentation, indicating its potential to improve risk stratification.

\begingroup
\setlength\aboverulesep{0pt}
\setlength\belowrulesep{0pt}
\begin{table}[!ht]
\scriptsize
\caption{Ablation results of proposed modules on multi-task prediction, including multi-grained radiomics attention (MGRA, ``M''), progressive orthogonal disentanglement (POD, ``P''), and clinically guided prognostic enhancement (CGPE, ``C''). An unchecked ``M'', ``P'' and ``C'' denote (1) single OT-based co-attention for interactions between radiomics and deep learning features, (2) exclusion of the orthogonality constraint $\mathcal{L}_{ortho}$, and (3) removal of cross-modal guidance from clinical features, respectively. $*$ indicates our default setting. Results for each model are reported in two rows, corresponding to DSDA-MT (first row) and MedSAM2 \cite{medsam2_ours} (second row) segmentation models, respectively. The best results for each are highlighted in \textbf{bold}.}
\label{tab: module ablation}
\centering
\setlength{\tabcolsep}{1pt}
\begin{tabular}{>{\centering\arraybackslash}m{30pt} 
                >{\centering\arraybackslash}m{15pt}
                >{\centering\arraybackslash}m{15pt}
                >{\centering\arraybackslash}m{15pt}
                >{\centering\arraybackslash}m{55pt}
                >{\centering\arraybackslash}m{55pt}
                >{\centering\arraybackslash}m{55pt}
                >{\centering\arraybackslash}m{55pt}
                >{\centering\arraybackslash}m{55pt}
                >{\centering\arraybackslash}m{55pt}}
\toprule[1pt]
\vspace{0.6em}
\multirow{2}{*}[0.05em]{Model} & \multirow{2}{*}[-0.25em]{M} & \multirow{2}{*}[-0.25em]{P} & \multirow{2}{*}[-0.25em]{C} & \multicolumn{2}{c}{Survival Analysis} &
\multicolumn{2}{c}{PPG Assessment} & \multicolumn{2}{c}{OHE Prediction} \\[-0.15em]
\cmidrule(lr){5-6} \cmidrule(lr){7-8} \cmidrule(lr){9-10}
 & & & & \vspace{0.25em}C-index$\uparrow$ & \vspace{0.25em}mBS$\downarrow$ & \vspace{0.25em}MAE$\downarrow$ & \vspace{0.25em}RMSE$\downarrow$ & \vspace{0.25em}C-index$\uparrow$ & \vspace{0.25em}mBS$\downarrow$ \\[0.2em]
 
\cline{1-10}
\rule{0pt}{0pt}
\vspace{0.15em}
\newcounter{tmp}
\setcounter{tmp}{1}
\multirow{2}{*}[-0.4em]{\hspace{-0.47em}\Roman{tmp}} 
& \multicolumn{3}{c}{\multirow{2}{*}{(Baseline)}} & 0.6657\textsubscript{{$\pm$0.0612}} & 0.1748\textsubscript{{$\pm$0.0181}} & 2.9111\textsubscript{{$\pm$0.4457}} & 3.8077\textsubscript{{$\pm$0.5121}} & 0.6281\textsubscript{{$\pm$0.0473}} & 0.1488\textsubscript{{$\pm$0.0213}} \\
& \multicolumn{3}{c}{} & 0.6761\textsubscript{{$\pm$0.0475}} & 0.1731\textsubscript{{$\pm$0.0078}} & 2.8862\textsubscript{{$\pm$0.4297}} & 3.7649\textsubscript{{$\pm$0.5012}} & 0.6332\textsubscript{{$\pm$0.0678}} & 0.1450\textsubscript{{$\pm$0.0313}} \\[0.15em]

\cline{1-10}
\rule{0pt}{0pt}
\vspace{0.15em}
\setcounter{tmp}{2}
\multirow{2}{*}[-0.4em]{\hspace{-0.7em}\Roman{tmp}} & \multirow{2}{*}[-0.2em]{\ding{51}} & \multirow{2}{*}[-0.2em]{} & \multirow{2}{*}[-0.2em]{} & 0.6821\textsubscript{{$\pm$0.0674}} & 0.1694\textsubscript{{$\pm$0.0219}} & 2.8739\textsubscript{{$\pm$0.4469}} & 3.6803\textsubscript{{$\pm$0.5235}} & 0.6487\textsubscript{{$\pm$0.0757}} & 0.1449\textsubscript{{$\pm$0.0285}} \\
& & & & 0.6892\textsubscript{{$\pm$0.0512}} & 0.1668\textsubscript{{$\pm$0.0106}} & 2.8445\textsubscript{{$\pm$0.4097}} & 3.6750\textsubscript{{$\pm$0.4827}} & 0.6511\textsubscript{{$\pm$0.0558}} & 0.1414\textsubscript{{$\pm$0.0296}} \\[0.15em]

\cline{1-10}
\rule{0pt}{0pt}
\vspace{0.15em}
\setcounter{tmp}{3}
\multirow{2}{*}[-0.4em]{\hspace{-0.7em}\Roman{tmp}} & \multirow{2}{*}[-0.2em]{} & \multirow{2}{*}[-0.2em]{\ding{51}} & \multirow{2}{*}[-0.2em]{} & 0.6836\textsubscript{{$\pm$0.0494}} & 0.1692\textsubscript{{$\pm$0.0199}} & 2.8612\textsubscript{{$\pm$0.4191}} & 3.6807\textsubscript{{$\pm$0.4795}} & 0.6423\textsubscript{{$\pm$0.0289}} & 0.1454\textsubscript{{$\pm$0.0170}} \\
& & & & 0.6908\textsubscript{{$\pm$0.0653}} & 0.1660\textsubscript{{$\pm$0.0138}} & 2.8476\textsubscript{{$\pm$0.4231}} & 3.6774\textsubscript{{$\pm$0.4792}} & 0.6484\textsubscript{{$\pm$0.0514}} & 0.1439\textsubscript{{$\pm$0.0394}} \\[0.15em]

\cline{1-10}
\rule{0pt}{0pt}
\vspace{0.15em}
\setcounter{tmp}{4}
\multirow{2}{*}[-0.4em]{\hspace{-0.7em}\Roman{tmp}} & \multirow{2}{*}[-0.2em]{} & \multirow{2}{*}[-0.2em]{} & \multirow{2}{*}[-0.2em]{\ding{51}} & 0.6851\textsubscript{{$\pm$0.0738}} & 0.1679\textsubscript{{$\pm$0.0181}} & 2.8456\textsubscript{{$\pm$0.4867}} & 3.6607\textsubscript{{$\pm$0.5203}} & 0.6494\textsubscript{{$\pm$0.0518}} & 0.1460\textsubscript{{$\pm$0.0447}} \\
& & & & 0.6931\textsubscript{{$\pm$0.0766}} & 0.1653\textsubscript{{$\pm$0.0158}} & 2.8342\textsubscript{{$\pm$0.4345}} & 3.6458\textsubscript{{$\pm$0.5104}} & 0.6524\textsubscript{{$\pm$0.0392}} & 0.1412\textsubscript{{$\pm$0.0401}} \\[0.15em]

\cline{1-10}
\rule{0pt}{0pt}
\vspace{0.15em}
\setcounter{tmp}{5}
\multirow{2}{*}[-0.4em]{\hspace{-0.7em}\Roman{tmp}} & \multirow{2}{*}[-0.2em]{\ding{51}} & \multirow{2}{*}[-0.2em]{\ding{51}} & \multirow{2}{*}[-0.2em]{} & 0.6963\textsubscript{{$\pm$0.0541}} & 0.1638\textsubscript{{$\pm$0.0117}} & 2.8138\textsubscript{{$\pm$0.4237}} & 3.6288\textsubscript{{$\pm$0.4854}} & 0.6552\textsubscript{{$\pm$0.0431}} & 0.1424\textsubscript{{$\pm$0.0424}} \\
& & & & 0.7055\textsubscript{{$\pm$0.0409}} & 0.1599\textsubscript{{$\pm$0.0105}} & 2.8077\textsubscript{{$\pm$0.4463}} & 3.5979\textsubscript{{$\pm$0.5302}} & 0.6692\textsubscript{{$\pm$0.0358}} & 0.1393\textsubscript{{$\pm$0.0394}} \\[0.15em]

\cline{1-10}
\rule{0pt}{0pt}
\vspace{0.15em}
\setcounter{tmp}{6}
\multirow{2}{*}[-0.4em]{\hspace{-0.7em}\Roman{tmp}} & \multirow{2}{*}[-0.2em]{\ding{51}} & \multirow{2}{*}[-0.2em]{} & \multirow{2}{*}[-0.2em]{\ding{51}} & 0.7004\textsubscript{{$\pm$0.0246}} & 0.1632\textsubscript{{$\pm$0.0075}} & 2.8103\textsubscript{{$\pm$0.4039}} & 3.6110\textsubscript{{$\pm$0.4574}} & 0.6452\textsubscript{{$\pm$0.0456}} & 0.1423\textsubscript{{$\pm$0.0333}} \\
& & & & 0.7077\textsubscript{{$\pm$0.0437}} & 0.1593\textsubscript{{$\pm$0.0199}} & 2.7988\textsubscript{{$\pm$0.4103}} & 3.5820\textsubscript{{$\pm$0.4544}} & 0.6493\textsubscript{{$\pm$0.1026}} & 0.1401\textsubscript{{$\pm$0.0310}} \\[0.15em]

\cline{1-10}
\rule{0pt}{0pt}
\vspace{0.15em}
\setcounter{tmp}{7}
\multirow{2}{*}[-0.4em]{\hspace{-0.7em}\Roman{tmp}} & \multirow{2}{*}[-0.2em]{} & \multirow{2}{*}[-0.2em]{\ding{51}} & \multirow{2}{*}[-0.2em]{\ding{51}} & 0.6969\textsubscript{{$\pm$0.0687}} & 0.1627\textsubscript{{$\pm$0.0160}} & 2.8095\textsubscript{{$\pm$0.4270}} & 3.6064\textsubscript{{$\pm$0.4794}} & 0.6546\textsubscript{{$\pm$0.0510}} & 0.1411\textsubscript{{$\pm$0.0392}} \\
& & & & 0.7022\textsubscript{{$\pm$0.0439}} & 0.1612\textsubscript{{$\pm$0.0170}} & 2.8007\textsubscript{{$\pm$0.4373}} & 3.5894\textsubscript{{$\pm$0.4930}} & 0.6493\textsubscript{{$\pm$0.0398}} & 0.1414\textsubscript{{$\pm$0.0401}} \\[0.15em]

\cline{1-10}
\rule{0pt}{0pt}
\vspace{0.15em}
\setcounter{tmp}{8}
\multirow{2}{*}[-0.4em]{\hspace{-0.3em}$\text{\Roman{tmp}}^{*}$} & \multirow{2}{*}[-0.2em]{\ding{51}} & \multirow{2}{*}[-0.2em]{\ding{51}} & \multirow{2}{*}[-0.2em]{\ding{51}} & \textbf{0.7060}\textsubscript{{$\pm$0.0542}} & \textbf{0.1604}\textsubscript{{$\pm$0.0144}} & \textbf{2.7699}\textsubscript{{$\pm$0.4140}} & \textbf{3.5602}\textsubscript{{$\pm$0.4636}} & \textbf{0.6698}\textsubscript{{$\pm$0.0624}} & \textbf{0.1404}\textsubscript{{$\pm$0.0320}} \\
& & & & \textbf{0.7178}\textsubscript{{$\pm$0.0407}} & \textbf{0.1559}\textsubscript{{$\pm$0.0050}} & \textbf{2.7657}\textsubscript{{$\pm$0.4163}} & \textbf{3.5410}\textsubscript{{$\pm$0.4690}} & \textbf{0.6802}\textsubscript{{$\pm$0.0669}} & \textbf{0.1373}\textsubscript{{$\pm$0.0350}} \\[0.2em]

\bottomrule[1pt]

\end{tabular}
\end{table}
\endgroup

\normalsize

\subsection{Multi-Task Prediction Results and Ablation Study}

In this section, we present the performance of the proposed method on three TIPS prognostic tasks and conduct ablation studies on the internal dataset to assess module contributions and configuration effects.

\subsubsection{Overall Results with Module Analysis}

To manifest the effectiveness of our approach, the proposed modules are progressively incorporated into the baseline model, including multi-grained radiomics attention (MGRA), progressive orthogonal disentanglement (POD), and clinically guided prognostic enhancement (CGPE). As presented in Table~\ref{tab: module ablation}, models equipped with a single module (II--IV) consistently outperform the baseline across all multi-task metrics in both rows of results, while two-module models (V--VII) further achieve superior overall performance compared with their respective single-module counterparts. This confirms the effectiveness of our modules and their mutual compatibility. Moreover, when all three modules are integrated, our model yields significant improvements over all ablation variants, demonstrating that the hierarchical feature interaction of MGRA, the redundancy elimination of POD, and the clinical guidance of CGPE are complementary and jointly enable optimal multi-task TIPS prognosis. Notably, improved outcomes are attained with MedSAM2 \cite{medsam2_ours} over DSDA-MT, validating the benefit of more accurate segmentation for our framework despite higher annotation costs.

\begin{figure*}[!t]
	\centering
	\includegraphics[width=0.635\paperwidth]{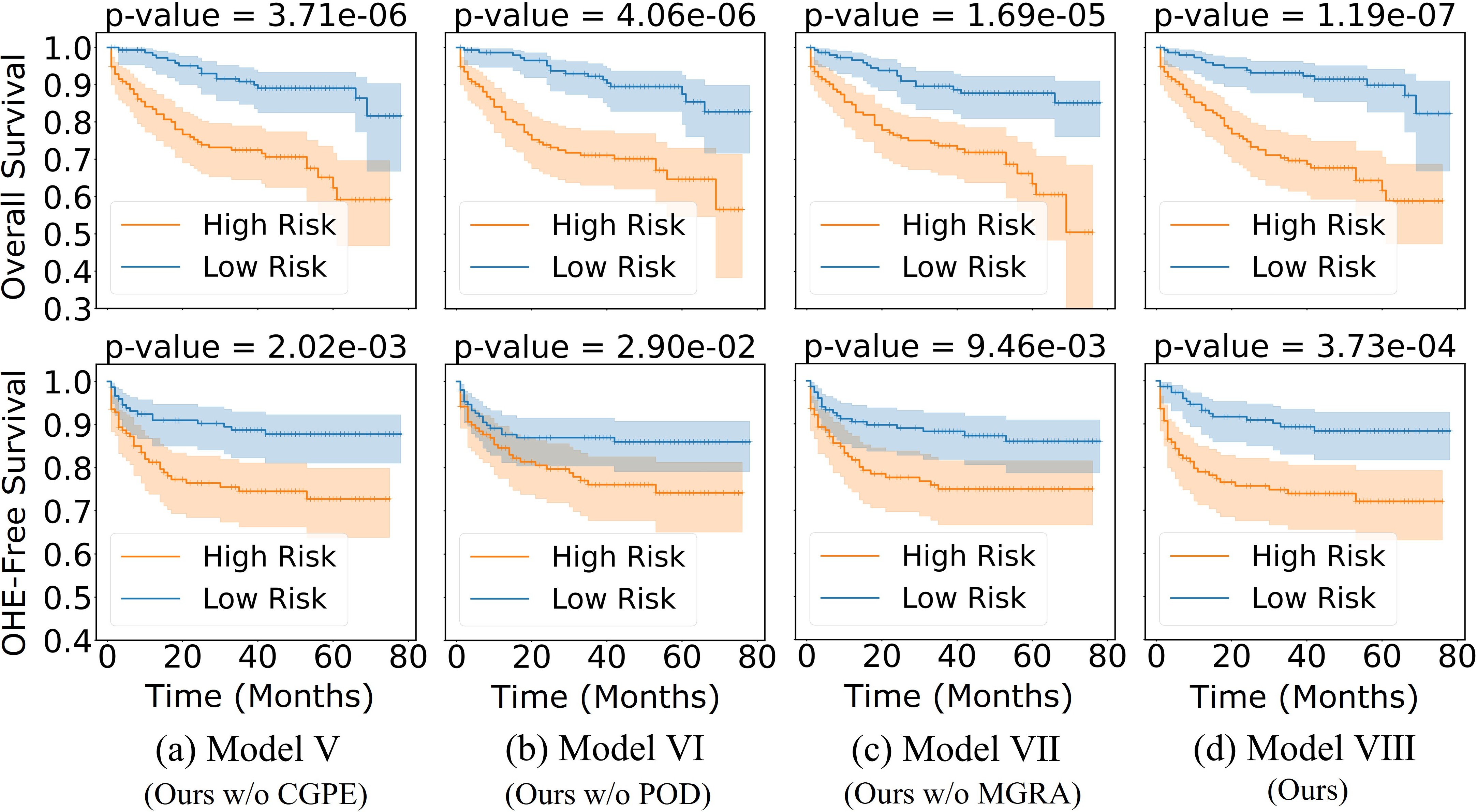}\caption{\label{fig:K_M_ablation}Kaplan-Meier analysis and log-rank test of the proposed method and its ablation variants based on the MedSAM2 segmentation model for survival (first row) and OHE (second row) prediction. ``w/o'' indicates removal of the corresponding module.
}
\end{figure*}

In addition, we employ KM analysis and the log-rank test to evaluate the patient stratification ability of the proposed method in survival and OHE prediction. Fig.~\ref{fig:K_M_ablation} illustrates the comparisons of Models V--VIII based on MedSAM2 in the two tasks. It can be observed that our method with all components exhibits the best stratification in both survival and OHE prediction, yielding lower p-values than the other ablation models. This demonstrates that the effective combination of the three proposed modules can enhance patient stratification, highlighting its potential clinical value for personalized treatment.

\begingroup
\setlength\aboverulesep{0pt}
\setlength\belowrulesep{0pt}
\begin{table}[ht]
\scriptsize
\caption{Comparative results of MGRA variants using the DSDA-MT segmentation model. In Variants I--III, radiomics features are partitioned by filter type ($N_{filt}=17$), feature type ($N_{feat}=7$), and joint filter-feature type ($N_g=103$), then interact with deep learning features through a single OT-based co-attention module. Variant IV denotes MGRA without fine-grained feature concatenation, and Variant V represents the default MGRA.}
\label{tab: MGRA ablation-1}
\centering
\setlength{\tabcolsep}{1pt}
\begin{tabular}{>{\centering\arraybackslash}m{55pt} 
                >{\centering\arraybackslash}m{55pt}
                >{\centering\arraybackslash}m{55pt}
                >{\centering\arraybackslash}m{55pt}
                >{\centering\arraybackslash}m{55pt}
                >{\centering\arraybackslash}m{55pt}
                >{\centering\arraybackslash}m{55pt}}
\toprule[1pt]
\vspace{0.6em}
\multirow{2}{*}[0.05em]{Variant}  & \multicolumn{2}{c}{Survival Analysis} & \multicolumn{2}{c}{PPG Assessment} & \multicolumn{2}{c}{OHE Prediction} \\[-0.15em]
\cmidrule(lr){2-3} \cmidrule(lr){4-5} \cmidrule(lr){6-7}
& \vspace{0.25em}C-index$\uparrow$ & \vspace{0.25em}mBS$\downarrow$ & \vspace{0.25em}MAE$\downarrow$& \vspace{0.25em}RMSE$\downarrow$ & \vspace{0.25em}C-index$\uparrow$ & \vspace{0.25em}mBS$\downarrow$ \\[0.2em]

\cline{1-7}
\setcounter{tmp}{1}
\vspace{0.3em}\Roman{tmp} & \vspace{0.3em}0.6855\textsubscript{$\pm$0.0714} & \vspace{0.3em}0.1696\textsubscript{$\pm$0.0283} & \vspace{0.3em}2.8710\textsubscript{$\pm$0.4870} & \vspace{0.3em}3.7040\textsubscript{$\pm$0.5678} & \vspace{0.3em}\textbf{0.6739}\textsubscript{$\pm$0.0423} & \vspace{0.3em}0.1407\textsubscript{$\pm$0.0418} \\

\setcounter{tmp}{2}
\Roman{tmp} & 0.6764\textsubscript{$\pm$0.0723} & 0.1764\textsubscript{$\pm$0.0183} & 2.8744\textsubscript{$\pm$0.4383} & 3.6684\textsubscript{$\pm$0.4840} & 0.6480\textsubscript{$\pm$0.0631} & 0.1444\textsubscript{$\pm$0.0291} \\

\setcounter{tmp}{3}
\Roman{tmp} & 0.6969\textsubscript{$\pm$0.0687} & 0.1627\textsubscript{$\pm$0.0160} & 2.8095\textsubscript{$\pm$0.4270} & 3.6064\textsubscript{$\pm$0.4794} & 0.6546\textsubscript{$\pm$0.0510} & 0.1411\textsubscript{$\pm$0.0392} \\

\setcounter{tmp}{4}
\Roman{tmp} & 0.6953\textsubscript{$\pm$0.0376} & 0.1623\textsubscript{$\pm$0.0103} & 2.8067\textsubscript{$\pm$0.3894} & 3.6020\textsubscript{$\pm$0.4267} & 0.6591\textsubscript{$\pm$0.0750} & 0.1413\textsubscript{$\pm$0.0265} \\

\setcounter{tmp}{5}
\Roman{tmp} (Ours) & \textbf{0.7060}\textsubscript{$\pm$0.0542} & \textbf{0.1604}\textsubscript{$\pm$0.0144} & \textbf{2.7699}\textsubscript{$\pm$0.4140} & \textbf{3.5602}\textsubscript{$\pm$0.4636} & 0.6698\textsubscript{$\pm$0.0624} & \textbf{0.1404}\textsubscript{$\pm$0.0320} \\[0.1em]

\bottomrule[1pt]

\end{tabular}
\end{table}
\endgroup

\normalsize

\begingroup
\setlength\aboverulesep{0pt}
\setlength\belowrulesep{0pt}
\begin{table}[ht]
\scriptsize
\caption{Ablation study on different configurations of $N_c$ using the DSDA-MT segmentation model. ``7 ($N_{feat}$)'' and ``17 ($N_{filt}$)'' indicate aggregation into 7 and 17 coarse-grained representations by feature and filter classes, respectively.}
\label{tab: MGRA ablation-2}
\centering
\setlength{\tabcolsep}{1pt}
\begin{tabular}{>{\centering\arraybackslash}m{60pt} 
                >{\centering\arraybackslash}m{55pt}
                >{\centering\arraybackslash}m{55pt}
                >{\centering\arraybackslash}m{55pt}
                >{\centering\arraybackslash}m{55pt}
                >{\centering\arraybackslash}m{55pt}
                >{\centering\arraybackslash}m{55pt}}
\toprule[1pt]
\vspace{0.6em}
\multirow{2}{*}[0.05em]{$N_c$}  & \multicolumn{2}{c}{Survival Analysis} & \multicolumn{2}{c}{PPG Assessment} & \multicolumn{2}{c}{OHE Prediction} \\[-0.15em]
\cmidrule(lr){2-3} \cmidrule(lr){4-5} \cmidrule(lr){6-7}
& \vspace{0.25em}C-index$\uparrow$ & \vspace{0.25em}mBS$\downarrow$ & \vspace{0.25em}MAE$\downarrow$& \vspace{0.25em}RMSE$\downarrow$ & \vspace{0.25em}C-index$\uparrow$ & \vspace{0.25em}mBS$\downarrow$ \\[0.2em]

\cline{1-7}
\vspace{0.3em}1 & \vspace{0.3em}0.6983\textsubscript{$\pm$0.0678} & \vspace{0.3em}0.1620\textsubscript{$\pm$0.0050} & \vspace{0.3em}2.7968\textsubscript{$\pm$0.4354} & \vspace{0.3em}3.5943\textsubscript{$\pm$0.4862} & \vspace{0.3em}0.6604\textsubscript{$\pm$0.0486} & \vspace{0.3em}0.1418\textsubscript{$\pm$0.0386} \\

3 & 0.7001\textsubscript{$\pm$0.0708} & 0.1617\textsubscript{$\pm$0.0068} & 2.7927\textsubscript{$\pm$0.4439} & 3.5844\textsubscript{$\pm$0.4868} & 0.6642\textsubscript{$\pm$0.0544} & 0.1414\textsubscript{$\pm$0.0259} \\

6 (Default) & \textbf{0.7060}\textsubscript{$\pm$0.0542} & \textbf{0.1604}\textsubscript{$\pm$0.0144} & \textbf{2.7699}\textsubscript{$\pm$0.4140} & \textbf{3.5602}\textsubscript{$\pm$0.4636} & \textbf{0.6698}\textsubscript{$\pm$0.0624} & 0.1404\textsubscript{$\pm$0.0320} \\

10 & 0.7015\textsubscript{$\pm$0.0465} & 0.1611\textsubscript{$\pm$0.0145} & 2.7739\textsubscript{$\pm$0.4262} & 3.5773\textsubscript{$\pm$0.4747} & 0.6682\textsubscript{$\pm$0.0771} & \textbf{0.1390}\textsubscript{$\pm$0.0327} \\

20 & 0.6971\textsubscript{$\pm$0.0441} & 0.1633\textsubscript{$\pm$0.0055} & 2.8062\textsubscript{$\pm$0.4461} & 3.6040\textsubscript{$\pm$0.5091} & 0.6572\textsubscript{$\pm$0.0295} & 0.1409\textsubscript{$\pm$0.0424} \\

50 & 0.6946\textsubscript{$\pm$0.0871} & 0.1643\textsubscript{$\pm$0.0119} & 2.8129\textsubscript{$\pm$0.4361} & 3.6101\textsubscript{$\pm$0.5197} & 0.6420\textsubscript{$\pm$0.0725} & 0.1439\textsubscript{$\pm$0.0313} \\

100 & 0.6950\textsubscript{$\pm$0.0728} & 0.1635\textsubscript{$\pm$0.0118} & 2.8223\textsubscript{$\pm$0.4555} & 3.6309\textsubscript{$\pm$0.4978} & 0.6443\textsubscript{$\pm$0.0586} & 0.1447\textsubscript{$\pm$0.0217} \\

7 ($N_{feat}$) & 0.6990\textsubscript{$\pm$0.0476} & 0.1616\textsubscript{$\pm$0.0099} & 2.7961\textsubscript{$\pm$0.4185} & 3.5818\textsubscript{$\pm$0.4894} & 0.6588\textsubscript{$\pm$0.0586} & 0.1408\textsubscript{$\pm$0.0362} \\

17 ($N_{filt}$) & 0.7004\textsubscript{$\pm$0.0672} & 0.1612\textsubscript{$\pm$0.0136} & 2.7801\textsubscript{$\pm$0.4133} & 3.5776\textsubscript{$\pm$0.4470} & 0.6503\textsubscript{$\pm$0.0789} & 0.1425\textsubscript{$\pm$0.0267} \\
[0.1em]

\bottomrule[1pt]

\end{tabular}
\end{table}
\endgroup

\normalsize

\subsubsection{Ablation Study on MGRA}
\textbf{Evaluation of MGRA Variants.} To investigate the effects of different interaction strategies and feature usage in radiomics and deep learning modalities, five variants of MGRA are considered, as follows:
\begin{itemize}
\item \textbf{Variant I}: Radiomics features are coarsely divided into $N_{filt}=17$ groups by filter type at Level III (Fig.~\ref{fig:radiomics_hierarchy}) and interact with deep learning features via a single OT-based co-attention module.
\item \textbf{Variant II}: Radiomics features are divided into $N_{feat}=7$ groups by feature type at Level III, with all other settings identical to Variant I.
\item \textbf{Variant III} (same as Model VII in Table~\ref{tab: module ablation}): Radiomics features are finely divided into $N_g=103$ groups at Level II, with all other settings identical to Variants I and II.
\item \textbf{Variant IV}: MGRA without fine-grained feature concatenation, i.e., using only $F^{d,c}$ and $F^{r,c}$ in Fig.~\ref{fig:main_method}.
\item \textbf{Variant V}: The default MGRA configuration.
\end{itemize}

Table~\ref{tab: MGRA ablation-1} reports the comparative results of the five models. Under the partitions of radiomics features at different granularities, Variant III demonstrates better overall performance than Variants I and II. This may be attributed to its finer-grained partitioning, facilitating semantic interactions with deep learning features and thus being adopted as the default setting. In the comparison among Variants III, IV, and V, the former two can be regarded as ablation models of Variant V, generating only fine-grained or coarse-grained representations of deep learning and radiomics, respectively. As observed, both models underperform Variant V across all six measures in the three tasks. This finding confirms that leveraging multi-granularity comprehensive features improves survival prediction and yields more robust shared representations, thereby benefiting PPG and OHE prediction.

\textbf{Impact of the Coarse-Grained Feature Count $N_c$.} $N_c$ determines the dimensionality of aggregated radiomics features, thereby influencing their interactions with deep learning features at the coarse level. To assess the effect of $N_c$ on model performance, ablation studies are performed with its varying values, as presented in Table~\ref{tab: MGRA ablation-2} (excluding the last two rows). We observe that the best overall performance is achieved at $N_c=6$, which also serves as the default setting in our experiments. Additionally, the impacts of different feature aggregation strategies are explored. Instead of applying the global attention pooling (GAP) \cite{attnmil} algorithm to all radiomics features, we aggregate them within their respective feature or filter classes (Level III in Fig.~\ref{fig:radiomics_hierarchy}) using GAP, thus preserving their intrinsic hierarchy. However, both results in the last two rows of Table~\ref{tab: MGRA ablation-2} exhibit some degradation compared to the default setting. This is mainly because the aggregation is restricted to predefined classes, which limits the representation of the resulting features.

\subsubsection{Ablation Study on POD}

\textbf{Configurations of the Orthogonal Ratio.} With the ratio of top-similar feature pairs selected for orthogonalization, the POD module decouples radiomics and deep learning features, thereby reducing shared redundancy.
Table~\ref{tab: POD strategy} presents the ablation results on various orthogonal ratio settings, involving fixed ratios and the dynamic threshold $\gamma_t$. Among the fixed ratio settings, a ratio of 0.1 performs better but remains inferior to $\gamma_t$, illustrating the efficacy of the well-designed dynamic threshold. Meanwhile, it is observed that applying full orthogonality to all feature pairs ($\mathrm{Ratio}=1.0$) leads to a substantial performance drop. This stems from the excessively wide scope of initial orthogonalization, which confuses the optimization direction and thus causes suboptimal prediction. The finding further validates the superiority of focusing on the most similar pairs to eliminate dominant redundancy. For the ablation analysis of $ \gamma_t $, model performance consistently degrades when either the temporal factor $\left(1 - \tfrac{t}{T} \right)^\alpha$ or the similarity scaling factor $\tfrac{\widetilde{\max}_t}{\max_0}$ is removed. The former can be attributed to the orthogonality constraint not extending to all feature pairs, thereby lacking global regularization, while the latter may be due to the absence of similarity-based adjustment, resulting in a ratio threshold misaligned with model training.

Moreover, an absolute cosine similarity-based thresholding strategy is evaluated, in which $\gamma_t$ is derived from the pairwise absolute similarity matrix $\left| S^{d,r} \right|$, since the objective is to fully decouple radiomics and deep learning features. The experimental results in the second-to-last row show comparable overall performance to the default configuration, but remain slightly inferior on most metrics. This supports the original design of prioritizing pairs with high cosine similarity, consistent with the distribution in Fig.~\ref{fig:histo} (a), and indicates that orthogonal regularization is insensitive to the sign of feature correlations after dominant redundancy is reduced. Consequently, the default $\gamma_t$ setting is adopted.

\begingroup
\setlength\aboverulesep{0pt}
\setlength\belowrulesep{0pt}
\begin{table}[!t]
\scriptsize
\caption{Ablation study on different configurations of the orthogonal ratio using the DSDA-MT segmentation model.}
\label{tab: POD strategy}
\centering
\setlength{\tabcolsep}{1pt}
\begin{tabular}{>{\centering\arraybackslash}m{65pt} 
                >{\centering\arraybackslash}m{55pt}
                >{\centering\arraybackslash}m{55pt}
                >{\centering\arraybackslash}m{55pt}
                >{\centering\arraybackslash}m{55pt}
                >{\centering\arraybackslash}m{55pt}
                >{\centering\arraybackslash}m{55pt}}
\toprule[1pt]
\vspace{0.6em}
\multirow{2}{*}[0.05em]{Ratio}  & \multicolumn{2}{c}{Survival Analysis} & \multicolumn{2}{c}{PPG Assessment} & \multicolumn{2}{c}{OHE Prediction} \\[-0.15em]
\cmidrule(lr){2-3} \cmidrule(lr){4-5} \cmidrule(lr){6-7}
& \vspace{0.25em}C-index$\uparrow$ & \vspace{0.25em}mBS$\downarrow$ & \vspace{0.25em}MAE$\downarrow$& \vspace{0.25em}RMSE$\downarrow$ & \vspace{0.25em}C-index$\uparrow$ & \vspace{0.25em}mBS$\downarrow$ \\[0.2em]

\cline{1-7}
\vspace{0.3em}0.05 & \vspace{0.3em}0.7036\textsubscript{$\pm$0.0528} & \vspace{0.3em}0.1618\textsubscript{$\pm$0.0118} & \vspace{0.3em}2.7921\textsubscript{$\pm$0.4261} & \vspace{0.3em}3.5805\textsubscript{$\pm$0.4744} & \vspace{0.3em}0.6566\textsubscript{$\pm$0.0515} & \vspace{0.3em}0.1416\textsubscript{$\pm$0.0384} \\

0.1 & 0.7046\textsubscript{$\pm$0.0592} & 0.1613\textsubscript{$\pm$0.0119} & 2.7844\textsubscript{$\pm$0.4183} & 3.5764\textsubscript{$\pm$0.4683} & 0.6596\textsubscript{$\pm$0.0561} & 0.1411\textsubscript{$\pm$0.0425} \\

0.2 & 0.7027\textsubscript{$\pm$0.0669} & 0.1619\textsubscript{$\pm$0.0107} & 2.8040\textsubscript{$\pm$0.4300} & 3.5943\textsubscript{$\pm$0.4933} & 0.6490\textsubscript{$\pm$0.0560} & 0.1417\textsubscript{$\pm$0.0294} \\

0.5 & 0.7002\textsubscript{$\pm$0.0438} & 0.1635\textsubscript{$\pm$0.0211} & 2.8050\textsubscript{$\pm$0.4390} & 3.6039\textsubscript{$\pm$0.4772} & 0.6504\textsubscript{$\pm$0.0527} & 0.1420\textsubscript{$\pm$0.0325} \\

1.0 & 0.6915\textsubscript{$\pm$0.0689} & 0.1662\textsubscript{$\pm$0.0099} & 2.8272\textsubscript{$\pm$0.4379} & 3.6299\textsubscript{$\pm$0.5024} & 0.6387\textsubscript{$\pm$0.0273} & 0.1441\textsubscript{$\pm$0.0421} \\[0.1em]

$\gamma _{t}$ w/o $(1-\frac{t}{T} )^{\alpha }$  & 0.7047\textsubscript{$\pm$0.0464} & 0.1616\textsubscript{$\pm$0.0119} & 2.7800\textsubscript{$\pm$0.4188} & 3.5746\textsubscript{$\pm$0.4976} & 0.6611\textsubscript{$\pm$0.0527} & 0.1413\textsubscript{$\pm$0.0411} \\[0.4em]

$\gamma _{t}$ w/o $\frac{\widetilde{\max}_t}{\max_0}$ & 0.7032\textsubscript{$\pm$0.0465} & 0.1619\textsubscript{$\pm$0.0051} & 2.7882\textsubscript{$\pm$0.4372} & 3.5836\textsubscript{$\pm$0.5300} & 0.6553\textsubscript{$\pm$0.0640} & 0.1415\textsubscript{$\pm$0.0395} \\[0.5em]

$\gamma_t$ w/ $|S^{d,r}|$ & 0.7051\textsubscript{$\pm$0.0462} & 0.1621\textsubscript{$\pm$0.0103} & 2.7703\textsubscript{$\pm$0.4038} & \textbf{3.5586}\textsubscript{$\pm$0.4552} & \textbf{0.6711}\textsubscript{$\pm$0.0588} & 0.1409\textsubscript{$\pm$0.0302} \\[0.5em]

$\gamma _{t}$ (Default)
 & \textbf{0.7060}\textsubscript{$\pm$0.0542} & \textbf{0.1604}\textsubscript{$\pm$0.0144} & \textbf{2.7699}\textsubscript{$\pm$0.4140} & 3.5602\textsubscript{$\pm$0.4636} & 0.6698\textsubscript{$\pm$0.0624} & \textbf{0.1404}\textsubscript{$\pm$0.0320} \\
[0.1em]

\bottomrule[1pt]

\end{tabular}
\end{table}
\endgroup

\normalsize

\begingroup
\setlength\aboverulesep{0pt}
\setlength\belowrulesep{0pt}
\begin{table}[!t]
\scriptsize
\caption{Ablation study on different orthogonality loss weights $\delta$ using the DSDA-MT segmentation model.}
\label{tab: POD loss weight}
\centering
\setlength{\tabcolsep}{1pt}
\begin{tabular}{>{\centering\arraybackslash}m{60pt} 
                >{\centering\arraybackslash}m{55pt}
                >{\centering\arraybackslash}m{55pt}
                >{\centering\arraybackslash}m{55pt}
                >{\centering\arraybackslash}m{55pt}
                >{\centering\arraybackslash}m{55pt}
                >{\centering\arraybackslash}m{55pt}}
\toprule[1pt]
\vspace{0.6em}
\multirow{2}{*}[0.05em]{$\delta$}  & \multicolumn{2}{c}{Survival Analysis} & \multicolumn{2}{c}{PPG Assessment} & \multicolumn{2}{c}{OHE Prediction} \\[-0.15em]
\cmidrule(lr){2-3} \cmidrule(lr){4-5} \cmidrule(lr){6-7}
& \vspace{0.25em}C-index$\uparrow$ & \vspace{0.25em}mBS$\downarrow$ & \vspace{0.25em}MAE$\downarrow$& \vspace{0.25em}RMSE$\downarrow$ & \vspace{0.25em}C-index$\uparrow$ & \vspace{0.25em}mBS$\downarrow$ \\[0.2em]

\cline{1-7}
\vspace{0.3em}0.01 & \vspace{0.3em}0.7036\textsubscript{$\pm$0.0450} & \vspace{0.3em}0.1610\textsubscript{$\pm$0.0117} & \vspace{0.3em}2.7883\textsubscript{$\pm$0.4228} & \vspace{0.3em}3.5860\textsubscript{$\pm$0.4815} & \vspace{0.3em}0.6686\textsubscript{$\pm$0.0414} & \vspace{0.3em}0.1409\textsubscript{$\pm$0.0410} \\

0.1 (Default) & \textbf{0.7060}\textsubscript{$\pm$0.0542} & \textbf{0.1604}\textsubscript{$\pm$0.0144} & 2.7699\textsubscript{$\pm$0.4140} & 3.5602\textsubscript{$\pm$0.4636} & \textbf{0.6698}\textsubscript{$\pm$0.0624} & \textbf{0.1404}\textsubscript{$\pm$0.0320} \\

1 & 0.7025\textsubscript{$\pm$0.0439} & 0.1613\textsubscript{$\pm$0.0168} & \textbf{2.7627}\textsubscript{$\pm$0.4147} & \textbf{3.5551}\textsubscript{$\pm$0.4576} & 0.6657\textsubscript{$\pm$0.0640} & 0.1414\textsubscript{$\pm$0.0363} \\

10 & 0.6964\textsubscript{$\pm$0.0323} & 0.1640\textsubscript{$\pm$0.0204} & 2.8056\textsubscript{$\pm$0.3958} & 3.5999\textsubscript{$\pm$0.4537} & 0.6511\textsubscript{$\pm$0.0302} & 0.1431\textsubscript{$\pm$0.0248} \\
[0.1em]

\bottomrule[1pt]

\end{tabular}
\end{table}
\endgroup

\normalsize

\textbf{Impact of the Orthogonality Loss Weight $\delta$.} Table~\ref{tab: POD loss weight} illustrates the ablation experiments with different $\delta$ values on the three prediction tasks. The best survival prediction is observed when $\delta=0.1$, indicating the optimal compatibility between the orthogonality loss and the survival loss during pretraining. Moreover, fine-tuning based on the corresponding shared features achieves comparable and superior performance for PPG and OHE prediction compared to other settings. Hence, we set $\delta$ to 0.1 to attain the best overall performance.

\begingroup
\setlength\aboverulesep{0pt}
\setlength\belowrulesep{0pt}
\begin{table}[!t]
\scriptsize
\caption{Comparative results of CGPE variants using the DSDA-MT segmentation model. Variant I indicates the removal of clinical guidance. Variant II retains solely the clinically guided deep learning and radiomics features, along with the clinical representation. Variant III applies clinical guidance before the transformers, while Variant IV denotes the default setting.}
\label{tab: CGPE}
\centering
\setlength{\tabcolsep}{1pt}
\begin{tabular}{>{\centering\arraybackslash}m{55pt} 
                >{\centering\arraybackslash}m{55pt}
                >{\centering\arraybackslash}m{55pt}
                >{\centering\arraybackslash}m{55pt}
                >{\centering\arraybackslash}m{55pt}
                >{\centering\arraybackslash}m{55pt}
                >{\centering\arraybackslash}m{55pt}}
\toprule[1pt]
\vspace{0.6em}
\multirow{2}{*}[0.05em]{Variant}  & \multicolumn{2}{c}{Survival Analysis} & \multicolumn{2}{c}{PPG Assessment} & \multicolumn{2}{c}{OHE Prediction} \\[-0.15em]
\cmidrule(lr){2-3} \cmidrule(lr){4-5} \cmidrule(lr){6-7}
& \vspace{0.25em}C-index$\uparrow$ & \vspace{0.25em}mBS$\downarrow$ & \vspace{0.25em}MAE$\downarrow$& \vspace{0.25em}RMSE$\downarrow$ & \vspace{0.25em}C-index$\uparrow$ & \vspace{0.25em}mBS$\downarrow$ \\[0.2em]

\cline{1-7}
\setcounter{tmp}{1}
\vspace{0.3em}\Roman{tmp} & \vspace{0.3em}0.6963\textsubscript{$\pm$0.0541} & \vspace{0.3em}0.1638\textsubscript{$\pm$0.0117} & \vspace{0.3em}2.8138\textsubscript{$\pm$0.4237} & \vspace{0.3em}3.6288\textsubscript{$\pm$0.4854} & \vspace{0.3em}0.6552\textsubscript{$\pm$0.0431} & \vspace{0.3em}0.1424\textsubscript{$\pm$0.0424} \\

\setcounter{tmp}{2}
\Roman{tmp} & 0.6928\textsubscript{$\pm$0.0474} & 0.1641\textsubscript{$\pm$0.0188} & 2.8008\textsubscript{$\pm$0.3998} & 3.6066\textsubscript{$\pm$0.4568} & 0.6590\textsubscript{$\pm$0.0360} & 0.1413\textsubscript{$\pm$0.0234} \\

% \setcounter{tmp}{3}
% \Roman{tmp} & 0.6880\textsubscript{$\pm$0.0454} & 0.1669\textsubscript{$\pm$0.0145} & 2.8195\textsubscript{$\pm$0.4380} & 3.6305\textsubscript{$\pm$0.4875} & 0.6531\textsubscript{$\pm$0.0595} & 0.1421\textsubscript{$\pm$0.0443} \\

\setcounter{tmp}{3}
\Roman{tmp} & 0.7005\textsubscript{$\pm$0.0561} & 0.1625\textsubscript{$\pm$0.0057} & 2.7887\textsubscript{$\pm$0.4181} & 3.5839\textsubscript{$\pm$0.4804} & 0.6634\textsubscript{$\pm$0.0390} & 0.1406\textsubscript{$\pm$0.0242} \\

\setcounter{tmp}{4}
\Roman{tmp} (Ours) & \textbf{0.7060}\textsubscript{$\pm$0.0542} & \textbf{0.1604}\textsubscript{$\pm$0.0144} & \textbf{2.7699}\textsubscript{$\pm$0.4140} & \textbf{3.5602}\textsubscript{$\pm$0.4636} & \textbf{0.6698}\textsubscript{$\pm$0.0624} & \textbf{0.1404}\textsubscript{$\pm$0.0320} \\[0.1em]

\bottomrule[1pt]

\end{tabular}
\end{table}
\endgroup

\normalsize

\subsubsection{Ablation Study on CGPE}
To identify the most effective cross-modal guidance scheme with clinical features, we conduct an ablation analysis on the CGPE module, including its four variants:
\begin{itemize}
\item \textbf{Variant I} (same as Model V in Table~\ref{tab: module ablation}): Clinical guidance is removed, and only $h^d$, $h^r$, and $h^{cli}$ in Fig.~\ref{fig:main_method} are used.
\item \textbf{Variant II}: Employing solely the clinically guided deep learning and radiomics features $h^{d, cli}$ and $h^{r, cli}$, along with the clinical representation $h^{cli}$.
\item \textbf{Variant III}: Clinical guidance is applied prior to the transformers.
\item \textbf{Variant IV}: The default CGPE module.
\end{itemize}

The comparative results of the four models are presented in Table~\ref{tab: CGPE}. Compared to Variant IV, both Variants I and II show inferior results, indicating that clinical guidance facilitates the generation of discriminative prognostic features (I vs. IV), but superior performance is obtained only when combined with the inherent radiomics and deep learning features (II vs. IV). Furthermore, Variant IV consistently outperforms Variant III across all metrics, demonstrating a more effective strategy of integrating clinical guidance after the transformers.

\begingroup
\setlength\aboverulesep{0pt}
\setlength\belowrulesep{0pt}
\begin{table}[ht]
\scriptsize
\caption{Comparative results of the PPG regression head variants. Variant I denotes the inherent task-specific subnet with its final fully connected layer modified to a single output dimension. Variant II represents the default regression head, with the tanh activation replaced by sigmoid. Both Variants I and II directly predict the postoperative PPG value. Variant III indicates the default configuration, with postoperative PPG variation as the prediction target.}
\label{tab: PPG head ablation}
\centering
\setlength{\tabcolsep}{1pt}
\begin{tabular}{>{\centering\arraybackslash}m{70pt} 
                >{\centering\arraybackslash}m{70pt}
                >{\centering\arraybackslash}m{70pt}
                >{\centering\arraybackslash}m{70pt}
                >{\centering\arraybackslash}m{70pt}}
\toprule[1pt]
\vspace{0.6em}
\multirow{2}{*}[0.05em]{Variant}  & \multicolumn{2}{c}{Metric\textsubscript{DSDA-MT}} & \multicolumn{2}{c}{Metric\textsubscript{MedSAM2}} \\[-0.15em]
\cmidrule(lr){2-3} \cmidrule(lr){4-5}
& \vspace{0.25em}MAE$\downarrow$ & \vspace{0.25em}RMSE$\downarrow$ & \vspace{0.25em}MAE$\downarrow$ & \vspace{0.25em}RMSE$\downarrow$ \\[0.2em]

\hline
\setcounter{tmp}{1}
\vspace{0.3em}\Roman{tmp} & \vspace{0.3em}2.8811\textsubscript{$\pm$0.4932} & \vspace{0.3em}3.6511\textsubscript{$\pm$0.6018} & \vspace{0.3em}2.8618\textsubscript{$\pm$0.4894} & \vspace{0.3em}3.6390\textsubscript{$\pm$0.5687} \\

\setcounter{tmp}{2}
\Roman{tmp} & 2.8227\textsubscript{$\pm$0.5077} & 3.5840\textsubscript{$\pm$0.5781} & 2.8091\textsubscript{$\pm$0.4819} & 3.5630\textsubscript{$\pm$0.5310} \\

\setcounter{tmp}{3}
\Roman{tmp} (Ours) & \textbf{2.7699}\textsubscript{$\pm$0.4140} & \textbf{3.5602}\textsubscript{$\pm$0.4636} & \textbf{2.7657}\textsubscript{$\pm$0.4163} & \textbf{3.5410}\textsubscript{$\pm$0.4690} \\[0.1em]

\bottomrule[1pt]

\end{tabular}
\end{table}
\endgroup

\normalsize

\subsubsection{Selection of the PPG Regression Head}
To explore the impacts of different PPG regression heads and objectives, we propose three variants as follows:
\begin{itemize}
\item \textbf{Variant I}: The inherent task-specific subnet (Fig.~\ref{fig:predict_head} (a)) is employed, with its final fully connected layer modified to an output dimension of 1, directly predicting the postoperative PPG value.
\item \textbf{Variant II}: An extra subnet fed with pressure-related indicators is introduced to augment the task-specific pathway, as illustrated in Fig.~\ref{fig:predict_head} (b). The tanh activation function is replaced with sigmoid for direct postoperative PPG prediction.
\item \textbf{Variant III}: The default configuration (same structure as Fig.~\ref{fig:predict_head} (b), predicting postoperative PPG variation).
\end{itemize}

As shown in Table~\ref{tab: PPG head ablation}, Variant II achieves consistently better results than Variant I, with either DSDA-MT or MedSAM2 \cite{medsam2_ours} as the segmentation model, illustrating the effectiveness of the simple extra subnet and its extracted preoperative pressure-related features. Variant III further surpasses Variant II on all four metrics, validating that the estimation of PPG variation improves accuracy when preoperative PPG is available. Given its optimal performance, Variant III is adopted in our experiments.

\begin{figure*}[!t]
	\centering
	\includegraphics[width=0.635\paperwidth]{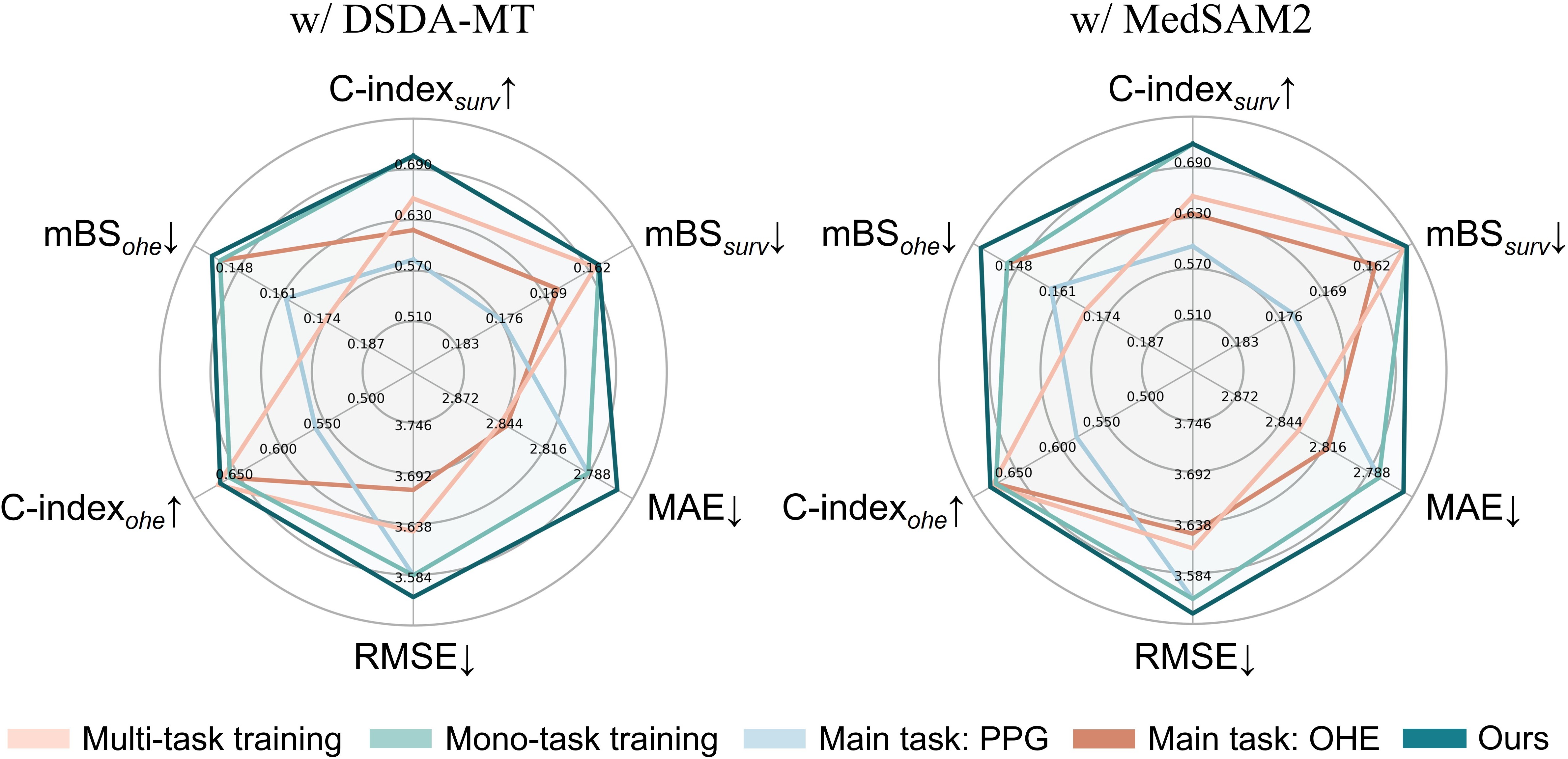}\caption{\label{fig:radar}Comparisons of overall performance across different training strategies. The subscripts ``surv'' and ``ohe'' indicate survival and OHE prediction metrics, respectively. Performance improves outward along each axis.
}
\end{figure*}

\subsubsection{Training Strategy Assessment}
To optimize the overall performance of the multi-task prediction framework and balance the outcomes across the three prognostic tasks, we evaluate several training strategies, including multi-task training, mono-task training, and three staged schemes with survival, PPG, and OHE prediction as the main task, respectively. Their details are presented as follows:
\begin{itemize}
\item \textbf{Multi-task training}: The three tasks are optimized concurrently with all loss functions.
\item \textbf{Mono-task training}: Three mono-task baselines are trained, each using its task-specific loss and the orthogonality constraint.
\item \textbf{Staged training (Main task: PPG estimation)}: The backbone is pretrained with $\mathcal{L}_{ppg} + \delta \mathcal{L}_{ortho}$, followed by fine-tuning of the survival and OHE prediction branches.
\item \textbf{Staged training (Main task: OHE prediction)}: The backbone is pretrained with $\mathcal{L}_{ohe} + \delta \mathcal{L}_{ortho}$, followed by fine-tuning of the survival and PPG prediction branches.
\item \textbf{Staged training (Main task: survival analysis)}: Our default setting.
\end{itemize}

Fig.~\ref{fig:radar} illustrates the mean performance across five folds for different training strategies. As observed, our approach attains the best overall performance under both segmentation models while yielding balanced outcomes on all six metrics of the three tasks. When compared with other methods, first, multi-task training exhibits significantly inferior performance to our method, with imbalanced results across metrics. Although comparable results are achieved on mBS\textsubscript{surv} and C-index\textsubscript{ohe}, it shows substantial drops on the remaining four measures. This may be attributed to concurrent multi-task training, which confuses model optimization, resulting in imbalanced results and reduced overall performance. Second, staged training with PPG or OHE prediction as the main task underperforms in fine-tuning the branches of the remaining tasks. It demonstrates the limited representation of the shared features and confirms the superiority of adopting the survival task, which involves broader prognostic factors and provides more generalized features. Third, compared to the mono-task baselines, the proposed method achieves consistent improvements across the four metrics for PPG and OHE prediction, particularly in MAE and RMSE. This result indicates that pretraining on the survival task generates more robust shared features, mitigating the risk of being trapped in local optima and thereby enhancing performance.

\begin{figure*}[!t]
	\centering
	\includegraphics[width=0.635\paperwidth]{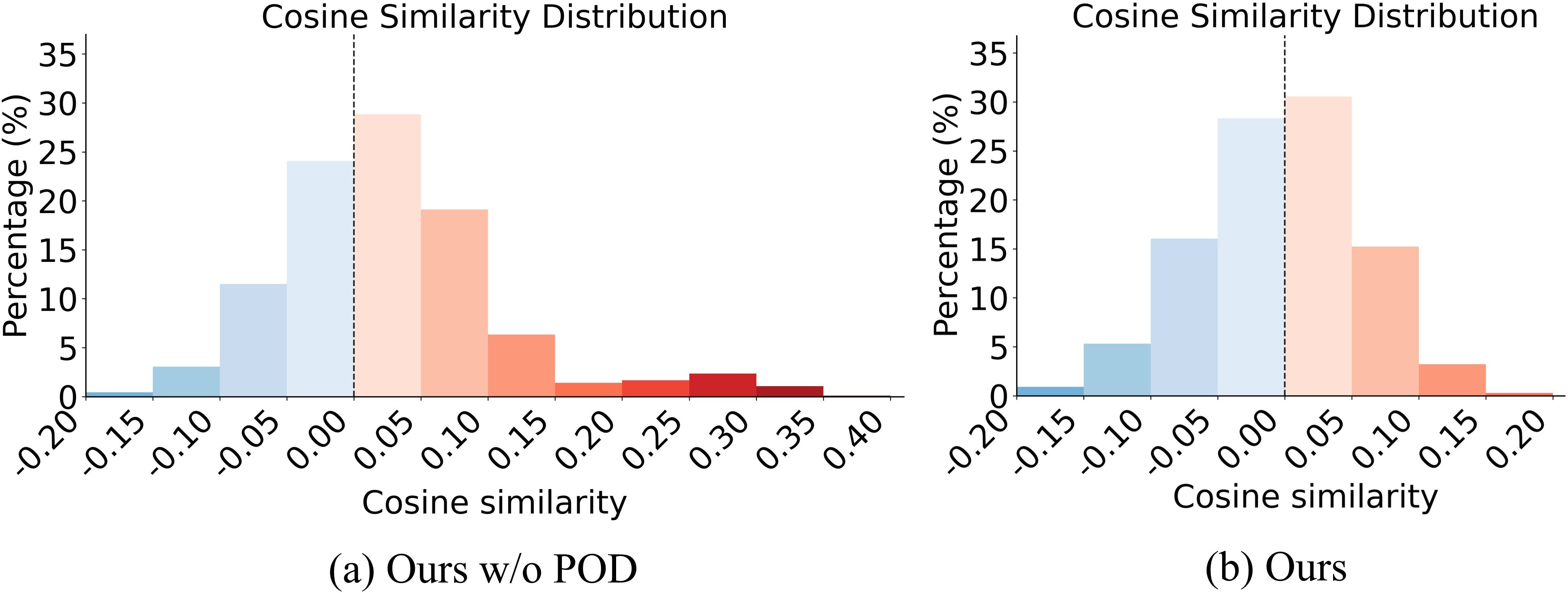}\caption{Cosine similarity distributions for the proposed method and its variant, based on the MedSAM2 \cite{medsam2_ours} segmentation model.}
    \label{fig:histo}
\end{figure*}

\subsection{Statistical Analysis}
% \subsubsection{Visualization of the Similarity Matrix}
To further examine the effect of POD on the orthogonalization between deep learning and radiomics features, the distributions of cosine similarity matrices $S^{d, r}$ on the internal dataset are shown in Fig.~\ref{fig:histo} for both the proposed model and its POD-free variant. Compared with the variant, our method substantially suppresses dominant redundancy, as evidenced by a notable reduction in high-similarity pairs. In addition, the increased proportion of low-similarity pairs between -0.05 and 0.05 demonstrates enhanced orthogonality of our approach, confirming the efficacy of the POD module.

\begin{figure}[H]
	\centering
	\includegraphics[width=0.635\paperwidth]{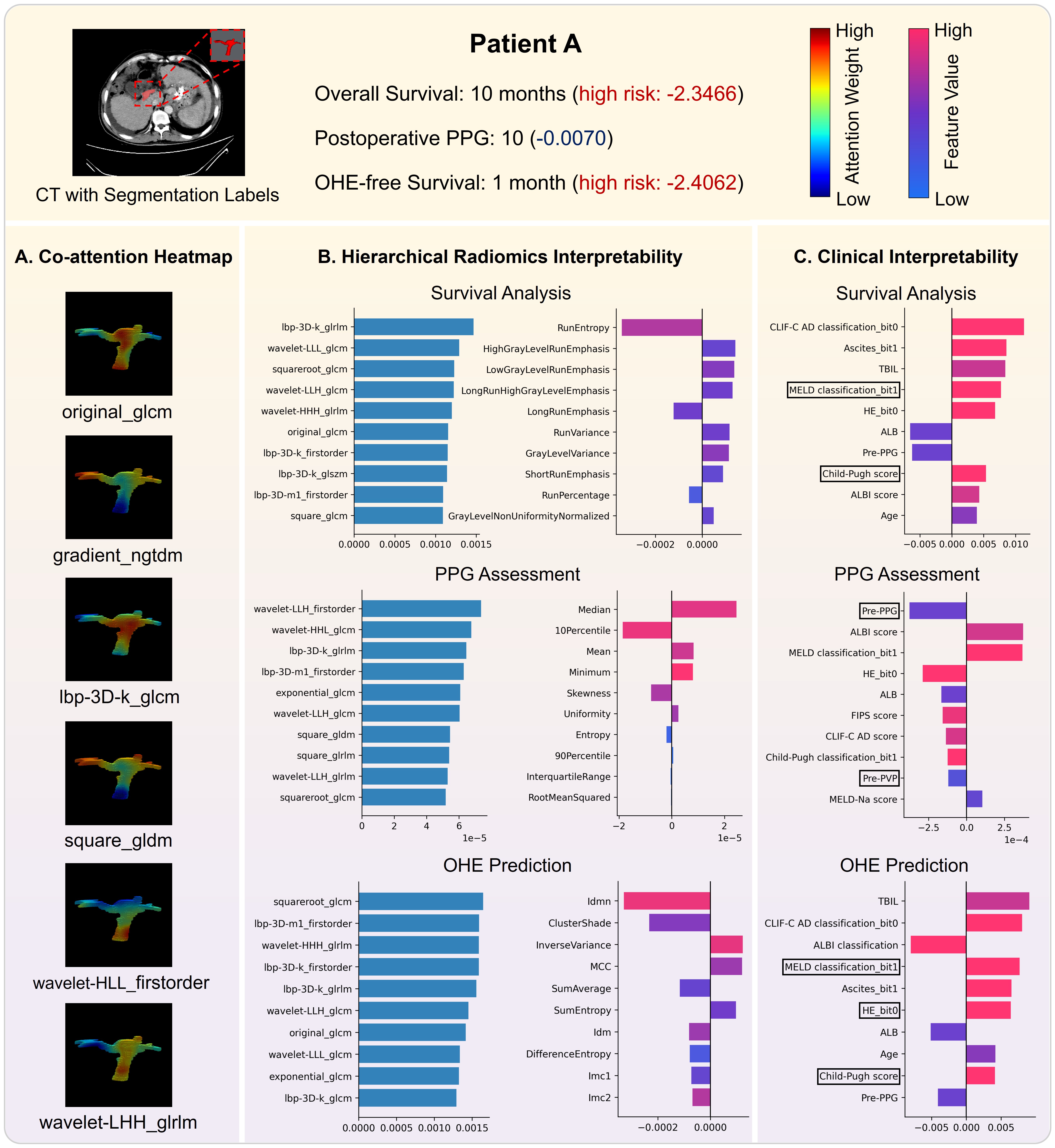}\caption{\label{fig:patient_high_risk}Multimodal interpretability of the proposed framework in a poor-prognosis case with MedSAM2 \cite{medsam2_ours} segmentation. In the top panel, the value in parentheses for postoperative PPG denotes the prediction deviation. Subfigures B and C show, from left to right, the absolute attributions of radiomics feature groups and the signed attributions of radiomics and clinical features. ``\_bitx'' in clinical features indicates the $x$-th position of one-hot encoding for categorical variables. Zoom in for a better view.
}
\end{figure}

\begin{figure}[H]
	\centering
	\includegraphics[width=0.635\paperwidth]{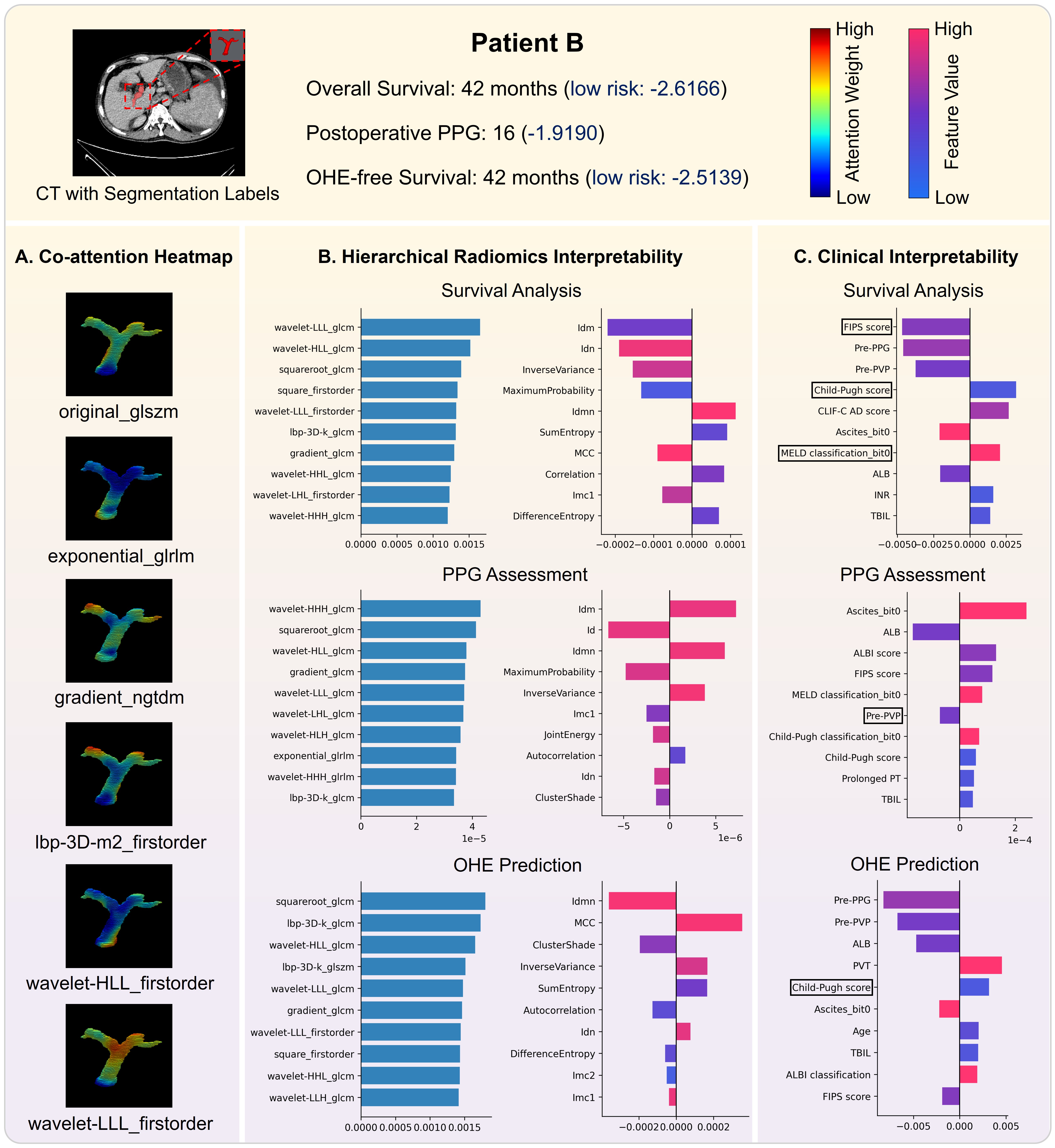}\caption{\label{fig:patient_low_risk}Multimodal interpretability of the proposed framework in a favorable-prognosis case with MedSAM2 \cite{medsam2_ours} segmentation. Zoom in for a better view.
}
\end{figure}

\subsection{Multimodal Interpretability}
Multimodal interpretability analysis can provide insights into the mechanisms of cross-modal interactions in our prognostic framework and explore the impacts of different modality features on multi-task prediction. To achieve this, the fine-grained co-attention scores between deep learning and radiomics features are mapped onto the original segmentation label space to visualize their interaction patterns. Additionally, Integrated Gradients (IG) \cite{IG} is employed to quantify the contributions of radiomics and clinical features to various tasks. As a gradient-based attribution method, IG attributes the model prediction to its input features, with positive attributions favoring an increase in the output value and negative attributions indicating a decrease. Figs.~\ref{fig:patient_high_risk} and \ref{fig:patient_low_risk} illustrate two representative cases for interpretation: Patient A with poor prognosis (high risk in survival and OHE prediction) and Patient B with favorable prognosis (low risk in both tasks).

In both subfigures A, six co-attention heatmaps between radiomics feature groups and deep learning features are presented. We can observe that different regions of the deep learning features are attended by distinct radiomics groups. Taking Fig.~\ref{fig:patient_high_risk}-A as an example, the high attention weights of groups ``original\_glcm'' and ``wavelet-LHH\_glrlm'' are mainly located in the trunk of the portal vein, whereas ``gradient\_ngtdm'' and ``square\_gldm'' show greater attention to the branches. Moreover, ``lbp-3D-k\_glcm'' and ``wavelet-HLL\_firstorder'' focus on the bifurcation and the lower segment of the main trunk, respectively. Note that our approach incorporates $N_g=103$ fine-grained co-attention units, enabling rich semantic interactions and diverse representations that may contribute to its superior performance.

Subfigures B demonstrate the interpretability of the employed radiomics features at both Level II (group, $N_g=103$) and Level I (element, $N_r=1,595$), reflecting their inherent hierarchical structure (Fig.~\ref{fig:radiomics_hierarchy}). Specifically, for each prognostic task, we apply IG to visualize the top 10 feature groups with the highest absolute attribution values (left panel). Furthermore, the 10 most contributing elements from the top-ranked group are presented in the right panel as an example (other groups can be displayed similarly). Consequently, key radiomics feature groups can be evaluated along with their constituent elements, thereby offering insights into the discovery of novel prognostic markers. For instance, gray level co-occurrence matrix (GLCM) features extracted from various preprocessed images highlight the critical role of texture representation, with the Inverse Difference Moment (IDM) indicating the significance of local homogeneity.

In subfigures C, the top 10 clinical features for each task are illustrated based on the IG algorithm. In survival and OHE prediction, the MELD \cite{MELD} classification indicator and Child-Pugh \cite{child-pugh} score exhibit high absolute importance for both Patients A and B. Meanwhile, the FIPS \cite{FIPS} score serves as the most crucial factor in reducing Patient B’s predicted mortality risk. The measures of MELD and Child-Pugh reflect hepatic function and compensation and have been widely confirmed to be associated with postoperative survival and OHE \cite{meld_child_survival, meld_child_ohe}, while the FIPS score is specifically designed for post-TIPS survival prediction. In addition, for Patient A, the preoperative grade of hepatic encephalopathy (HE\_bit0) increases the predicted risk of OHE, which is consistent with the patient’s prior history of HE. For PPG assessment, preoperative pressure-related indicators, e.g., Pre-PPG and Pre-PVP, show substantial contributions, indicating strong correlations with the postoperative PPG value. These findings demonstrate that our model can effectively identify and validate key clinical characteristics related to prognosis. Furthermore, its clinical interpretability may help uncover new prognostic factors, thereby providing guidance for therapeutic optimization.

\begingroup
\setlength\aboverulesep{0pt}
\setlength\belowrulesep{0pt}
\begin{table}[t]
\scriptsize
\caption{Comparative results of our method with trimodal SOTA approaches for survival analysis on the external dataset. Bootstrap 95\% CIs are reported as ranges in smaller type to reflect estimation uncertainty.}
\label{tab: external_evaluation_SOTA}
\centering
\setlength{\tabcolsep}{1pt}
\begin{tabular}{
    >{\centering\arraybackslash}m{60pt}
    >{\centering\arraybackslash}m{80pt}
    >{\centering\arraybackslash}m{80pt}
    >{\centering\arraybackslash}m{80pt}
    >{\centering\arraybackslash}m{80pt}
}
\toprule[1pt]
\vspace{0.6em}
\multirow{2}{*}[0.05em]{Method}
& \multicolumn{2}{c}{Metric\textsubscript{DSDA-MT}}
& \multicolumn{2}{c}{Metric\textsubscript{MedSAM2}} \\[-0.15em]

\cmidrule(lr){2-3} \cmidrule(lr){4-5}

& \vspace{0.25em}C-index$\uparrow$
& \vspace{0.25em}mBS$\downarrow$
& \vspace{0.25em}C-index$\uparrow$
& \vspace{0.25em}mBS$\downarrow$ \\[0.2em]

\hline

\vspace{0.3em}MCAT \cite{mcat}
& \vspace{0.3em}0.6066\textsubscript{~0.4913-0.7186}
& \vspace{0.3em}0.1780\textsubscript{~0.1761-0.1796}
& \vspace{0.3em}0.6328\textsubscript{~0.5224-0.7455}
& \vspace{0.3em}0.1732\textsubscript{~0.1677-0.1791}
\\

MOTCat \cite{motcat}
& 0.6085\textsubscript{~0.4995-0.7099}
& 0.1729\textsubscript{~0.1621-0.1831}
& 0.6355\textsubscript{~0.5276-0.7402}
& 0.1727\textsubscript{~0.1638-0.1818}
\\

CMTA \cite{cmta}
& 0.6081\textsubscript{~0.4805-0.7099}
& 0.1712\textsubscript{~0.1512-0.1934}
& 0.6124\textsubscript{~0.4831-0.7296}
& \textbf{0.1652}\textsubscript{~0.1647-0.1658}
\\[0.1em]

SurvPath \cite{survpath}
& 0.6359\textsubscript{~0.5066-0.7525}
& 0.1742\textsubscript{~0.1737-0.1751}
& 0.6551\textsubscript{~0.5177-0.7715}
& 0.1742\textsubscript{~0.1740-0.1744}
\\[0.1em]

MoME \cite{mome}
& 0.6230\textsubscript{~0.5126-0.7192}
& 0.1747\textsubscript{~0.1641-0.1859}
& 0.6390\textsubscript{~0.5175-0.7618}
& 0.1707\textsubscript{~0.1683-0.1731}
\\[0.1em]

CCL \cite{ccl}
& 0.6402\textsubscript{~0.5383-0.7414}
& 0.1650\textsubscript{~0.1632-0.1671}
& 0.6476\textsubscript{~0.5172-0.7645}
& 0.1673\textsubscript{~0.1665-0.1682}
\\[0.1em]

LD-CVAE \cite{ld-cvae}
& 0.6179\textsubscript{~0.4976-0.7336}
& \textbf{0.1649}\textsubscript{~0.1647-0.1651}
& 0.6676\textsubscript{~0.5584-0.7581}
& 0.1680\textsubscript{~0.1678-0.1682}
\\[0.1em]

HSFSurv \cite{hsfsurv}
& 0.6093\textsubscript{~0.4828-0.7328}
& 0.1686\textsubscript{~0.1683-0.1689}
& 0.6371\textsubscript{~0.5205-0.7471}
& 0.1768\textsubscript{~0.1691-0.1844}
\\[0.1em]

SlotSPE \cite{slotspe}
& 0.6449\textsubscript{~0.5269-0.7533}
& 0.1699\textsubscript{~0.1696-0.1703}
& 0.6590\textsubscript{~0.5581-0.7492}
& 0.1683\textsubscript{~0.1678-0.1687}
\\[0.1em]

Ours
& \textbf{0.6711}\textsubscript{~0.5604-0.7840}
& 0.1661\textsubscript{~0.1653-0.1670}
& \textbf{0.6891}\textsubscript{~0.5841-0.7968}
& 0.1683\textsubscript{~0.1680-0.1687}
\\[0.1em]

\bottomrule[1pt]

\end{tabular}
\end{table}
\endgroup

\normalsize

\begingroup
\setlength\aboverulesep{0pt}
\setlength\belowrulesep{0pt}

\begin{table}[t]
\scriptsize
\caption{Multi-task prediction results of our method on the external dataset. Bootstrap 95\% CIs are reported as ranges in smaller type.}
\label{tab: external_evaluation_multi-task}
\centering
\setlength{\tabcolsep}{1pt}
\renewcommand{\arraystretch}{1.15}

\begin{tabular}{
    >{\centering\arraybackslash}m{65pt}
    >{\centering\arraybackslash}m{55pt}
    >{\centering\arraybackslash}m{55pt}
    >{\centering\arraybackslash}m{55pt}
    >{\centering\arraybackslash}m{55pt}
    >{\centering\arraybackslash}m{55pt}
    >{\centering\arraybackslash}m{55pt}
}

\toprule[1pt]

\vspace{0.6em}
\multirow{2}{*}[0.05em]{Method}
& \multicolumn{2}{c}{Survival Analysis}
& \multicolumn{2}{c}{PPG Assessment}
& \multicolumn{2}{c}{OHE Prediction}
\\[-0.15em]

\cmidrule(lr){2-3}
\cmidrule(lr){4-5}
\cmidrule(lr){6-7}

& \vspace{0.25em}C-index$\uparrow$
& \vspace{0.25em}mBS$\downarrow$
& \vspace{0.25em}MAE$\downarrow$
& \vspace{0.25em}RMSE$\downarrow$
& \vspace{0.25em}C-index$\uparrow$
& \vspace{0.25em}mBS$\downarrow$
\\[0.2em]

\cline{1-7}

\vspace{0.7em}\shortstack{Ours w/\\DSDA-MT}
& \raisebox{0.4em}{0.6711\textsubscript{\shortstack{~0.5604-\\0.7840}}}
& \raisebox{0.4em}{\textbf{0.1661}\textsubscript{\shortstack{~0.1653-\\0.1670}}}
& \raisebox{0.4em}{3.0189\textsubscript{\shortstack{~2.6262-\\3.4254}}}
& \raisebox{0.4em}{3.8857\textsubscript{\shortstack{~3.3999-\\4.3651}}}
& \raisebox{0.4em}{0.6592\textsubscript{\shortstack{~0.5482-\\0.7709}}}
& \raisebox{0.4em}{\textbf{0.1470}\textsubscript{\shortstack{~0.1462-\\0.1478}}}
\\[0.8em]

\vspace{0.2em}\shortstack{Ours w/\\MedSAM2}
& \raisebox{0.7em}{\textbf{0.6891}\textsubscript{\shortstack{~0.5841-\\0.7968}}}
& \raisebox{0.7em}{0.1683\textsubscript{\shortstack{~0.1680-\\0.1687}}}
& \raisebox{0.7em}{\textbf{2.9847}\textsubscript{\shortstack{~2.6257-\\3.3757}}}
& \raisebox{0.7em}{\textbf{3.8586}\textsubscript{\shortstack{~3.4104-\\4.3058}}}
& \raisebox{0.7em}{\textbf{0.6686}\textsubscript{\shortstack{~0.5509-\\0.7942}}}
& \raisebox{0.7em}{0.1484\textsubscript{\shortstack{~0.1479-\\0.1489}}}
\\[0.3em]

\bottomrule[1pt]

\end{tabular}
\end{table}

\endgroup

\normalsize

\subsection{External Validation}
To further evaluate the generalization ability, models trained internally are tested on the external dataset ($n=137$), which comprises two additional cohorts. In particular, to fully utilize internal data, we retrain DSDA-MT with all unlabeled samples while leaving MedSAM2 \cite{medsam2_ours} unchanged. For post-TIPS prediction, training is conducted on the entire internal dataset, with evaluation on the external set. Tables~\ref{tab: external_evaluation_SOTA} and \ref{tab: external_evaluation_multi-task} present, respectively, the comparison of the proposed method with SOTA approaches on the survival task and its performance on multi-task prediction. Bootstrap 95\% confidence intervals (CIs) are also reported for all the validation metrics.

As observed, our method achieves the highest C-index under both the DSDA-MT and MedSAM2 settings, reaching 0.6711 (95\% CI, 0.5604–0.7840) and 0.6891 (95\% CI, 0.5841–0.7968), respectively. These results improve upon the previous best by 2.62\% and 2.15\%, demonstrating its superior discriminative ability on external cases. Moreover, it attains considerable accuracy and discrimination in PPG and OHE prediction, providing evidence of cross-domain generalization and robustness, which supports its potential clinical utility.

\section{Discussion}
\label{sec:Discussion}
To promote research on TIPS prognosis, we construct the first publicly available multi-center dataset, MultiTIPS, and propose a multimodal prognostic framework, which has demonstrated robust performance and promising clinical value across internal and external cohorts. Specifically, the framework integrates semi-supervised and foundation model-based segmentation pipelines that generate reliable portal vein labels with limited manual annotations, thus significantly reducing the workload of radiologists. The multimodal interaction module overcomes the single-modality limitations in prior works, effectively leveraging complementary information across modalities, and thereby enhancing clinical relevance. Furthermore, instead of single-endpoint assessment, our model performs a joint evaluation of postoperative survival, PPG, and OHE, offering more comprehensive guidance for individualized clinical decision-making. Additionally, multimodal interpretability is accessible in our method, which facilitates the validation of established prognostic markers and the discovery of novel ones. Note that most multimodal approaches rely on pathological whole-slide images and genomic data, with radiological imaging rarely incorporated. Hence, our study presents new insights into radiology-based diagnosis and prognosis.

However, this study does have some limitations. First, the MultiTIPS dataset currently covers only three centers in China and lacks cases from diverse countries and ethnicities. Therefore, its clinical utility and value for cross-domain evaluation remain limited. In addition, continuously incorporating more recent cohorts is important to capture evolving clinical practice and future technical advances in TIPS. Future work will involve larger, more comprehensive, and regularly updated cohorts to support broader diagnostic and prognostic tasks and facilitate real-world clinical translation of the predictive framework.

In terms of data utilization, this study primarily focuses on the portal vein, while overlooking other anatomical structures relevant to PH and TIPS prognosis, such as the inferior vena cava and spleen. Thus, leveraging multiple clinical structures in model design may lead to enhanced prognostic performance. Moreover, only portal phase CT images are employed in our study, with other phases excluded. With advances in multiphase CT-based prediction \cite{dv1, dv2}, superior performance may be achieved by integrating supplementary information across phases.

In terms of method design, the semi-supervised segmentation pipeline is solely based on two robust algorithms and a well-validated network, with minor modifications. Although the segmentation results are reliable, dedicated methods and network architectures for TIPS-related structures, such as the portal vein, still need to be developed for improved accuracy. For the foundation model-based pipeline, 3D segmentation techniques using bounding box propagation remain to be explored for anatomical regions exhibiting substantial inter-slice variation, aiming to balance precision with weakly supervised annotations. To enhance multimodal interaction, large language models (LLMs) are expected to provide innovative perspectives and methodologies. A recent pathology-based approach, CiMP \cite{cimp}, utilizes GPT-4 with clinical staging to generate patient-specific survival priors that induce prognostic representations in pathological and genomic modalities, leading to improved survival prediction. VLSA \cite{vlsa} employs GPT-4o to obtain critical prognostic features visible in WSIs for each cancer type, thereby guiding the aggregation of patch-level features. Hence, future research will introduce LLMs into the TIPS prognostic model to optimize unimodal representations, guide cross-modal interactions, and augment clinical features, thus advancing prognostic performance. Additionally, inter-task correlations have not been fully explored in multi-task prediction. More effective training strategies are needed for the synergistic optimization of survival, PPG, and OHE prediction.

Furthermore, in terms of clinical application, the current prognostic framework primarily assumes the availability of complete multimodal data. Although the simple zero-imputation strategy is used to maintain model continuity when individual features are missing, dedicated optimization for cases with incomplete data has not been implemented. However, in clinical practice, suboptimal imaging quality, anatomical variability, and incomplete examinations may lead to missing imaging or clinical features. Recent studies have explored methods for addressing incomplete data modalities. G-HANet \cite{G-HANet} captures histo-genomic associations via genomic data reconstruction during training, which are utilized at inference to enhance prognostic representations from unimodal WSIs. MSRL \cite{msrl} further leverages inter-case structural relevance and authentic genomic data from the training set to reconstruct missing modalities, thereby enabling accurate prediction based solely on histopathology. Accordingly, our future work will investigate more advanced approaches to improve robustness against incomplete data while preserving effective cross-modal interactions, ultimately enhancing the applicability and generalizability of TIPS prognostic models under complex real-world clinical conditions.

\section{Conclusion}
In this paper, we present the first multi-center dataset for TIPS prognosis, MultiTIPS, which comprises multiphase CT scans, clinical data, and expert annotations from 443 patients across three distinct cohorts. Driven by clinical needs, a multimodal TIPS prognostic framework is proposed, with the core objective of using limited manual annotations to achieve multimodal information fusion and comprehensive postoperative prediction. We realize this objective through three modules: dual-option portal vein segmentation, multimodal interaction, and multi-task prediction. Within the multimodal interaction module, MGRA, POD, and CGPE are further introduced to enhance inter-modal interaction and complementary representation, resulting in improved predictive accuracy. Moreover, the adopted staged training scheme ensures stable multi-task learning and robust overall performance. Extensive experiments on MultiTIPS demonstrate that the proposed method yields superior performance in survival, PPG, and OHE prediction, while exhibiting cross-domain generalization and interpretability, indicating its clinical potential. The limitations and future research directions are also discussed candidly. Moreover, the dataset and source code are publicly available for reproducing and developing novel algorithms, thus contributing to the clinical application.

\section*{Acknowledgments}
This work was supported by Chinese National Natural Science Foundation (Grant 62401069); the Beijing Advanced Innovation Center for Future Blockchain and Privacy Computing (GJJ-24-021); and the Innovation and Entrepreneurship Support Program of Beijing University of Posts and Telecommunications (2025-YC-T031).

\appendix
\section{Fully Automated Segmentation with MedSAM2}
\label{app1: auto segment}

\begin{table}[!t]
\scriptsize
\caption{Quantitative evaluation of MedSAM2 \cite{medsam2_ours} under different bounding box settings on the MultiTIPS validation set. Box prompts are generated from DSDA-MT segmentation masks or ground-truth labels. ``Fixed-pixel expansion'' denotes expanding each side of the box by fixed pixels at the original image scale, whereas ``Percentage expansion'' symmetrically extends the box width and height by a given percentage. The subscripts ``px'' and ``\%'' in method names indicate the expansion pixels and percentages, respectively. The best results are in \textbf{bold}, and the second-best are \underline{underlined}.}
\label{table appendix:MedSAM2 with DSDA-MT}
\centering
\setlength{\tabcolsep}{1pt}
\begin{tabular}{>{\centering\arraybackslash}m{52pt}
                >{\centering\arraybackslash}m{65pt}
                >{\centering\arraybackslash}m{55pt}
                >{\centering\arraybackslash}m{52pt}
                >{\centering\arraybackslash}m{52pt}
                >{\centering\arraybackslash}m{52pt}
                >{\centering\arraybackslash}m{52pt}}
\toprule[1pt]
\multirow{2}{*}[-0.3em]{Prompt source} 
& \multirow{2}{*}[-0.3em]{Box setting} 
& \multirow{2}{*}[-0.3em]{Method} 
& \multicolumn{4}{c}{Metric} \\
\cmidrule(lr){4-7}
& & & Dice(\%)$\uparrow$ & Jaccard(\%)$\uparrow$ & 95HD(Voxel)$\downarrow$ & ASD(Voxel)$\downarrow$ \\
\midrule

/ 
& / 
& DSDA-MT 
& 79.53\textsubscript{{$\pm$0.56}} 
& 66.42\textsubscript{{$\pm$0.68}} 
& \textbf{8.96}\textsubscript{{$\pm$1.35}} 
& 1.16\textsubscript{{$\pm$0.33}} \\
\midrule

\multirow{11}{*}[0em]{\makecell[c]{DSDA-MT\\mask}} 
& Minimum box 
& MedSAM2\textsubscript{Min} 
& 80.16\textsubscript{{$\pm$0.47}} 
& 67.32\textsubscript{{$\pm$0.59}} 
& 9.42\textsubscript{{$\pm$1.35}} 
& \textbf{1.09}\textsubscript{{$\pm$0.36}} \\
\cmidrule(lr){2-7}

& \multirow{5}{*}[0em]{\makecell[c]{Fixed-pixel\\ expansion}} 
& MedSAM2\textsubscript{1px} 
& 80.77\textsubscript{{$\pm$0.44}} 
& 68.15\textsubscript{{$\pm$0.55}} 
& 9.35\textsubscript{{$\pm$1.50}} 
& 1.18\textsubscript{{$\pm$0.38}} \\

& 
& MedSAM2\textsubscript{2px} 
& 80.83\textsubscript{{$\pm$0.45}}
& 68.21\textsubscript{{$\pm$0.56}} 
& 9.40\textsubscript{{$\pm$1.58}} 
& 1.27\textsubscript{{$\pm$0.41}} \\

& 
& MedSAM2\textsubscript{3px} 
& 80.46\textsubscript{{$\pm$0.55}} 
& 67.71\textsubscript{{$\pm$0.66}} 
& 9.21\textsubscript{{$\pm$1.58}} 
& 1.34\textsubscript{{$\pm$0.42}} \\

& 
& MedSAM2\textsubscript{5px} 
& 78.49\textsubscript{{$\pm$0.53}} 
& 65.04\textsubscript{{$\pm$0.61}} 
& 9.39\textsubscript{{$\pm$1.62}} 
& 1.57\textsubscript{{$\pm$0.45}} \\

& 
& MedSAM2\textsubscript{10px} 
& 65.32\textsubscript{{$\pm$0.71}} 
& 48.95\textsubscript{{$\pm$0.71}} 
& 11.28\textsubscript{{$\pm$3.19}} 
& 2.55\textsubscript{{$\pm$0.54}} \\
\cmidrule(lr){2-7}

& \multirow{5}{*}[0em]{\makecell[c]{Percentage\\ expansion}} 
& MedSAM2\textsubscript{2\%} 
& 80.31\textsubscript{{$\pm$0.46}} 
& 67.53\textsubscript{{$\pm$0.59}} 
& 9.35\textsubscript{{$\pm$1.31}} 
& \textbf{1.09}\textsubscript{{$\pm$0.36}} \\

& 
& MedSAM2\textsubscript{5\%} 
& 80.55\textsubscript{{$\pm$0.40}} 
& 67.85\textsubscript{{$\pm$0.49}} 
& 9.36\textsubscript{{$\pm$1.40}} 
& \underline{1.13\textsubscript{{$\pm$0.36}}} \\

& 
& MedSAM2\textsubscript{10\%} 
& \textbf{80.92}\textsubscript{{$\pm$0.48}} 
& \textbf{68.36}\textsubscript{{$\pm$0.60}} 
& 9.20\textsubscript{{$\pm$1.47}} 
& 1.17\textsubscript{{$\pm$0.38}} \\

& 
& MedSAM2\textsubscript{20\%} 
& \underline{80.84\textsubscript{{$\pm$0.50}}} 
& \underline{68.23\textsubscript{{$\pm$0.61}}} 
& \underline{9.08\textsubscript{{$\pm$1.54}}} 
& 1.26\textsubscript{{$\pm$0.38}} \\

& 
& MedSAM2\textsubscript{50\%} 
& 74.93\textsubscript{{$\pm$0.67}} 
& 60.34\textsubscript{{$\pm$0.75}} 
& 9.94\textsubscript{{$\pm$1.86}} 
& 1.96\textsubscript{{$\pm$0.39}} \\
\midrule

\graytext{GT mask} 
& \graytext{GT box} 
& \graytext{MedSAM2\textsubscript{GT}} 
& \graytext{91.32} 
& \graytext{84.07} 
& \graytext{1.00} 
& \graytext{0.20} \\

\bottomrule[1pt]
\end{tabular}
\end{table}
\normalsize

To explore the potential for fully automated MedSAM2 \cite{medsam2_ours} segmentation, we evaluate a workflow in which the required bounding box prompts are derived from semi-supervised segmentation masks rather than manual annotations during inference. Specifically, the trained DSDA-MT first generates pre-segmentation masks across slices, and the minimum bounding rectangles of all ROIs are extracted as box prompts for automated MedSAM2 segmentation. Furthermore, to mitigate potential boundary bias in pre-segmentation (particularly under-segmentation), both fixed-pixel and percentage-based box expansion strategies are adopted to improve target-region coverage. Experimental results are presented as mean $\pm$ standard deviation based on the five DSDA-MT models. Table~\ref{table appendix:MedSAM2 with DSDA-MT} shows the performance of MedSAM2 under different bounding box strategies on the MultiTIPS validation set, compared with DSDA-MT and MedSAM2 using GT-derived boxes.

The quantitative results demonstrate that, with appropriate bounding box settings, MedSAM2 achieves further segmentation gains over the DSDA-MT baseline. First, with DSDA-MT-derived minimum bounding boxes as prompts, MedSAM2\textsubscript{Min} outperforms DSDA-MT in Dice, Jaccard, and ASD, demonstrating the effectiveness of automatic box-prompt generation. Moreover, segmentation performance continues to improve with moderate box expansion, which is observed under both fixed-pixel ($\leq3$ pixels) and percentage-based ($\leq20\%$) settings. For example, MedSAM2\textsubscript{10\%} markedly surpasses DSDA-MT, achieving gains of 1.39\% in Dice and 1.94\% in Jaccard, with comparable results for the other two metrics. These findings confirm MedSAM2’s robustness to bounding box perturbations and indicate that appropriate box expansion provides a more complete effective prompt region. In contrast, performance tends to degrade when boxes are excessively extended, which may be attributed to the inclusion of more non-target regions. Additionally, MedSAM2 with automatically generated prompts remains inferior to MedSAM2\textsubscript{GT}, suggesting limitations in DSDA-MT segmentation quality.

\begin{figure*}[!t]
	\centering
	\includegraphics[width=0.635\paperwidth]{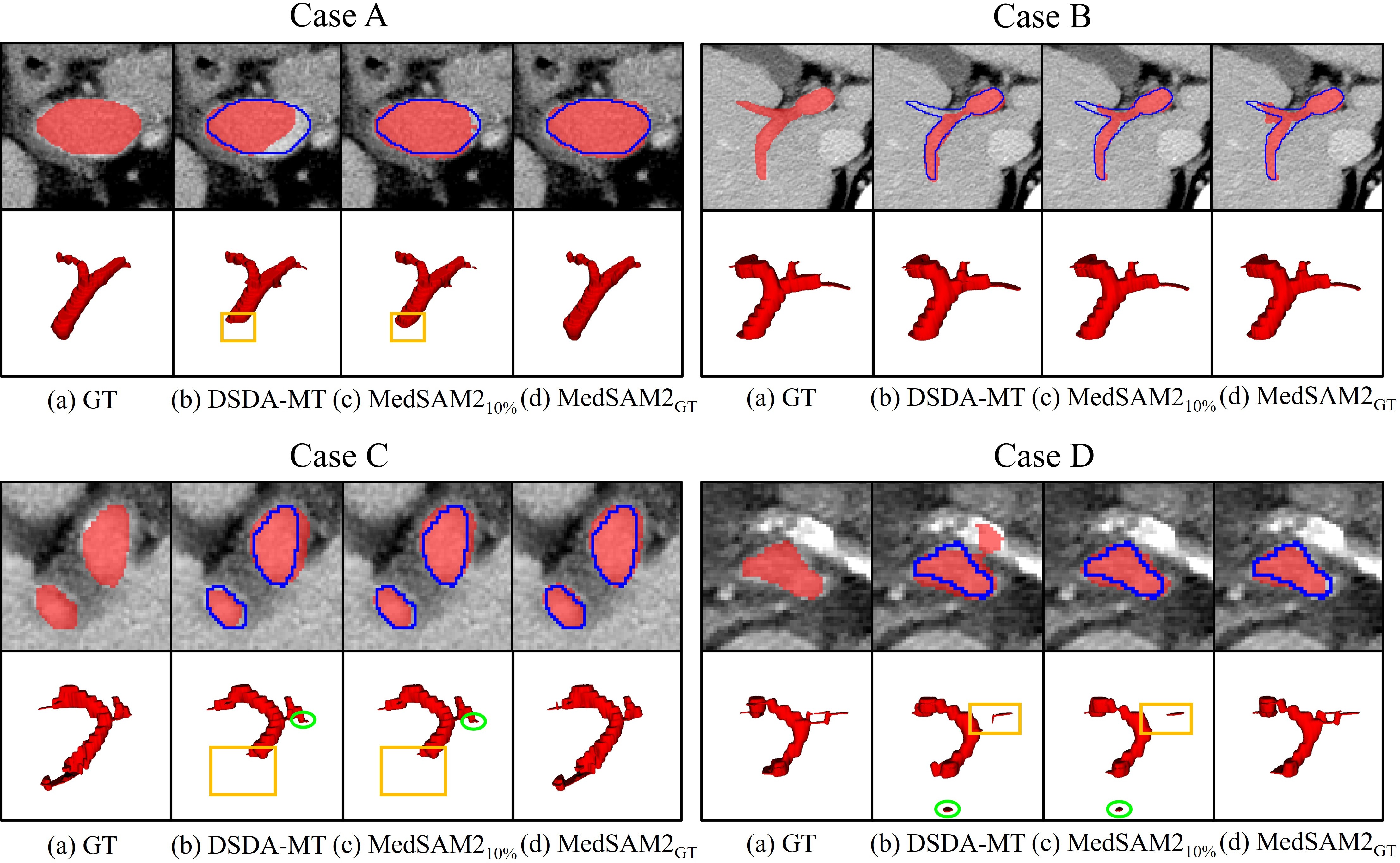}\caption{\label{fig_a:fully auto segment}Qualitative comparisons of fully automated MedSAM2 segmentation with reference methods in 2D axial and 3D views. Blue contours denote the ground-truth (GT) boundaries, yellow rectangles highlight false negative (FN) regions, and green ellipses indicate false positive (FP) regions.
}
\end{figure*}

To illustrate the segmentation performance of MedSAM2 using generated box prompts, we compare the robust MedSAM2\textsubscript{10\%} with DSDA-MT and MedSAM2\textsubscript{GT} in 2D axial and 3D views. As shown in Fig.~\ref{fig_a:fully auto segment}, MedSAM2\textsubscript{10\%} produces more refined boundaries than DSDA-MT, with predictions closer to the GT (indicated by the blue contours), supporting the effectiveness of DSDA-MT-derived box prompts. This is also consistent with the quantitative improvements, suggesting that MedSAM2’s multi-scale feature extraction and large-scale pretraining contribute to improved segmentation details. However, MedSAM2\textsubscript{10\%} exhibits limited ability to address false negative (FN) and false positive (FP) errors in DSDA-MT segmentation. Particularly, the lower main portal vein and fine branches are under-segmented in DSDA-MT (Cases A, C, and D; yellow rectangles) and remain poorly recovered by MedSAM2\textsubscript{10\%} due to the lack of corresponding prompts. Meanwhile, FP regions generated by DSDA-MT (green ellipses in Cases C and D) are not effectively removed by MedSAM2\textsubscript{10\%}, under the misleading guidance of the generated box prompts. In contrast, MedSAM2\textsubscript{GT} achieves more complete segmentation results closer to the GT with accurate GT-derived boxes.

Therefore, these quantitative and qualitative results provide preliminary evidence for the feasibility of automated MedSAM2 inference, which achieves superior segmentation performance over DSDA-MT alone. Nevertheless, its performance is mainly constrained by pre-segmentation accuracy, particularly the impact of FN and FP regions on automatic prompt generation. Further enhancements may be attained by developing more robust semi-supervised segmentation methods, incorporating negative box-prompt training, and applying adaptive box refinement.

\bibliographystyle{elsarticle-num-names} 

\bibliography{ref}

\begin{thebibliography}{84}
\expandafter\ifx\csname natexlab\endcsname\relax\def\natexlab#1{#1}\fi
\providecommand{\url}[1]{\texttt{#1}}
\providecommand{\href}[2]{#2}
\providecommand{\path}[1]{#1}
\providecommand{\DOIprefix}{doi:}
\providecommand{\ArXivprefix}{arXiv:}
\providecommand{\URLprefix}{URL: }
\providecommand{\Pubmedprefix}{pmid:}
\providecommand{\doi}[1]{\href{http://dx.doi.org/#1}{\path{#1}}}
\providecommand{\Pubmed}[1]{\href{pmid:#1}{\path{#1}}}
\providecommand{\bibinfo}[2]{#2}
\ifx\xfnm\relax \def\xfnm[#1]{\unskip,\space#1}\fi
%Type = Article
\bibitem[{Gracia-Sancho et~al.(2019)Gracia-Sancho, Marrone, and Fern{\'a}ndez-Iglesias}]{ph1}
\bibinfo{author}{J.~Gracia-Sancho}, \bibinfo{author}{G.~Marrone}, \bibinfo{author}{A.~Fern{\'a}ndez-Iglesias},
\newblock \bibinfo{title}{Hepatic microcirculation and mechanisms of portal hypertension},
\newblock \bibinfo{journal}{Nature reviews Gastroenterology \& hepatology} \bibinfo{volume}{16} (\bibinfo{year}{2019}) \bibinfo{pages}{221--234}.
%Type = Article
\bibitem[{Tonon and Piano(2023)}]{ph2}
\bibinfo{author}{M.~Tonon}, \bibinfo{author}{S.~Piano},
\newblock \bibinfo{title}{Cirrhosis and portal hypertension: how do we deal with ascites and its consequences},
\newblock \bibinfo{journal}{Medical Clinics} \bibinfo{volume}{107} (\bibinfo{year}{2023}) \bibinfo{pages}{505--516}.
%Type = Article
\bibitem[{Tripathi et~al.(2020)Tripathi, Stanley, Hayes, Travis, Armstrong, Tsochatzis, Rowe, Roslund, Ireland, Lomax et~al.}]{tips1}
\bibinfo{author}{D.~Tripathi}, \bibinfo{author}{A.~J. Stanley}, \bibinfo{author}{P.~C. Hayes}, \bibinfo{author}{S.~Travis}, \bibinfo{author}{M.~J. Armstrong}, \bibinfo{author}{E.~A. Tsochatzis}, \bibinfo{author}{I.~A. Rowe}, \bibinfo{author}{N.~Roslund}, \bibinfo{author}{H.~Ireland}, \bibinfo{author}{M.~Lomax}, et~al.,
\newblock \bibinfo{title}{Transjugular intrahepatic portosystemic stent-shunt in the management of portal hypertension},
\newblock \bibinfo{journal}{Gut} \bibinfo{volume}{69} (\bibinfo{year}{2020}) \bibinfo{pages}{1173--1192}.
%Type = Article
\bibitem[{Angeli et~al.(2018{\natexlab{a}})Angeli, Bernardi, Villanueva, Francoz, Mookerjee, Trebicka, Krag, Laleman, and Gines}]{tips2_guide}
\bibinfo{author}{P.~Angeli}, \bibinfo{author}{M.~Bernardi}, \bibinfo{author}{C.~Villanueva}, \bibinfo{author}{C.~Francoz}, \bibinfo{author}{R.~P. Mookerjee}, \bibinfo{author}{J.~Trebicka}, \bibinfo{author}{A.~Krag}, \bibinfo{author}{W.~Laleman}, \bibinfo{author}{P.~Gines},
\newblock \bibinfo{title}{Easl clinical practice guidelines for the management of patients with decompensated cirrhosis},
\newblock \bibinfo{journal}{Journal of hepatology} \bibinfo{volume}{69} (\bibinfo{year}{2018}{\natexlab{a}}) \bibinfo{pages}{406--460}.
%Type = Article
\bibitem[{Angeli et~al.(2018{\natexlab{b}})Angeli, Bernardi, Villanueva, Francoz, Mookerjee, Trebicka, Krag, Laleman, and Gines}]{tips3_guide}
\bibinfo{author}{P.~Angeli}, \bibinfo{author}{M.~Bernardi}, \bibinfo{author}{C.~Villanueva}, \bibinfo{author}{C.~Francoz}, \bibinfo{author}{R.~P. Mookerjee}, \bibinfo{author}{J.~Trebicka}, \bibinfo{author}{A.~Krag}, \bibinfo{author}{W.~Laleman}, \bibinfo{author}{P.~Gines},
\newblock \bibinfo{title}{Corrigendum to “easl clinical practice guidelines for the management of patients with decompensated cirrhosis”[j hepatol 69 (2018) 406--460]},
\newblock \bibinfo{journal}{Journal of Hepatology} \bibinfo{volume}{69} (\bibinfo{year}{2018}{\natexlab{b}}) \bibinfo{pages}{1207}.
%Type = Article
\bibitem[{Mamone et~al.(2024)Mamone, Comelli, Porrello, Milazzo, Di~Piazza, Stefano, Benfante, Tuttolomondo, Sparacia, Maruzzelli et~al.}]{N4}
\bibinfo{author}{G.~Mamone}, \bibinfo{author}{A.~Comelli}, \bibinfo{author}{G.~Porrello}, \bibinfo{author}{M.~Milazzo}, \bibinfo{author}{A.~Di~Piazza}, \bibinfo{author}{A.~Stefano}, \bibinfo{author}{V.~Benfante}, \bibinfo{author}{A.~Tuttolomondo}, \bibinfo{author}{G.~Sparacia}, \bibinfo{author}{L.~Maruzzelli}, et~al.,
\newblock \bibinfo{title}{Radiomics analysis of preprocedural ct imaging for outcome prediction after transjugular intrahepatic portosystemic shunt creation},
\newblock \bibinfo{journal}{Life} \bibinfo{volume}{14} (\bibinfo{year}{2024}) \bibinfo{pages}{726}.
%Type = Article
\bibitem[{Liu et~al.(2025)Liu, Jia, Zeng, Zeng, Qin, Peng, Tan, Zeng, Ou, Kun et~al.}]{N2}
\bibinfo{author}{D.-J. Liu}, \bibinfo{author}{L.-X. Jia}, \bibinfo{author}{F.-X. Zeng}, \bibinfo{author}{W.-X. Zeng}, \bibinfo{author}{G.-G. Qin}, \bibinfo{author}{Q.-F. Peng}, \bibinfo{author}{Q.~Tan}, \bibinfo{author}{H.~Zeng}, \bibinfo{author}{Z.-Y. Ou}, \bibinfo{author}{L.-Z. Kun}, et~al.,
\newblock \bibinfo{title}{Machine learning prediction of hepatic encephalopathy for long-term survival after transjugular intrahepatic portosystemic shunt in acute variceal bleeding},
\newblock \bibinfo{journal}{World Journal of Gastroenterology} \bibinfo{volume}{31} (\bibinfo{year}{2025}) \bibinfo{pages}{100401}.
%Type = Article
\bibitem[{Da et~al.(2025)Da, Chen, Wu, Guo, Zhou, Yin, Gao, Chen, Xiao, Wang et~al.}]{N6}
\bibinfo{author}{B.~Da}, \bibinfo{author}{H.~Chen}, \bibinfo{author}{W.~Wu}, \bibinfo{author}{W.~Guo}, \bibinfo{author}{A.~Zhou}, \bibinfo{author}{Q.~Yin}, \bibinfo{author}{J.~Gao}, \bibinfo{author}{J.~Chen}, \bibinfo{author}{J.~Xiao}, \bibinfo{author}{L.~Wang}, et~al.,
\newblock \bibinfo{title}{Development and validation of a machine learning-based model to predict survival in patients with cirrhosis after transjugular intrahepatic portosystemic shunt},
\newblock \bibinfo{journal}{EClinicalMedicine} \bibinfo{volume}{79} (\bibinfo{year}{2025}).
%Type = Article
\bibitem[{Elhakim et~al.(2025)Elhakim, Mansur, Kondo, Omar, Ahmed, Tabari, Brea, Ndakwah, Iqbal, Allegretti et~al.}]{N7}
\bibinfo{author}{T.~Elhakim}, \bibinfo{author}{A.~Mansur}, \bibinfo{author}{J.~Kondo}, \bibinfo{author}{O.~M.~F. Omar}, \bibinfo{author}{K.~Ahmed}, \bibinfo{author}{A.~Tabari}, \bibinfo{author}{A.~Brea}, \bibinfo{author}{G.~Ndakwah}, \bibinfo{author}{S.~Iqbal}, \bibinfo{author}{A.~S. Allegretti}, et~al.,
\newblock \bibinfo{title}{Beyond meld score: association of machine learning-derived ct body composition with 90-day mortality post transjugular intrahepatic portosystemic shunt placement},
\newblock \bibinfo{journal}{CardioVascular and Interventional Radiology} \bibinfo{volume}{48} (\bibinfo{year}{2025}) \bibinfo{pages}{221--230}.
%Type = Inproceedings
\bibitem[{Ronneberger et~al.(2015)Ronneberger, Fischer, and Brox}]{U-Net}
\bibinfo{author}{O.~Ronneberger}, \bibinfo{author}{P.~Fischer}, \bibinfo{author}{T.~Brox},
\newblock \bibinfo{title}{U-net: Convolutional networks for biomedical image segmentation},
\newblock in: \bibinfo{booktitle}{International Conference on Medical image computing and computer-assisted intervention}, \bibinfo{organization}{Springer}, \bibinfo{year}{2015}, pp. \bibinfo{pages}{234--241}.
%Type = Article
\bibitem[{Chen et~al.(2024)Chen, Huang, Yu, Chen, Xu, Jiang, Li, Zhao, Duan, Luo et~al.}]{xueguan}
\bibinfo{author}{X.~Chen}, \bibinfo{author}{M.~Huang}, \bibinfo{author}{X.~Yu}, \bibinfo{author}{J.~Chen}, \bibinfo{author}{C.~Xu}, \bibinfo{author}{Y.~Jiang}, \bibinfo{author}{Y.~Li}, \bibinfo{author}{Y.~Zhao}, \bibinfo{author}{C.~Duan}, \bibinfo{author}{Y.~Luo}, et~al.,
\newblock \bibinfo{title}{Hepatic-associated vascular morphological assessment to predict overt hepatic encephalopathy before tips: a multicenter study},
\newblock \bibinfo{journal}{Hepatology International} \bibinfo{volume}{18} (\bibinfo{year}{2024}) \bibinfo{pages}{1238--1248}.
%Type = Article
\bibitem[{Tarvainen and Valpola(2017)}]{mean_teacher}
\bibinfo{author}{A.~Tarvainen}, \bibinfo{author}{H.~Valpola},
\newblock \bibinfo{title}{Mean teachers are better role models: Weight-averaged consistency targets improve semi-supervised deep learning results},
\newblock \bibinfo{journal}{Advances in neural information processing systems} \bibinfo{volume}{30} (\bibinfo{year}{2017}).
%Type = Inproceedings
\bibitem[{Yang et~al.(2023)Yang, Qi, Feng, Zhang, and Shi}]{Unimatch}
\bibinfo{author}{L.~Yang}, \bibinfo{author}{L.~Qi}, \bibinfo{author}{L.~Feng}, \bibinfo{author}{W.~Zhang}, \bibinfo{author}{Y.~Shi},
\newblock \bibinfo{title}{Revisiting weak-to-strong consistency in semi-supervised semantic segmentation},
\newblock in: \bibinfo{booktitle}{Proceedings of the IEEE/CVF conference on computer vision and pattern recognition}, \bibinfo{year}{2023}, pp. \bibinfo{pages}{7236--7246}.
%Type = Article
\bibitem[{Ravi et~al.(2024)Ravi, Gabeur, Hu, Hu, Ryali, Ma, Khedr, R{\"a}dle, Rolland, Gustafson et~al.}]{sam2}
\bibinfo{author}{N.~Ravi}, \bibinfo{author}{V.~Gabeur}, \bibinfo{author}{Y.-T. Hu}, \bibinfo{author}{R.~Hu}, \bibinfo{author}{C.~Ryali}, \bibinfo{author}{T.~Ma}, \bibinfo{author}{H.~Khedr}, \bibinfo{author}{R.~R{\"a}dle}, \bibinfo{author}{C.~Rolland}, \bibinfo{author}{L.~Gustafson}, et~al.,
\newblock \bibinfo{title}{Sam 2: Segment anything in images and videos},
\newblock \bibinfo{journal}{arXiv preprint arXiv:2408.00714}  (\bibinfo{year}{2024}).
%Type = Article
\bibitem[{Zhou et~al.(2024)Zhou, Zhou, and Chen}]{ccl}
\bibinfo{author}{H.~Zhou}, \bibinfo{author}{F.~Zhou}, \bibinfo{author}{H.~Chen},
\newblock \bibinfo{title}{Cohort-individual cooperative learning for multimodal cancer survival analysis},
\newblock \bibinfo{journal}{IEEE Transactions on Medical Imaging}  (\bibinfo{year}{2024}).
%Type = Article
\bibitem[{Chen et~al.(2019)Chen, Ma, and Zheng}]{MedicalNet}
\bibinfo{author}{S.~Chen}, \bibinfo{author}{K.~Ma}, \bibinfo{author}{Y.~Zheng},
\newblock \bibinfo{title}{Med3d: Transfer learning for 3d medical image analysis},
\newblock \bibinfo{journal}{arXiv preprint arXiv:1904.00625}  (\bibinfo{year}{2019}).
%Type = Article
\bibitem[{Chen et~al.(2023)Chen, Wang, Ji, Luo, Lv, Wang, Zhao, Duan, Yu, Li et~al.}]{T23}
\bibinfo{author}{X.~Chen}, \bibinfo{author}{T.~Wang}, \bibinfo{author}{Z.~Ji}, \bibinfo{author}{J.~Luo}, \bibinfo{author}{W.~Lv}, \bibinfo{author}{H.~Wang}, \bibinfo{author}{Y.~Zhao}, \bibinfo{author}{C.~Duan}, \bibinfo{author}{X.~Yu}, \bibinfo{author}{Q.~Li}, et~al.,
\newblock \bibinfo{title}{3d automatic liver and spleen assessment in predicting overt hepatic encephalopathy before tips: a multi-center study},
\newblock \bibinfo{journal}{Hepatology International} \bibinfo{volume}{17} (\bibinfo{year}{2023}) \bibinfo{pages}{1545--1556}.
%Type = Article
\bibitem[{Chen et~al.(2024)Chen, Luo, Zhang, Liu, Liu, and Zhang}]{T30}
\bibinfo{author}{R.~Chen}, \bibinfo{author}{L.~Luo}, \bibinfo{author}{Y.-Z. Zhang}, \bibinfo{author}{Z.~Liu}, \bibinfo{author}{A.-L. Liu}, \bibinfo{author}{Y.-W. Zhang},
\newblock \bibinfo{title}{Bayesian network-based survival prediction model for patients having undergone post-transjugular intrahepatic portosystemic shunt for portal hypertension},
\newblock \bibinfo{journal}{World Journal of Gastroenterology} \bibinfo{volume}{30} (\bibinfo{year}{2024}) \bibinfo{pages}{1859}.
%Type = Article
\bibitem[{Zhao et~al.(2024)Zhao, Yang, Lv, Zhu, Chen, Wang, Huang, An, Duan, Yu et~al.}]{N5}
\bibinfo{author}{Y.~Zhao}, \bibinfo{author}{Y.~Yang}, \bibinfo{author}{W.~Lv}, \bibinfo{author}{S.~Zhu}, \bibinfo{author}{X.~Chen}, \bibinfo{author}{T.~Wang}, \bibinfo{author}{M.~Huang}, \bibinfo{author}{T.~An}, \bibinfo{author}{C.~Duan}, \bibinfo{author}{X.~Yu}, et~al.,
\newblock \bibinfo{title}{A modified model for predicting mortality after transjugular intrahepatic portosystemic shunt: A multicentre study},
\newblock \bibinfo{journal}{Liver International} \bibinfo{volume}{44} (\bibinfo{year}{2024}) \bibinfo{pages}{472--482}.
%Type = Article
\bibitem[{Sun et~al.(2022)Sun, Eche, Dorczynski, Otal, Revel-Mouroz, Zadro, Partouche, Fares, Maulat, Bureau et~al.}]{T20}
\bibinfo{author}{S.~H. Sun}, \bibinfo{author}{T.~Eche}, \bibinfo{author}{C.~Dorczynski}, \bibinfo{author}{P.~Otal}, \bibinfo{author}{P.~Revel-Mouroz}, \bibinfo{author}{C.~Zadro}, \bibinfo{author}{E.~Partouche}, \bibinfo{author}{N.~Fares}, \bibinfo{author}{C.~Maulat}, \bibinfo{author}{C.~Bureau}, et~al.,
\newblock \bibinfo{title}{Predicting death or recurrence of portal hypertension symptoms after tips procedures},
\newblock \bibinfo{journal}{European Radiology} \bibinfo{volume}{32} (\bibinfo{year}{2022}) \bibinfo{pages}{3346--3357}.
%Type = Article
\bibitem[{Queck et~al.(2023)Queck, Schwierz, Gu, Ferstl, Jansen, Uschner, Praktiknjo, Chang, Brol, Schepis et~al.}]{ppg1}
\bibinfo{author}{A.~Queck}, \bibinfo{author}{L.~Schwierz}, \bibinfo{author}{W.~Gu}, \bibinfo{author}{P.~G. Ferstl}, \bibinfo{author}{C.~Jansen}, \bibinfo{author}{F.~E. Uschner}, \bibinfo{author}{M.~Praktiknjo}, \bibinfo{author}{J.~Chang}, \bibinfo{author}{M.~J. Brol}, \bibinfo{author}{F.~Schepis}, et~al.,
\newblock \bibinfo{title}{Targeted decrease of portal hepatic pressure gradient improves ascites control after tips},
\newblock \bibinfo{journal}{Hepatology} \bibinfo{volume}{77} (\bibinfo{year}{2023}) \bibinfo{pages}{466--475}.
%Type = Article
\bibitem[{Xia et~al.(2023)Xia, Tie, Wang, Zhuge, Wu, Xue, Xu, Zhang, Zhao, Huang et~al.}]{ppg2}
\bibinfo{author}{Y.~Xia}, \bibinfo{author}{J.~Tie}, \bibinfo{author}{G.~Wang}, \bibinfo{author}{Y.~Zhuge}, \bibinfo{author}{H.~Wu}, \bibinfo{author}{H.~Xue}, \bibinfo{author}{J.~Xu}, \bibinfo{author}{F.~Zhang}, \bibinfo{author}{L.~Zhao}, \bibinfo{author}{G.~Huang}, et~al.,
\newblock \bibinfo{title}{Individualized portal pressure gradient threshold based on liver function categories in preventing rebleeding after tips},
\newblock \bibinfo{journal}{Hepatology International} \bibinfo{volume}{17} (\bibinfo{year}{2023}) \bibinfo{pages}{967--978}.
%Type = Article
\bibitem[{Kabelitz et~al.(2025)Kabelitz, Hartl, Schaub, Tiede, Rieland, Kornfehl, H{\"u}bener, Jachs, Hinrichs, Sch{\"u}tte et~al.}]{ppg0}
\bibinfo{author}{M.~A. Kabelitz}, \bibinfo{author}{L.~Hartl}, \bibinfo{author}{G.~Schaub}, \bibinfo{author}{A.~Tiede}, \bibinfo{author}{H.~Rieland}, \bibinfo{author}{A.~Kornfehl}, \bibinfo{author}{P.~H{\"u}bener}, \bibinfo{author}{M.~Jachs}, \bibinfo{author}{J.~Hinrichs}, \bibinfo{author}{S.~L. Sch{\"u}tte}, et~al.,
\newblock \bibinfo{title}{Identification of optimal portal pressure decrease to control ascites while minimizing he after tips: A multicenter study},
\newblock \bibinfo{journal}{Hepatology (Baltimore, Md.)}  (\bibinfo{year}{2025}).
%Type = Article
\bibitem[{De~Franchis et~al.(2022)De~Franchis, Bosch, Garcia-Tsao, Reiberger, Ripoll, Abraldes, Albillos, Baiges, Bajaj, Ba{\~n}ares et~al.}]{baveno}
\bibinfo{author}{R.~De~Franchis}, \bibinfo{author}{J.~Bosch}, \bibinfo{author}{G.~Garcia-Tsao}, \bibinfo{author}{T.~Reiberger}, \bibinfo{author}{C.~Ripoll}, \bibinfo{author}{J.~G. Abraldes}, \bibinfo{author}{A.~Albillos}, \bibinfo{author}{A.~Baiges}, \bibinfo{author}{J.~Bajaj}, \bibinfo{author}{R.~Ba{\~n}ares}, et~al.,
\newblock \bibinfo{title}{Baveno vii--renewing consensus in portal hypertension},
\newblock \bibinfo{journal}{Journal of hepatology} \bibinfo{volume}{76} (\bibinfo{year}{2022}) \bibinfo{pages}{959--974}.
%Type = Article
\bibitem[{Lee et~al.(2024)Lee, Eghtesad, Garcia-Tsao, Haskal, Hernandez-Gea, Jalaeian, Kalva, Mohanty, Thabut, and Abraldes}]{aasld}
\bibinfo{author}{E.~W. Lee}, \bibinfo{author}{B.~Eghtesad}, \bibinfo{author}{G.~Garcia-Tsao}, \bibinfo{author}{Z.~J. Haskal}, \bibinfo{author}{V.~Hernandez-Gea}, \bibinfo{author}{H.~Jalaeian}, \bibinfo{author}{S.~P. Kalva}, \bibinfo{author}{A.~Mohanty}, \bibinfo{author}{D.~Thabut}, \bibinfo{author}{J.~G. Abraldes},
\newblock \bibinfo{title}{Aasld practice guidance on the use of tips, variceal embolization, and retrograde transvenous obliteration in the management of variceal hemorrhage},
\newblock \bibinfo{journal}{Hepatology} \bibinfo{volume}{79} (\bibinfo{year}{2024}) \bibinfo{pages}{224--250}.
%Type = Article
\bibitem[{{European Association For The Study Of The Liver and others}(2025)}]{easl}
\bibinfo{author}{{European Association For The Study Of The Liver and others}},
\newblock \bibinfo{title}{Easl clinical practice guidelines on tips},
\newblock \bibinfo{journal}{Journal of hepatology} \bibinfo{volume}{83} (\bibinfo{year}{2025}) \bibinfo{pages}{177--210}.
%Type = Inproceedings
\bibitem[{Yushkevich et~al.(2016)Yushkevich, Gao, and Gerig}]{itk-snap}
\bibinfo{author}{P.~A. Yushkevich}, \bibinfo{author}{Y.~Gao}, \bibinfo{author}{G.~Gerig},
\newblock \bibinfo{title}{Itk-snap: An interactive tool for semi-automatic segmentation of multi-modality biomedical images},
\newblock in: \bibinfo{booktitle}{2016 38th annual international conference of the IEEE engineering in medicine and biology society (EMBC)}, \bibinfo{organization}{IEEE}, \bibinfo{year}{2016}, pp. \bibinfo{pages}{3342--3345}.
%Type = Article
\bibitem[{Garcia-Tsao et~al.(2017)Garcia-Tsao, Abraldes, Berzigotti, and Bosch}]{tips_guideline_2016}
\bibinfo{author}{G.~Garcia-Tsao}, \bibinfo{author}{J.~G. Abraldes}, \bibinfo{author}{A.~Berzigotti}, \bibinfo{author}{J.~Bosch},
\newblock \bibinfo{title}{Portal hypertensive bleeding in cirrhosis: Risk stratification, diagnosis, and management: 2016 practice guidance by the american association for the study of liver diseases},
\newblock \bibinfo{journal}{Hepatology} \bibinfo{volume}{65} (\bibinfo{year}{2017}) \bibinfo{pages}{310--335}.
%Type = Inproceedings
\bibitem[{Ryali et~al.(2023)Ryali, Hu, Bolya, Wei, Fan, Huang, Aggarwal, Chowdhury, Poursaeed, Hoffman et~al.}]{hiera}
\bibinfo{author}{C.~Ryali}, \bibinfo{author}{Y.-T. Hu}, \bibinfo{author}{D.~Bolya}, \bibinfo{author}{C.~Wei}, \bibinfo{author}{H.~Fan}, \bibinfo{author}{P.-Y. Huang}, \bibinfo{author}{V.~Aggarwal}, \bibinfo{author}{A.~Chowdhury}, \bibinfo{author}{O.~Poursaeed}, \bibinfo{author}{J.~Hoffman}, et~al.,
\newblock \bibinfo{title}{Hiera: A hierarchical vision transformer without the bells-and-whistles},
\newblock in: \bibinfo{booktitle}{International conference on machine learning}, \bibinfo{organization}{PMLR}, \bibinfo{year}{2023}, pp. \bibinfo{pages}{29441--29454}.
%Type = Article
\bibitem[{Ma et~al.(2024)Ma, Kim, Li, Baharoon, Asakereh, Lyu, and Wang}]{medsam2_ours}
\bibinfo{author}{J.~Ma}, \bibinfo{author}{S.~Kim}, \bibinfo{author}{F.~Li}, \bibinfo{author}{M.~Baharoon}, \bibinfo{author}{R.~Asakereh}, \bibinfo{author}{H.~Lyu}, \bibinfo{author}{B.~Wang},
\newblock \bibinfo{title}{Segment anything in medical images and videos: Benchmark and deployment},
\newblock \bibinfo{journal}{arXiv preprint arXiv:2408.03322}  (\bibinfo{year}{2024}).
%Type = Inproceedings
\bibitem[{He et~al.(2016)He, Zhang, Ren, and Sun}]{resnet}
\bibinfo{author}{K.~He}, \bibinfo{author}{X.~Zhang}, \bibinfo{author}{S.~Ren}, \bibinfo{author}{J.~Sun},
\newblock \bibinfo{title}{Deep residual learning for image recognition},
\newblock in: \bibinfo{booktitle}{Proceedings of the IEEE conference on computer vision and pattern recognition}, \bibinfo{year}{2016}, pp. \bibinfo{pages}{770--778}.
%Type = Inproceedings
\bibitem[{Chen et~al.(2021)Chen, Lu, Weng, Chen, Williamson, Manz, Shady, and Mahmood}]{mcat}
\bibinfo{author}{R.~J. Chen}, \bibinfo{author}{M.~Y. Lu}, \bibinfo{author}{W.-H. Weng}, \bibinfo{author}{T.~Y. Chen}, \bibinfo{author}{D.~F. Williamson}, \bibinfo{author}{T.~Manz}, \bibinfo{author}{M.~Shady}, \bibinfo{author}{F.~Mahmood},
\newblock \bibinfo{title}{Multimodal co-attention transformer for survival prediction in gigapixel whole slide images},
\newblock in: \bibinfo{booktitle}{Proceedings of the IEEE/CVF international conference on computer vision}, \bibinfo{year}{2021}, pp. \bibinfo{pages}{4015--4025}.
%Type = Inproceedings
\bibitem[{Xu and Chen(2023)}]{motcat}
\bibinfo{author}{Y.~Xu}, \bibinfo{author}{H.~Chen},
\newblock \bibinfo{title}{Multimodal optimal transport-based co-attention transformer with global structure consistency for survival prediction},
\newblock in: \bibinfo{booktitle}{Proceedings of the IEEE/CVF international conference on computer vision}, \bibinfo{year}{2023}, pp. \bibinfo{pages}{21241--21251}.
%Type = Inproceedings
\bibitem[{Jaume et~al.(2024)Jaume, Vaidya, Chen, Williamson, Liang, and Mahmood}]{survpath}
\bibinfo{author}{G.~Jaume}, \bibinfo{author}{A.~Vaidya}, \bibinfo{author}{R.~J. Chen}, \bibinfo{author}{D.~F. Williamson}, \bibinfo{author}{P.~P. Liang}, \bibinfo{author}{F.~Mahmood},
\newblock \bibinfo{title}{Modeling dense multimodal interactions between biological pathways and histology for survival prediction},
\newblock in: \bibinfo{booktitle}{Proceedings of the IEEE/CVF Conference on Computer Vision and Pattern Recognition}, \bibinfo{year}{2024}, pp. \bibinfo{pages}{11579--11590}.
%Type = Article
\bibitem[{Klambauer et~al.(2017)Klambauer, Unterthiner, Mayr, and Hochreiter}]{snn}
\bibinfo{author}{G.~Klambauer}, \bibinfo{author}{T.~Unterthiner}, \bibinfo{author}{A.~Mayr}, \bibinfo{author}{S.~Hochreiter},
\newblock \bibinfo{title}{Self-normalizing neural networks},
\newblock \bibinfo{journal}{Advances in neural information processing systems} \bibinfo{volume}{30} (\bibinfo{year}{2017}).
%Type = Inproceedings
\bibitem[{Song et~al.(2024)Song, Chen, Jaume, Vaidya, Baras, and Mahmood}]{mmp}
\bibinfo{author}{A.~H. Song}, \bibinfo{author}{R.~J. Chen}, \bibinfo{author}{G.~Jaume}, \bibinfo{author}{A.~Vaidya}, \bibinfo{author}{A.~S. Baras}, \bibinfo{author}{F.~Mahmood},
\newblock \bibinfo{title}{Multimodal prototyping for cancer survival prediction},
\newblock in: \bibinfo{booktitle}{Proceedings of the 41st International Conference on Machine Learning}, \bibinfo{year}{2024}, pp. \bibinfo{pages}{46050--46073}.
%Type = Article
\bibitem[{Zhang et~al.(2025)Zhang, Meng, Dong, Su, and Zhao}]{icfnet}
\bibinfo{author}{B.~Zhang}, \bibinfo{author}{Z.~Meng}, \bibinfo{author}{J.~Dong}, \bibinfo{author}{F.~Su}, \bibinfo{author}{Z.~Zhao},
\newblock \bibinfo{title}{Icfnet: Integrated cross-modal fusion network for survival prediction},
\newblock \bibinfo{journal}{arXiv preprint arXiv:2501.02778}  (\bibinfo{year}{2025}).
%Type = Article
\bibitem[{Kantorovich(2006)}]{kantorovich}
\bibinfo{author}{L.~V. Kantorovich},
\newblock \bibinfo{title}{On the translocation of masses.},
\newblock \bibinfo{journal}{Journal of mathematical sciences} \bibinfo{volume}{133} (\bibinfo{year}{2006}).
%Type = Article
\bibitem[{Frogner et~al.(2015)Frogner, Zhang, Mobahi, Araya, and Poggio}]{Sinkhorn-Knopp1}
\bibinfo{author}{C.~Frogner}, \bibinfo{author}{C.~Zhang}, \bibinfo{author}{H.~Mobahi}, \bibinfo{author}{M.~Araya}, \bibinfo{author}{T.~A. Poggio},
\newblock \bibinfo{title}{Learning with a wasserstein loss},
\newblock \bibinfo{journal}{Advances in neural information processing systems} \bibinfo{volume}{28} (\bibinfo{year}{2015}).
%Type = Article
\bibitem[{Chizat et~al.(2016)Chizat, Peyr{\'e}, Schmitzer, and Vialard}]{Sinkhorn-Knopp2}
\bibinfo{author}{L.~Chizat}, \bibinfo{author}{G.~Peyr{\'e}}, \bibinfo{author}{B.~Schmitzer}, \bibinfo{author}{F.-X. Vialard},
\newblock \bibinfo{title}{Scaling algorithms for unbalanced transport problems},
\newblock \bibinfo{journal}{arXiv preprint arXiv:1607.05816}  (\bibinfo{year}{2016}).
%Type = Inproceedings
\bibitem[{Ilse et~al.(2018)Ilse, Tomczak, and Welling}]{attnmil}
\bibinfo{author}{M.~Ilse}, \bibinfo{author}{J.~Tomczak}, \bibinfo{author}{M.~Welling},
\newblock \bibinfo{title}{Attention-based deep multiple instance learning},
\newblock in: \bibinfo{booktitle}{International conference on machine learning}, \bibinfo{organization}{PMLR}, \bibinfo{year}{2018}, pp. \bibinfo{pages}{2127--2136}.
%Type = Inproceedings
\bibitem[{Li et~al.(2021)Li, Yang, Xing, Zhao, Zhang, Liu, Han, Huang, Wang, and Yao}]{tab_cli}
\bibinfo{author}{H.~Li}, \bibinfo{author}{F.~Yang}, \bibinfo{author}{X.~Xing}, \bibinfo{author}{Y.~Zhao}, \bibinfo{author}{J.~Zhang}, \bibinfo{author}{Y.~Liu}, \bibinfo{author}{M.~Han}, \bibinfo{author}{J.~Huang}, \bibinfo{author}{L.~Wang}, \bibinfo{author}{J.~Yao},
\newblock \bibinfo{title}{Multi-modal multi-instance learning using weakly correlated histopathological images and tabular clinical information},
\newblock in: \bibinfo{booktitle}{International Conference on Medical Image Computing and Computer-Assisted Intervention}, \bibinfo{organization}{Springer}, \bibinfo{year}{2021}, pp. \bibinfo{pages}{529--539}.
%Type = Inproceedings
\bibitem[{Zhou et~al.(2025)Zhou, Tang, Zuo, Wan, Zhang, and Shao}]{ld-cvae}
\bibinfo{author}{J.~Zhou}, \bibinfo{author}{J.~Tang}, \bibinfo{author}{Y.~Zuo}, \bibinfo{author}{P.~Wan}, \bibinfo{author}{D.~Zhang}, \bibinfo{author}{W.~Shao},
\newblock \bibinfo{title}{Robust multimodal survival prediction with conditional latent differentiation variational autoencoder},
\newblock in: \bibinfo{booktitle}{Proceedings of the Computer Vision and Pattern Recognition Conference}, \bibinfo{year}{2025}, pp. \bibinfo{pages}{10384--10393}.
%Type = Article
\bibitem[{Zadeh and Schmid(2020)}]{nllloss}
\bibinfo{author}{S.~G. Zadeh}, \bibinfo{author}{M.~Schmid},
\newblock \bibinfo{title}{Bias in cross-entropy-based training of deep survival networks},
\newblock \bibinfo{journal}{IEEE transactions on pattern analysis and machine intelligence} \bibinfo{volume}{43} (\bibinfo{year}{2020}) \bibinfo{pages}{3126--3137}.
%Type = Inproceedings
\bibitem[{Chen et~al.(2018)Chen, Zhu, Papandreou, Schroff, and Adam}]{deeplabv3+}
\bibinfo{author}{L.-C. Chen}, \bibinfo{author}{Y.~Zhu}, \bibinfo{author}{G.~Papandreou}, \bibinfo{author}{F.~Schroff}, \bibinfo{author}{H.~Adam},
\newblock \bibinfo{title}{Encoder-decoder with atrous separable convolution for semantic image segmentation},
\newblock in: \bibinfo{booktitle}{Proceedings of the European conference on computer vision (ECCV)}, \bibinfo{year}{2018}, pp. \bibinfo{pages}{801--818}.
%Type = Article
\bibitem[{Soler et~al.(2010)Soler, Hostettler, Agnus, Charnoz, Fasquel, Moreau, Osswald, Bouhadjar, and Marescaux}]{3d-ircadb}
\bibinfo{author}{L.~Soler}, \bibinfo{author}{A.~Hostettler}, \bibinfo{author}{V.~Agnus}, \bibinfo{author}{A.~Charnoz}, \bibinfo{author}{J.-B. Fasquel}, \bibinfo{author}{J.~Moreau}, \bibinfo{author}{A.-B. Osswald}, \bibinfo{author}{M.~Bouhadjar}, \bibinfo{author}{J.~Marescaux},
\newblock \bibinfo{title}{3d image reconstruction for comparison of algorithm database},
\newblock \bibinfo{journal}{URL: https://www.ircad.fr/research/data-sets/liver-segmentation-3d-ircadb-01} \bibinfo{volume}{13} (\bibinfo{year}{2010}).
%Type = Article
\bibitem[{Micikevicius et~al.(2017)Micikevicius, Narang, Alben, Diamos, Elsen, Garcia, Ginsburg, Houston, Kuchaiev, Venkatesh et~al.}]{mixed}
\bibinfo{author}{P.~Micikevicius}, \bibinfo{author}{S.~Narang}, \bibinfo{author}{J.~Alben}, \bibinfo{author}{G.~Diamos}, \bibinfo{author}{E.~Elsen}, \bibinfo{author}{D.~Garcia}, \bibinfo{author}{B.~Ginsburg}, \bibinfo{author}{M.~Houston}, \bibinfo{author}{O.~Kuchaiev}, \bibinfo{author}{G.~Venkatesh}, et~al.,
\newblock \bibinfo{title}{Mixed precision training},
\newblock \bibinfo{journal}{arXiv preprint arXiv:1710.03740}  (\bibinfo{year}{2017}).
%Type = Article
\bibitem[{Wang et~al.(2019)Wang, Li, and Reddy}]{brier_score}
\bibinfo{author}{P.~Wang}, \bibinfo{author}{Y.~Li}, \bibinfo{author}{C.~K. Reddy},
\newblock \bibinfo{title}{Machine learning for survival analysis: A survey},
\newblock \bibinfo{journal}{ACM Computing Surveys (CSUR)} \bibinfo{volume}{51} (\bibinfo{year}{2019}) \bibinfo{pages}{1--36}.
%Type = Article
\bibitem[{Kaplan and Meier(1958)}]{KM}
\bibinfo{author}{E.~L. Kaplan}, \bibinfo{author}{P.~Meier},
\newblock \bibinfo{title}{Nonparametric estimation from incomplete observations},
\newblock \bibinfo{journal}{Journal of the American statistical association} \bibinfo{volume}{53} (\bibinfo{year}{1958}) \bibinfo{pages}{457--481}.
%Type = Article
\bibitem[{Mantel et~al.(1966)}]{logrank}
\bibinfo{author}{N.~Mantel}, et~al.,
\newblock \bibinfo{title}{Evaluation of survival data and two new rank order statistics arising in its consideration},
\newblock \bibinfo{journal}{Cancer Chemother Rep} \bibinfo{volume}{50} (\bibinfo{year}{1966}) \bibinfo{pages}{163--170}.
%Type = Article
\bibitem[{Sohn et~al.(2020)Sohn, Berthelot, Carlini, Zhang, Zhang, Raffel, Cubuk, Kurakin, and Li}]{fixmatch}
\bibinfo{author}{K.~Sohn}, \bibinfo{author}{D.~Berthelot}, \bibinfo{author}{N.~Carlini}, \bibinfo{author}{Z.~Zhang}, \bibinfo{author}{H.~Zhang}, \bibinfo{author}{C.~A. Raffel}, \bibinfo{author}{E.~D. Cubuk}, \bibinfo{author}{A.~Kurakin}, \bibinfo{author}{C.-L. Li},
\newblock \bibinfo{title}{Fixmatch: Simplifying semi-supervised learning with consistency and confidence},
\newblock \bibinfo{journal}{Advances in neural information processing systems} \bibinfo{volume}{33} (\bibinfo{year}{2020}) \bibinfo{pages}{596--608}.
%Type = Inproceedings
\bibitem[{Sun et~al.(2024)Sun, Yang, Zhang, Cheng, and Hou}]{corrmatch}
\bibinfo{author}{B.~Sun}, \bibinfo{author}{Y.~Yang}, \bibinfo{author}{L.~Zhang}, \bibinfo{author}{M.-M. Cheng}, \bibinfo{author}{Q.~Hou},
\newblock \bibinfo{title}{Corrmatch: Label propagation via correlation matching for semi-supervised semantic segmentation},
\newblock in: \bibinfo{booktitle}{Proceedings of the IEEE/CVF conference on computer vision and pattern recognition}, \bibinfo{year}{2024}, pp. \bibinfo{pages}{3097--3107}.
%Type = Inproceedings
\bibitem[{Wang et~al.(2024)Wang, Bai, Yu, Zhao, and Xiao}]{ddfp}
\bibinfo{author}{X.~Wang}, \bibinfo{author}{H.~Bai}, \bibinfo{author}{L.~Yu}, \bibinfo{author}{Y.~Zhao}, \bibinfo{author}{J.~Xiao},
\newblock \bibinfo{title}{Towards the uncharted: Density-descending feature perturbation for semi-supervised semantic segmentation},
\newblock in: \bibinfo{booktitle}{Proceedings of the IEEE/CVF Conference on Computer Vision and Pattern Recognition}, \bibinfo{year}{2024}, pp. \bibinfo{pages}{3303--3312}.
%Type = Inproceedings
\bibitem[{Kirillov et~al.(2023)Kirillov, Mintun, Ravi, Mao, Rolland, Gustafson, Xiao, Whitehead, Berg, Lo et~al.}]{sam}
\bibinfo{author}{A.~Kirillov}, \bibinfo{author}{E.~Mintun}, \bibinfo{author}{N.~Ravi}, \bibinfo{author}{H.~Mao}, \bibinfo{author}{C.~Rolland}, \bibinfo{author}{L.~Gustafson}, \bibinfo{author}{T.~Xiao}, \bibinfo{author}{S.~Whitehead}, \bibinfo{author}{A.~C. Berg}, \bibinfo{author}{W.-Y. Lo}, et~al.,
\newblock \bibinfo{title}{Segment anything},
\newblock in: \bibinfo{booktitle}{Proceedings of the IEEE/CVF international conference on computer vision}, \bibinfo{year}{2023}, pp. \bibinfo{pages}{4015--4026}.
%Type = Article
\bibitem[{Ma et~al.(2024)Ma, He, Li, Han, You, and Wang}]{medsam}
\bibinfo{author}{J.~Ma}, \bibinfo{author}{Y.~He}, \bibinfo{author}{F.~Li}, \bibinfo{author}{L.~Han}, \bibinfo{author}{C.~You}, \bibinfo{author}{B.~Wang},
\newblock \bibinfo{title}{Segment anything in medical images},
\newblock \bibinfo{journal}{Nature Communications} \bibinfo{volume}{15} (\bibinfo{year}{2024}) \bibinfo{pages}{654}.
%Type = Article
\bibitem[{Shao et~al.(2021)Shao, Bian, Chen, Wang, Zhang, Ji et~al.}]{transmil}
\bibinfo{author}{Z.~Shao}, \bibinfo{author}{H.~Bian}, \bibinfo{author}{Y.~Chen}, \bibinfo{author}{Y.~Wang}, \bibinfo{author}{J.~Zhang}, \bibinfo{author}{X.~Ji}, et~al.,
\newblock \bibinfo{title}{Transmil: Transformer based correlated multiple instance learning for whole slide image classification},
\newblock \bibinfo{journal}{Advances in neural information processing systems} \bibinfo{volume}{34} (\bibinfo{year}{2021}) \bibinfo{pages}{2136--2147}.
%Type = Article
\bibitem[{Elmarakeby et~al.(2021)Elmarakeby, Hwang, Arafeh, Crowdis, Gang, Liu, AlDubayan, Salari, Kregel, Richter et~al.}]{s-mlp}
\bibinfo{author}{H.~A. Elmarakeby}, \bibinfo{author}{J.~Hwang}, \bibinfo{author}{R.~Arafeh}, \bibinfo{author}{J.~Crowdis}, \bibinfo{author}{S.~Gang}, \bibinfo{author}{D.~Liu}, \bibinfo{author}{S.~H. AlDubayan}, \bibinfo{author}{K.~Salari}, \bibinfo{author}{S.~Kregel}, \bibinfo{author}{C.~Richter}, et~al.,
\newblock \bibinfo{title}{Biologically informed deep neural network for prostate cancer discovery},
\newblock \bibinfo{journal}{Nature} \bibinfo{volume}{598} (\bibinfo{year}{2021}) \bibinfo{pages}{348--352}.
%Type = Article
\bibitem[{Zaheer et~al.(2017)Zaheer, Kottur, Ravanbakhsh, Poczos, Salakhutdinov, and Smola}]{deepsets}
\bibinfo{author}{M.~Zaheer}, \bibinfo{author}{S.~Kottur}, \bibinfo{author}{S.~Ravanbakhsh}, \bibinfo{author}{B.~Poczos}, \bibinfo{author}{R.~R. Salakhutdinov}, \bibinfo{author}{A.~J. Smola},
\newblock \bibinfo{title}{Deep sets},
\newblock \bibinfo{journal}{Advances in neural information processing systems} \bibinfo{volume}{30} (\bibinfo{year}{2017}).
%Type = Article
\bibitem[{Yao et~al.(2020)Yao, Zhu, Jonnagaddala, Hawkins, and Huang}]{deepattnmisl}
\bibinfo{author}{J.~Yao}, \bibinfo{author}{X.~Zhu}, \bibinfo{author}{J.~Jonnagaddala}, \bibinfo{author}{N.~Hawkins}, \bibinfo{author}{J.~Huang},
\newblock \bibinfo{title}{Whole slide images based cancer survival prediction using attention guided deep multiple instance learning networks},
\newblock \bibinfo{journal}{Medical image analysis} \bibinfo{volume}{65} (\bibinfo{year}{2020}) \bibinfo{pages}{101789}.
%Type = Article
\bibitem[{Lu et~al.(2021)Lu, Williamson, Chen, Chen, Barbieri, and Mahmood}]{clam}
\bibinfo{author}{M.~Y. Lu}, \bibinfo{author}{D.~F. Williamson}, \bibinfo{author}{T.~Y. Chen}, \bibinfo{author}{R.~J. Chen}, \bibinfo{author}{M.~Barbieri}, \bibinfo{author}{F.~Mahmood},
\newblock \bibinfo{title}{Data-efficient and weakly supervised computational pathology on whole-slide images},
\newblock \bibinfo{journal}{Nature biomedical engineering} \bibinfo{volume}{5} (\bibinfo{year}{2021}) \bibinfo{pages}{555--570}.
%Type = Inproceedings
\bibitem[{Zhang et~al.(2022)Zhang, Meng, Zhao, Qiao, Yang, Coupland, and Zheng}]{dtfd}
\bibinfo{author}{H.~Zhang}, \bibinfo{author}{Y.~Meng}, \bibinfo{author}{Y.~Zhao}, \bibinfo{author}{Y.~Qiao}, \bibinfo{author}{X.~Yang}, \bibinfo{author}{S.~E. Coupland}, \bibinfo{author}{Y.~Zheng},
\newblock \bibinfo{title}{Dtfd-mil: Double-tier feature distillation multiple instance learning for histopathology whole slide image classification},
\newblock in: \bibinfo{booktitle}{Proceedings of the IEEE/CVF conference on computer vision and pattern recognition}, \bibinfo{year}{2022}, pp. \bibinfo{pages}{18802--18812}.
%Type = Inproceedings
\bibitem[{Xiang and Zhang(2023)}]{ilra}
\bibinfo{author}{J.~Xiang}, \bibinfo{author}{J.~Zhang},
\newblock \bibinfo{title}{Exploring low-rank property in multiple instance learning for whole slide image classification},
\newblock in: \bibinfo{booktitle}{The Eleventh International Conference on Learning Representations}, \bibinfo{year}{2023}.
%Type = Inproceedings
\bibitem[{Li et~al.(2024)Li, Chen, Chu, Sun, Guan, Han, and He}]{wikg}
\bibinfo{author}{J.~Li}, \bibinfo{author}{Y.~Chen}, \bibinfo{author}{H.~Chu}, \bibinfo{author}{Q.~Sun}, \bibinfo{author}{T.~Guan}, \bibinfo{author}{A.~Han}, \bibinfo{author}{Y.~He},
\newblock \bibinfo{title}{Dynamic graph representation with knowledge-aware attention for histopathology whole slide image analysis},
\newblock in: \bibinfo{booktitle}{Proceedings of the IEEE/CVF conference on computer vision and pattern recognition}, \bibinfo{year}{2024}, pp. \bibinfo{pages}{11323--11332}.
%Type = Inproceedings
\bibitem[{Ren et~al.(2025)Ren, Wang, Fang, Ling, and You}]{otsurv}
\bibinfo{author}{Q.~Ren}, \bibinfo{author}{Y.~Wang}, \bibinfo{author}{R.~Fang}, \bibinfo{author}{H.~Ling}, \bibinfo{author}{C.~You},
\newblock \bibinfo{title}{Otsurv: A novel multiple instance learning framework for survival prediction with heterogeneity-aware optimal transport},
\newblock in: \bibinfo{booktitle}{International Conference on Medical Image Computing and Computer-Assisted Intervention}, \bibinfo{organization}{Springer}, \bibinfo{year}{2025}, pp. \bibinfo{pages}{439--449}.
%Type = Article
\bibitem[{Chen et~al.(2020)Chen, Lu, Wang, Williamson, Rodig, Lindeman, and Mahmood}]{pathomic}
\bibinfo{author}{R.~J. Chen}, \bibinfo{author}{M.~Y. Lu}, \bibinfo{author}{J.~Wang}, \bibinfo{author}{D.~F. Williamson}, \bibinfo{author}{S.~J. Rodig}, \bibinfo{author}{N.~I. Lindeman}, \bibinfo{author}{F.~Mahmood},
\newblock \bibinfo{title}{Pathomic fusion: an integrated framework for fusing histopathology and genomic features for cancer diagnosis and prognosis},
\newblock \bibinfo{journal}{IEEE Transactions on Medical Imaging} \bibinfo{volume}{41} (\bibinfo{year}{2020}) \bibinfo{pages}{757--770}.
%Type = Article
\bibitem[{Chen et~al.(2022)Chen, Lu, Williamson, Chen, Lipkova, Noor, Shaban, Shady, Williams, Joo et~al.}]{porpoise}
\bibinfo{author}{R.~J. Chen}, \bibinfo{author}{M.~Y. Lu}, \bibinfo{author}{D.~F. Williamson}, \bibinfo{author}{T.~Y. Chen}, \bibinfo{author}{J.~Lipkova}, \bibinfo{author}{Z.~Noor}, \bibinfo{author}{M.~Shaban}, \bibinfo{author}{M.~Shady}, \bibinfo{author}{M.~Williams}, \bibinfo{author}{B.~Joo}, et~al.,
\newblock \bibinfo{title}{Pan-cancer integrative histology-genomic analysis via multimodal deep learning},
\newblock \bibinfo{journal}{Cancer cell} \bibinfo{volume}{40} (\bibinfo{year}{2022}) \bibinfo{pages}{865--878}.
%Type = Inproceedings
\bibitem[{Zhou and Chen(2023)}]{cmta}
\bibinfo{author}{F.~Zhou}, \bibinfo{author}{H.~Chen},
\newblock \bibinfo{title}{Cross-modal translation and alignment for survival analysis},
\newblock in: \bibinfo{booktitle}{Proceedings of the IEEE/CVF International Conference on Computer Vision}, \bibinfo{year}{2023}, pp. \bibinfo{pages}{21485--21494}.
%Type = Inproceedings
\bibitem[{Xiong et~al.(2024)Xiong, Chen, Zheng, Wei, Zheng, Sung, and King}]{mome}
\bibinfo{author}{C.~Xiong}, \bibinfo{author}{H.~Chen}, \bibinfo{author}{H.~Zheng}, \bibinfo{author}{D.~Wei}, \bibinfo{author}{Y.~Zheng}, \bibinfo{author}{J.~J. Sung}, \bibinfo{author}{I.~King},
\newblock \bibinfo{title}{Mome: Mixture of multimodal experts for cancer survival prediction},
\newblock in: \bibinfo{booktitle}{International Conference on Medical Image Computing and Computer-Assisted Intervention}, \bibinfo{organization}{Springer}, \bibinfo{year}{2024}, pp. \bibinfo{pages}{318--328}.
%Type = Inproceedings
\bibitem[{Eijpe et~al.(2025)Eijpe, Lakbir, Erdal~Cesur, Oliveira, Abeln, and Silva}]{dimaf}
\bibinfo{author}{A.~Eijpe}, \bibinfo{author}{S.~Lakbir}, \bibinfo{author}{M.~Erdal~Cesur}, \bibinfo{author}{S.~P. Oliveira}, \bibinfo{author}{S.~Abeln}, \bibinfo{author}{W.~Silva},
\newblock \bibinfo{title}{Disentangled and interpretable multimodal attention fusion for cancer survival prediction},
\newblock in: \bibinfo{booktitle}{International Conference on Medical Image Computing and Computer-Assisted Intervention}, \bibinfo{organization}{Springer}, \bibinfo{year}{2025}, pp. \bibinfo{pages}{117--127}.
%Type = Inproceedings
\bibitem[{Liu et~al.(2025)Liu, Cai, Wang, Fan, Yu, Chen, and Zhang}]{prosurv}
\bibinfo{author}{F.~Liu}, \bibinfo{author}{L.~Cai}, \bibinfo{author}{Z.~Wang}, \bibinfo{author}{Z.~Fan}, \bibinfo{author}{J.-g. Yu}, \bibinfo{author}{H.~Chen}, \bibinfo{author}{Y.~Zhang},
\newblock \bibinfo{title}{Prototype-guided cross-modal knowledge enhancement for adaptive survival prediction},
\newblock in: \bibinfo{booktitle}{International Conference on Medical Image Computing and Computer-Assisted Intervention}, \bibinfo{organization}{Springer}, \bibinfo{year}{2025}, pp. \bibinfo{pages}{522--532}.
%Type = Article
\bibitem[{Fu et~al.(2025)Fu, He, Zhang, Peng, Zhang, Tang, Liu, Cao, and Wang}]{hsfsurv}
\bibinfo{author}{B.~Fu}, \bibinfo{author}{J.~He}, \bibinfo{author}{X.~Zhang}, \bibinfo{author}{Y.~Peng}, \bibinfo{author}{Z.~Zhang}, \bibinfo{author}{Q.~Tang}, \bibinfo{author}{X.~Liu}, \bibinfo{author}{Y.~Cao}, \bibinfo{author}{R.~Wang},
\newblock \bibinfo{title}{Hsfsurv: A hybrid supervision framework at individual and feature levels for multimodal cancer survival analysis},
\newblock \bibinfo{journal}{Medical Image Analysis}  (\bibinfo{year}{2025}) \bibinfo{pages}{103810}.
%Type = Inproceedings
\bibitem[{Wu et~al.(2026)Wu, Jiang, Zhu, Shi, and Zheng}]{msrl}
\bibinfo{author}{K.~Wu}, \bibinfo{author}{Z.~Jiang}, \bibinfo{author}{X.~Zhu}, \bibinfo{author}{J.~Shi}, \bibinfo{author}{Y.~Zheng},
\newblock \bibinfo{title}{Histopathology-genomics multi-modal structural representation learning for data-efficient precision oncology},
\newblock in: \bibinfo{booktitle}{The Fourteenth International Conference on Learning Representations}, \bibinfo{year}{2026}. \URLprefix \url{https://openreview.net/forum?id=24QX6XpvSL}.
%Type = Inproceedings
\bibitem[{Zhang et~al.(2026)Zhang, Nanbo, Yang, Schmidhuber, and Gao}]{slotspe}
\bibinfo{author}{Y.~Zhang}, \bibinfo{author}{L.~Nanbo}, \bibinfo{author}{C.~Yang}, \bibinfo{author}{J.~Schmidhuber}, \bibinfo{author}{X.~Gao},
\newblock \bibinfo{title}{Structural prognostic event modeling for multimodal cancer survival analysis},
\newblock in: \bibinfo{booktitle}{The Fourteenth International Conference on Learning Representations}, \bibinfo{year}{2026}. \URLprefix \url{https://openreview.net/forum?id=WqCRSn2WAY}.
%Type = Inproceedings
\bibitem[{Sundararajan et~al.(2017)Sundararajan, Taly, and Yan}]{IG}
\bibinfo{author}{M.~Sundararajan}, \bibinfo{author}{A.~Taly}, \bibinfo{author}{Q.~Yan},
\newblock \bibinfo{title}{Axiomatic attribution for deep networks},
\newblock in: \bibinfo{booktitle}{International conference on machine learning}, \bibinfo{organization}{PMLR}, \bibinfo{year}{2017}, pp. \bibinfo{pages}{3319--3328}.
%Type = Article
\bibitem[{Malinchoc et~al.(2000)Malinchoc, Kamath, Gordon, Peine, Rank, and Ter~Borg}]{MELD}
\bibinfo{author}{M.~Malinchoc}, \bibinfo{author}{P.~S. Kamath}, \bibinfo{author}{F.~D. Gordon}, \bibinfo{author}{C.~J. Peine}, \bibinfo{author}{J.~Rank}, \bibinfo{author}{P.~C. Ter~Borg},
\newblock \bibinfo{title}{A model to predict poor survival in patients undergoing transjugular intrahepatic portosystemic shunts},
\newblock \bibinfo{journal}{Hepatology} \bibinfo{volume}{31} (\bibinfo{year}{2000}) \bibinfo{pages}{864--871}.
%Type = Article
\bibitem[{Pugh et~al.(1973)Pugh, Murray-Lyon, Dawson, Pietroni, and Williams}]{child-pugh}
\bibinfo{author}{R.~Pugh}, \bibinfo{author}{I.~Murray-Lyon}, \bibinfo{author}{J.~Dawson}, \bibinfo{author}{M.~Pietroni}, \bibinfo{author}{R.~Williams},
\newblock \bibinfo{title}{Transection of the oesophagus for bleeding oesophageal varices},
\newblock \bibinfo{journal}{British journal of surgery} \bibinfo{volume}{60} (\bibinfo{year}{1973}) \bibinfo{pages}{646--649}.
%Type = Article
\bibitem[{Bettinger et~al.(2021)Bettinger, Sturm, Pfaff, Hahn, Kloeckner, Volkwein, Praktiknjo, Lv, Han, Huber et~al.}]{FIPS}
\bibinfo{author}{D.~Bettinger}, \bibinfo{author}{L.~Sturm}, \bibinfo{author}{L.~Pfaff}, \bibinfo{author}{F.~Hahn}, \bibinfo{author}{R.~Kloeckner}, \bibinfo{author}{L.~Volkwein}, \bibinfo{author}{M.~Praktiknjo}, \bibinfo{author}{Y.~Lv}, \bibinfo{author}{G.~Han}, \bibinfo{author}{J.~P. Huber}, et~al.,
\newblock \bibinfo{title}{Refining prediction of survival after tips with the novel freiburg index of post-tips survival},
\newblock \bibinfo{journal}{Journal of hepatology} \bibinfo{volume}{74} (\bibinfo{year}{2021}) \bibinfo{pages}{1362--1372}.
%Type = Article
\bibitem[{Angermayr et~al.(2003)Angermayr, Cejna, Karnel, Gschwantler, Koenig, Pidlich, Mendel, Pichler, Wichlas, Kreil et~al.}]{meld_child_survival}
\bibinfo{author}{B.~Angermayr}, \bibinfo{author}{M.~Cejna}, \bibinfo{author}{F.~Karnel}, \bibinfo{author}{M.~Gschwantler}, \bibinfo{author}{F.~Koenig}, \bibinfo{author}{J.~Pidlich}, \bibinfo{author}{H.~Mendel}, \bibinfo{author}{L.~Pichler}, \bibinfo{author}{M.~Wichlas}, \bibinfo{author}{A.~Kreil}, et~al.,
\newblock \bibinfo{title}{Child-pugh versus meld score in predicting survival in patients undergoing transjugular intrahepatic portosystemic shunt},
\newblock \bibinfo{journal}{Gut} \bibinfo{volume}{52} (\bibinfo{year}{2003}) \bibinfo{pages}{879--885}.
%Type = Article
\bibitem[{Bai et~al.(2011)Bai, Qi, Yang, Yin, Nie, Yuan, Wu, Han, and Fan}]{meld_child_ohe}
\bibinfo{author}{M.~Bai}, \bibinfo{author}{X.~Qi}, \bibinfo{author}{Z.~Yang}, \bibinfo{author}{Z.~Yin}, \bibinfo{author}{Y.~Nie}, \bibinfo{author}{S.~Yuan}, \bibinfo{author}{K.~Wu}, \bibinfo{author}{G.~Han}, \bibinfo{author}{D.~Fan},
\newblock \bibinfo{title}{Predictors of hepatic encephalopathy after transjugular intrahepatic portosystemic shunt in cirrhotic patients: a systematic review},
\newblock \bibinfo{journal}{Journal of gastroenterology and hepatology} \bibinfo{volume}{26} (\bibinfo{year}{2011}) \bibinfo{pages}{943--951}.
%Type = Inproceedings
\bibitem[{Dong et~al.(2023)Dong, Yao, Tang, Yuan, Xia, Zhou, Lu, Zhou, Dong, Lu et~al.}]{dv1}
\bibinfo{author}{H.~Dong}, \bibinfo{author}{J.~Yao}, \bibinfo{author}{Y.~Tang}, \bibinfo{author}{M.~Yuan}, \bibinfo{author}{Y.~Xia}, \bibinfo{author}{J.~Zhou}, \bibinfo{author}{H.~Lu}, \bibinfo{author}{J.~Zhou}, \bibinfo{author}{B.~Dong}, \bibinfo{author}{L.~Lu}, et~al.,
\newblock \bibinfo{title}{Improved prognostic prediction of pancreatic cancer using multi-phase ct by integrating neural distance and texture-aware transformer},
\newblock in: \bibinfo{booktitle}{International Conference on Medical Image Computing and Computer-Assisted Intervention}, \bibinfo{organization}{Springer}, \bibinfo{year}{2023}, pp. \bibinfo{pages}{241--251}.
%Type = Article
\bibitem[{Xiong et~al.(2025)Xiong, Yao, Lin, Yao, Bai, Huang, Zhang, Huang, Wang, Wang et~al.}]{dv2}
\bibinfo{author}{Y.~Xiong}, \bibinfo{author}{L.~Yao}, \bibinfo{author}{J.~Lin}, \bibinfo{author}{J.~Yao}, \bibinfo{author}{Q.~Bai}, \bibinfo{author}{Y.~Huang}, \bibinfo{author}{X.~Zhang}, \bibinfo{author}{R.~Huang}, \bibinfo{author}{R.~Wang}, \bibinfo{author}{K.~Wang}, et~al.,
\newblock \bibinfo{title}{Artificial intelligence links ct images to pathologic features and survival outcomes of renal masses},
\newblock \bibinfo{journal}{Nature Communications} \bibinfo{volume}{16} (\bibinfo{year}{2025}) \bibinfo{pages}{1425}.
%Type = Article
\bibitem[{Jin et~al.(2025)Jin, Xie, Li, Li, and Wang}]{cimp}
\bibinfo{author}{T.~Jin}, \bibinfo{author}{X.~Xie}, \bibinfo{author}{Q.~Li}, \bibinfo{author}{X.~Li}, \bibinfo{author}{Y.~Wang},
\newblock \bibinfo{title}{Clinical stage prompt induced multi-modal prognosis},
\newblock \bibinfo{journal}{IEEE Transactions on Medical Imaging}  (\bibinfo{year}{2025}).
%Type = Article
\bibitem[{Liu et~al.(2024)Liu, Ji, Gou, Fu, and Ye}]{vlsa}
\bibinfo{author}{P.~Liu}, \bibinfo{author}{L.~Ji}, \bibinfo{author}{J.~Gou}, \bibinfo{author}{B.~Fu}, \bibinfo{author}{M.~Ye},
\newblock \bibinfo{title}{Interpretable vision-language survival analysis with ordinal inductive bias for computational pathology},
\newblock \bibinfo{journal}{arXiv preprint arXiv:2409.09369}  (\bibinfo{year}{2024}).
%Type = Article
\bibitem[{Wang et~al.(2025)Wang, Zhang, Xu, Imoto, Chen, and Song}]{G-HANet}
\bibinfo{author}{Z.~Wang}, \bibinfo{author}{Y.~Zhang}, \bibinfo{author}{Y.~Xu}, \bibinfo{author}{S.~Imoto}, \bibinfo{author}{H.~Chen}, \bibinfo{author}{J.~Song},
\newblock \bibinfo{title}{Histo-genomic knowledge association for cancer prognosis from histopathology whole slide images},
\newblock \bibinfo{journal}{IEEE Transactions on Medical Imaging} \bibinfo{volume}{44} (\bibinfo{year}{2025}) \bibinfo{pages}{2170--2181}.

\end{thebibliography}

\end{document}